\pdfoutput=1
\documentclass[11pt]{article}

\usepackage[final]{acl}
\PassOptionsToPackage{hyphens}{url}
\usepackage{hyperref}
\usepackage{array}
\usepackage{xcolor}  % Optional: for colored links if you want
\usepackage{hyperref}
\usepackage{makecell}
\usepackage[normalem]{ulem}

\usepackage{times}
\usepackage{latexsym}
\usepackage{subcaption}

\usepackage[T1]{fontenc}
\usepackage[utf8]{inputenc}

\usepackage{microtype}
\usepackage{xspace}

\usepackage{inconsolata}

\usepackage{booktabs}
\usepackage{graphicx}    % For including graphics
\usepackage{caption}     % For customizing captions
\usepackage{subcaption}  % For creating subfigures
\usepackage{adjustbox} 
\usepackage{multirow}
\usepackage{placeins}

\usepackage{siunitx}
\usepackage{changepage}

\usepackage{pgf}       % needed for \pgfmathsetmacro

\usepackage{tcolorbox}
\tcbuselibrary{skins} 

\usepackage{longtable} % For full-page tables
\usepackage{array}     % For flexible column definitions
\usepackage{tabularx}
\newcolumntype{Y}{>{\raggedright\arraybackslash}X}

\usepackage{amsmath}
\usepackage{listings}
\lstdefinelanguage{Markdown}{
  basicstyle=\ttfamily\footnotesize,
  sensitive=false,
  morecomment=[l]{\#},   
  morecomment=[s]{```}{```},
  morestring=[b]",        
  morestring=[b]', 
}

\usepackage{xspace}
\usepackage{pgfmath}
\usepackage{makecell}
\usepackage{fontawesome5}
\usepackage{hyperref}
\usepackage{epigraph}

\definecolor{teal100}{RGB}{0, 90, 90}  
\definecolor{teal90}{RGB}{0, 110, 110} 
\definecolor{teal80}{RGB}{0, 150, 150}  
\definecolor{teal70}{RGB}{0, 220, 220}  
\definecolor{OIBlue}{HTML}{0072B2}      % good for Human
\definecolor{OIPurple}{HTML}{CC79A7}    % good for Mixed
\definecolor{OIVermil}{HTML}{D55E00}    % good for Fully AI
\definecolor{OIGreen}{HTML}{009E73}

\definecolor{burntorange}{RGB}{204, 85, 0}

\definecolor{tealbg}{RGB}{200,255,255}   % light teal 
\definecolor{orangebg}{RGB}{255,220,180} % light orange for background
\definecolor{orangetext}{RGB}{255,100,0} % orange text
\definecolor{purplebg}{RGB}{250,210,255} % light purple for background
\definecolor{lightredbg}{RGB}{255,220,220} % light red for background

\newcommand{\labelhuman}{\colorbox{tealbg}{\textsc{\textcolor{teal}{Human-written}}}}
\newcommand{\labelmixed}{\colorbox{orangebg}{\textsc{\textcolor{orangetext}{Mixed}}}}
\newcommand{\labelai}{\colorbox{purplebg}{\textsc{\textcolor{purple}{AI-Generated}}}}

\newcommand{\aiword}[1]{
  \begingroup
  \setlength{\fboxsep}{0.5pt}
  \colorbox{lightredbg}{\strut #1}
  \endgroup
}

\newcommand{\allowedicon}{\textcolor{green!60!black}{\faCheckCircle}}
\newcommand{\prohibitedicon}{\textcolor{red!70!black}{\faBan}}
\newcommand{\shrugicon}{\textcolor{gray!70}{\faQuestionCircle}}

\newcommand{\pangram}[1]{
  \href{#1}{\includegraphics[height=1em]{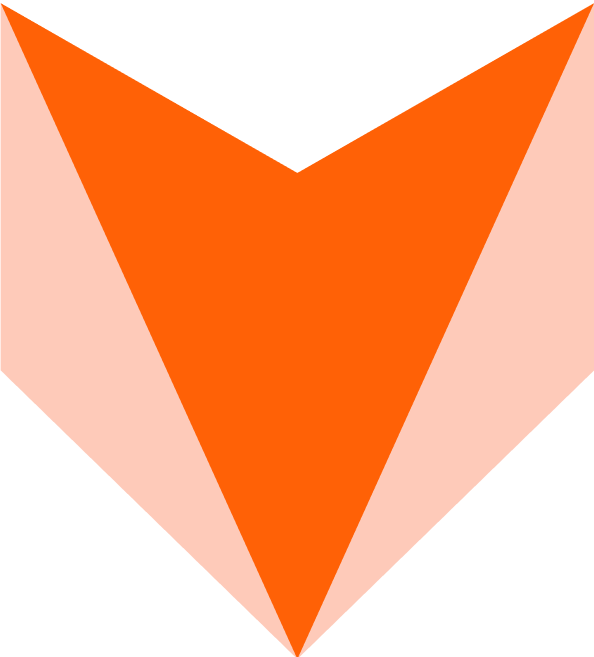}}
}

\newcommand{\news}[1]{
  \href{#1}{\includegraphics[height=1em]{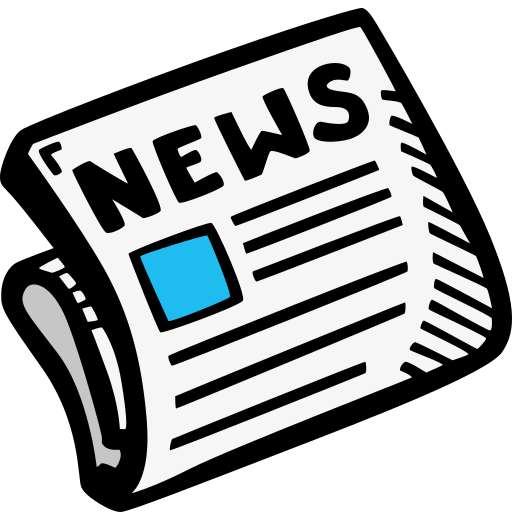}}
}

\newcommand{\wayback}[1]{
  \href{#1}{\includegraphics[height=1em]{figures/wayback-machine.svg}}
}

\usepackage[normalem]{ulem}

\definecolor{lightred}{HTML}{e99090}

\newcommand{\maindata}{\texttt{recent\_news}\xspace}
\newcommand{\opinionsdata}{\texttt{opinions}\xspace}
\newcommand{\reporterdata}{\texttt{ai\_reporters}\xspace}

\title{AI use in American newspapers is widespread, uneven, and rarely disclosed}

\author{%
\textbf{Jenna Russell\textsuperscript{\includegraphics[height=1.2em]{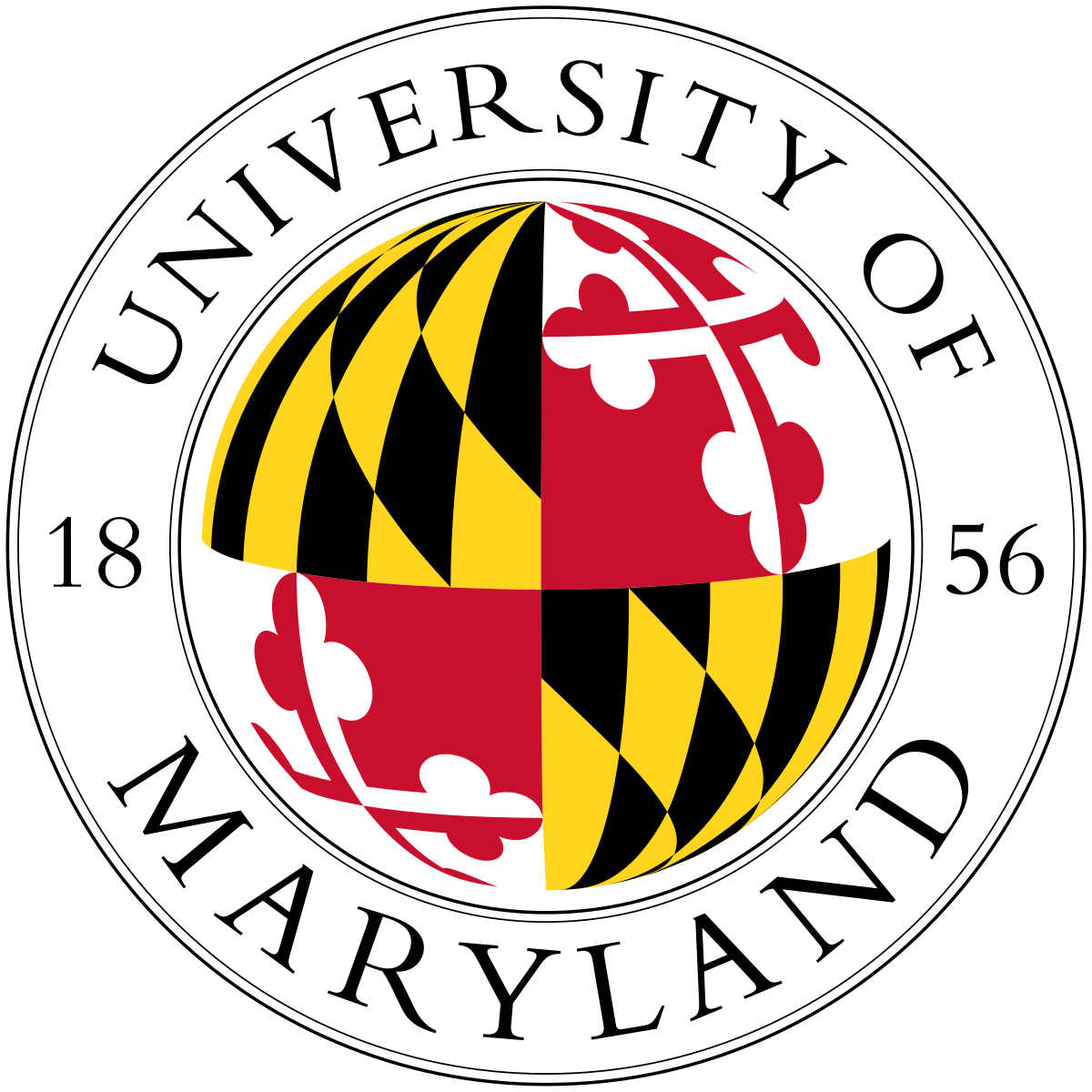}} \quad
Marzena Karpinska\textsuperscript{\includegraphics[height=1em]{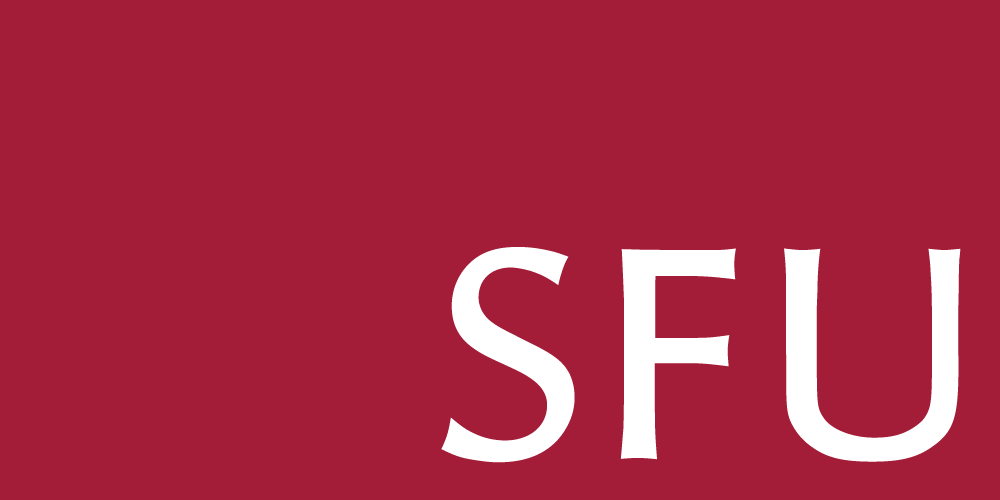}}} \quad
Destiny Akinode\textsuperscript{\includegraphics[height=1em]{figures/pangram.png}} \quad
James Zhou \textsuperscript{\includegraphics[height=1.2em]{figures/umd.png}} \protect\\
\textbf{Katherine Thai}\textsuperscript{\includegraphics[height=1em]{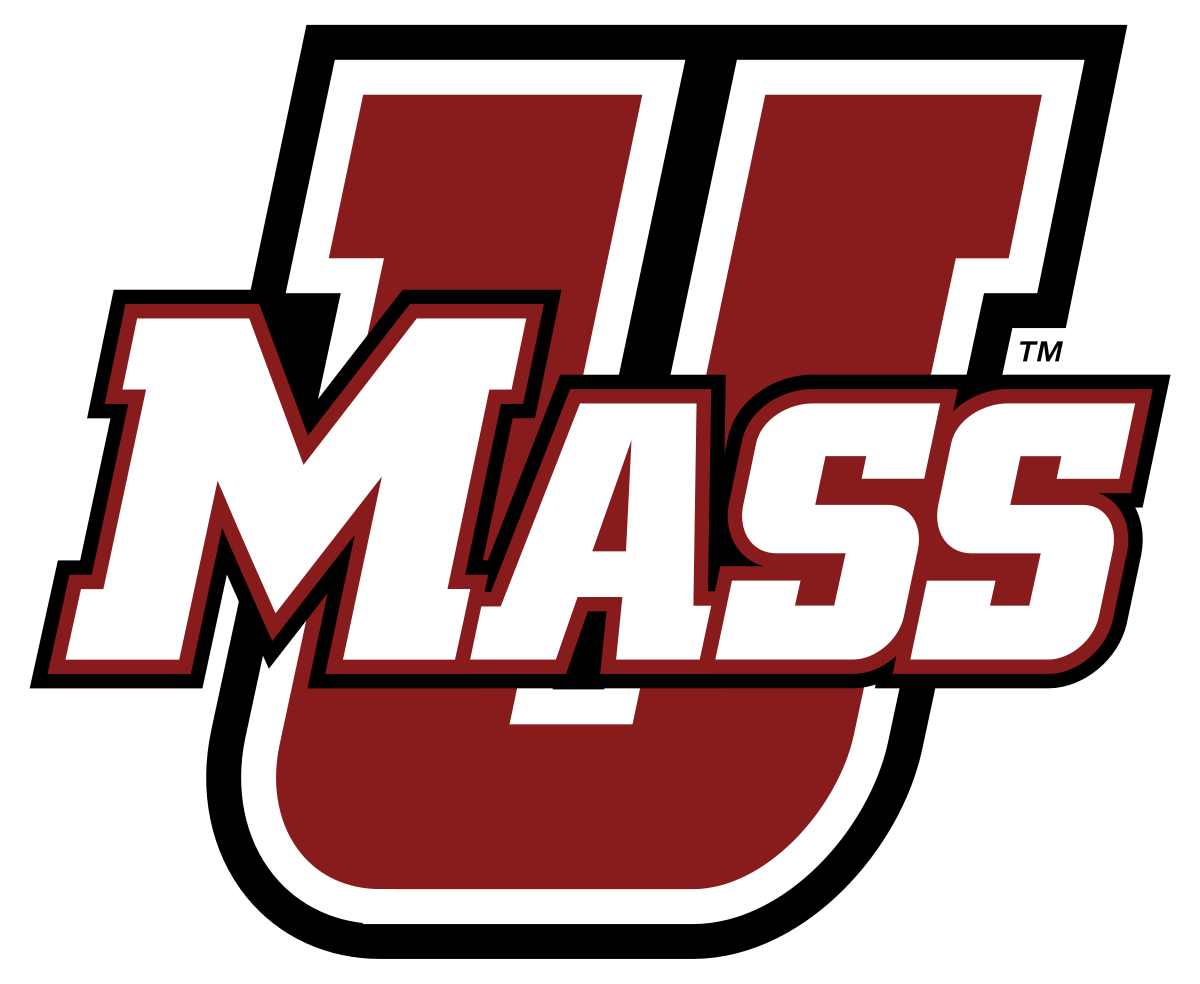} \includegraphics[height=1em]{figures/pangram.png}} \quad
\textbf{Bradley Emi\textsuperscript{\includegraphics[height=1em]{figures/pangram.png}} \quad
Max Spero\textsuperscript{\includegraphics[height=1em]{figures/pangram.png}} \quad
Mohit Iyyer\textsuperscript{\includegraphics[height=1.2em]{figures/umd.png} \includegraphics[height=1em]{figures/umass.png}}}\protect\\[0.75ex]
\textsuperscript{\includegraphics[height=1.2em]{figures/umd.png}}University of Maryland, College Park \quad
\textsuperscript{\includegraphics[height=1em]{figures/SFU_block_colour_rgb_1000px.png}}Simon Fraser University \quad
\textsuperscript{\includegraphics[height=1em]{figures/umass.png}}UMass Amherst\quad \\
\textsuperscript{\includegraphics[height=1em]{figures/pangram.png}}Pangram Labs\protect\\
 \texttt{\{jennarus,miyyer\}@umd.edu}, \texttt{karpinsk@sfu.ca}\\
 \href{https://ainewsaudit.github.io}{{\texttt{ainewsaudit.github.io}}}
}

\usepackage[utf8]{inputenc}
\DeclareUnicodeCharacter{2212}{-}
\DeclareUnicodeCharacter{0301}{\'{}}

\begin{document}
\maketitle 
\begin{abstract}
AI is rapidly transforming journalism, but the extent of its use in published newspaper articles remains unclear. We address this gap by auditing a large-scale dataset of 186K articles from online editions of 1.5K American newspapers published in the summer of 2025. Using Pangram, a state-of-the-art AI detector, we discover that approximately 9\% of newly-published articles are detected as either partially or fully AI-generated. This AI use is unevenly distributed, appearing more frequently in smaller, local outlets, in specific topics such as weather and technology, and within certain ownership groups. We also analyze 45K opinion pieces from \emph{Washington Post}, \emph{New York Times}, and  \emph{Wall Street Journal}, finding that they are 6.4 times more likely to contain AI-generated content than news articles from the same publications, with many AI-flagged op-eds authored by prominent public figures. Despite this prevalence, we find that AI use is rarely disclosed: a manual audit of 200 AI-flagged articles, 96.5\% of authors and 94.0\% of publishers did not disclose AI use. A factuality analysis shows articles flagged as AI-generated are 8.2 times more likely to contain hallucinated claims than human-labeled news. Overall, our audit highlights the immediate need for greater transparency and updated editorial standards regarding the use of AI in journalism to maintain public trust.
\end{abstract}

\definecolor{GrayBack}{RGB}{248,248,248}
\definecolor{GrayFrame}{RGB}{64,64,64}
\definecolor{GrayTitle}{RGB}{40,40,40}
\definecolor{GrayText}{RGB}{40,40,40}

\newtcolorbox{AlertBoxMinimal}[1][]{
  enhanced,
  colback=GrayBack,
  colframe=GrayFrame,
  coltitle=GrayTitle,
  coltext=GrayText,
  boxrule=0.8pt,
  arc=2pt,
  boxsep=2pt,
  left=5pt,right=5pt,top=5pt,bottom=5pt, 
  title={\textsc{\textcolor{white}{In this paper...}}},
  fonttitle=\bfseries\scshape,
  before skip=8pt, after skip=8pt,
  #1
}

\newtcolorbox{AlertBoxMaximal}[1][]{
  enhanced,
  colback=GrayBack,
  colframe=GrayFrame,
  coltitle=GrayTitle,
  coltext=GrayText,
  boxrule=0.8pt,
  arc=2pt,
  boxsep=2pt,
  left=5pt,right=5pt,top=5pt,bottom=5pt, 
  title={\textsc{\textcolor{white}{In this paper...}}},
  fonttitle=\bfseries\scshape,
  before skip=8pt, after skip=8pt,
  #1
}

\newtcolorbox{AlertBoxMaximal2}[1][]{
  enhanced,
  colback=GrayBack,
  colframe=GrayFrame,
  coltitle=GrayTitle,
  coltext=GrayText,
  boxrule=0.8pt,
  arc=2pt,
  boxsep=2pt,
  left=5pt,right=5pt,top=5pt,bottom=5pt, 
  title={\textsc{\textcolor{white}{Disclaimer}}},
  fonttitle=\bfseries\scshape,
  before skip=8pt, after skip=8pt,
  #1
}

\section{Introduction}
\label{sec:intro}

How much AI-generated content is being published in newspapers across America? To answer this question, 
we conduct a large-scale audit of recently-published articles using Pangram~\citep{emi2024technicalreportpangramaigenerated}, a high-precision AI detector that has previously been used to audit consumer reviews \cite{cavazos2024highcost}, research papers \cite{evanko2025quantifying}, and Medium articles \cite{knibbs2024aiSlop}. Our analysis reveals that \textbf{$\sim$9\%} of newly-published U.S. newspaper articles are detected as either partially or fully AI-generated.

\begin{figure}[t]
  \centering
    \begin{AlertBoxMaximal2}
 All findings in this paper rely on the outputs of an automated AI detector and should be interpreted accordingly. We do not assert definitive authorship attributions for any article. We do not draw conclusions about the intent, conduct, or practices of individual journalists, outlets, or companies. Results should not be interpreted as rankings, qualitative judgments, or accusations of wrongdoing. See \S\ref{sec:ethical} for further discussion.
    \end{AlertBoxMaximal2}
\end{figure}

\paragraph{Why does this matter?} 
Our audit is largely motivated by concerns of \emph{transparency} and \emph{factuality}. We do not claim that all AI use is inherently harmful; in fact, limited applications like grammar / style checking and template-driven article creation (e.g., weather reports) can improve article quality and accessibility~\citep{medill2024ai_local,radcliffe2025journalism}. However, LLMs often hallucinate \cite{maynez-etal-2020-faithfulness, ji2023surveyofhallucination, su-etal-2024-adapting} and inherit social biases from their training data~\cite{gallegos-etal-2024-bias, hu2025identitybias}; thus, public opinion is highly sensitive to \emph{undisclosed} AI use.\footnote{Recent studies from Pew Research show that (1) 49\% of Americans who get news directly from AI assistants report encountering inaccurate information~\citep{pew2025-chatbots-news}; (2) 56\% of Americans would feel less confident about a news article if they knew an AI wrote it~\citep{pew2025-ai-reaction}; and (3) 76\% believe it is extremely important for them to know if the text they are reading is AI-generated~\citep{pew2025-ai-views}.} 
We manually analyze 200 articles flagged for AI use by Pangram and find that only \textbf{7} of them disclose AI use, while only \textbf{12} of the newspapers have any public policies on AI use, leaving readers largely unable to determine the role AI plays in article authorship.

\paragraph{Audit design:}
We collect and audit two large-scale datasets of American newspapers:
\begin{itemize}
    \item[\faNewspaper] \maindata\ contains \textbf{186K} articles published online by \textbf{1.5K} local and national newspapers from June to September 2025.
    \item[\faComment] \opinionsdata contains \textbf{45K} \emph{opinion} articles published by the New York Times, Washington Post, and Wall Street Journal between August 2022 and September 2025.
\end{itemize}

We feed each  article in both datasets through Pangram's API to obtain both an AI likelihood (from 0-100\%) and a categorical label in \{\labelhuman, \labelmixed, \labelai \}. Pangram has a reported false positive rate (FPR) of $\sim$0.001\% on news text~\citep{emi2025falsepositives} and \citet{jabarian2025artificial} verified Pangram's high accuracy, finding an FPR of 0.08\%, substantially below the rates of AI use we observe.  
Importantly, for articles labeled \labelmixed, we cannot infer the role that AI played in the authorship process: we only know that some parts of a \labelmixed\ article are classified as human-written while other parts are classified as AI-generated. 

\paragraph{AI use in published articles is increasingly common.}
In our \maindata\ dataset, 9.1\% of articles are labeled by Pangram as either \labelai\ or \labelmixed, and disclosure of AI use is rare in our manual sample. Digging deeper, we observe that AI usage is unevenly distributed: it is much higher in smaller local outlets than nationally-circulated papers;
it occurs more frequently in topics such as weather, science / technology, and health (\autoref{fig:topic_ai_likelihood}); it varies across ownership group, with Boone Newsmedia and Advance Publications among the heaviest AI adopters (\autoref{fig:owner_topic_heatmap}); and it is higher in languages other than English (\autoref{fig:eng_vs_other_langs}).\footnote{Example articles linked within this paper are solely for illustrative purposes. Their inclusion does not attribute any intent or misconduct on part of any reporters: Pangram has a small but non-zero FPR, and we also emphasize that it is infeasible to tease apart exactly how AI is used in \labelmixed\ authorship cases.} 

\begin{figure}[t]
  \centering
    \begin{AlertBoxMinimal}
    \colorbox{gray!40}{\textsc{\textcolor{black}{\textbf{\small AI USE}}}}: We use the term \emph{AI use} throughout this paper to denote articles labeled by Pangram as either {\small \labelmixed} or {\small \labelai}.\vspace{0.2cm}\\
    \colorbox{gray!40}{\textsc{\textcolor{black}{\textbf{\small EXAMPLES}}}}: Each article discussed in this paper is associated with a 
    \pangram{https://www.pangram.com/history/9c2828da-4231-45da-a469-20851d6158f5/} icon that links to an AI prediction dashboard and a
    \news{https://ainewsaudit.github.io/} icon that links to the original news article.
    \end{AlertBoxMinimal}
\end{figure}

\paragraph{Undisclosed AI use compounds risks of factual hallucinations.}
While AI use is becoming more common, we find it is rarely disclosed and often factually unreliable. When investigating whether AI use is disclosed, only 4\% of articles were transparent about writing articles with the help of AI, leaving readers in the dark about the true sources of information. At the same time, we find that 41\% of articles labeled as AI-generated contain hallucinations, showing that AI-generated information is often inaccurate and not corrected by human revision.

\paragraph{AI use is concentrated in opinion articles at top newspapers.}
Opinion articles published at the NYT, WaPo, and WSJ are \textbf{6.4} times more likely to be flagged for AI use than contemporaneous news articles from the same three newspapers (4.5\% vs. 0.7\%).
Many opinion articles flagged for AI use are written by prominent guest contributors, including Nobel prize winners, US Senators and Governors, Pulitzer Prize-winning journalists, and CEOs (see \autoref{tab:opinion-ai-controversial} for examples). Analysis of both \opinionsdata\ and \reporterdata\ show AI usage rising over time, with reporters in the latter dataset increasing their AI use from $\sim0$\% prior to 2023 to over 40\% in 2025 on average (\autoref{fig:ai_reporters}). 

\paragraph{Contributions:}
We release our datasets,\footnote{We release links to the articles involved in our study (not full texts).} 
analysis code, and an interactive dashboard to facilitate further exploration of AI use in newspapers.
\footnote{\href{https://ainewsaudit.github.io}{\texttt{ainewsaudit.github.io}}} 
We also commit to periodically updating our dashboard with new articles and annotations (i.e., disclosure audits) to measure future changes to AI adoption.

\section{Collecting newspaper articles}
\label{sec:data_methodology}

We collect two datasets of published newspaper articles (\autoref{tab:dataset-meta}):
\maindata and \opinionsdata. \footnote{We collect a \textit{third} dataset, \reporterdata, which is detailed and analyzed in \S\ref{app:ai_reporters}.}
% \maindata, \opinionsdata, and \reporterdata.
The datasets vary significantly in terms of coverage, diversity, and publication date, which allows us to analyze differences in AI use across local vs. national papers, staff reporters vs. guest contributors, and articles written pre- vs. post-ChatGPT. 
This section outlines our dataset creation process, in which the full text and metadata for each article is paired with a label indicating whether the article was generated via AI.

\begin{table}[t]
\centering
\resizebox{\columnwidth}{!}{%
\begin{tabular}{lrr}
\toprule
 & \faNewspaper\ \textbf{\maindata} & \faComment\ \textbf{\opinionsdata} \\
\midrule
\multicolumn{3}{l}{\textbf{Temporal coverage}} \\
Years & 2025 & 2022--2025 \\
\midrule
\multicolumn{3}{l}{\textbf{Dataset statistics}} \\
\# Articles            & 186{,}507 & 44{,}803 \\
\# Authors             & 34{,}608  & 9{,}863 \\
\# Newspapers          & 1{,}528   & 3 \\
Avg. tokens            & 787.4     & 1078.4 \\
\midrule
\multicolumn{3}{l}{\textbf{AI use statistics}} \\
\labelhuman           & 90.85\% & 99.04\% \\
\labelmixed           & 3.98\%  & 0.85\% \\
\labelai              & 5.24\%  & 0.11\% \\
\bottomrule
\end{tabular}%
}
\caption{Dataset and AI use statistics for \maindata\ and \opinionsdata. The token counts are reported as per \href{https://github.com/openai/tiktoken}{\texttt{tiktoken}} tokenizer (\texttt{o200k\_base}).}
\label{tab:dataset-meta}
\end{table}

\subsection{\maindata} To examine AI use in present-day newspaper journalism, we form the \maindata\ dataset by collecting 186,507 articles\footnote{Each instance in our datasets includes the full article text along with title, author, publication date, newspaper, and URL.} published online by 1,528 unique newspapers between June 15th, 2025 and September 15th, 2025, accessed via url RSS feeds.\footnote{See \S\ref{app:dataset} for additional collection details.} 
 
 \subsection{\opinionsdata} 
 While \maindata\ contains a broad sample across many different American newspapers, topics, and journalists, another object of our study is AI use in \emph{opinion} articles written by prominent people in highly-reputable newspapers. To facilitate this, we also collect a dataset of opinions articles published by The New York Times, Wall Street Journal, and Washington Post between August 2022 and September 2025. The full text and metadata of these articles were accessed via \emph{ProQuest Recent Newspapers}. In total, we collected 44,803 articles during this time period: 16,964 from WSJ, 15,977 from WP, and 11,862 from NYT. 

\subsection{Labeling the datasets}

For each of the 251,442 total articles across all datasets, we use Pangram (v2.0) to obtain an AI detection label and score.\footnote{All Pangram links display v3.2 predictions, and had consistent labels with Pangram v2.0.} We also classify each article into one of 19 topics, and we use an existing database to match about half of the newspapers in our \maindata with print circulation statistics. 

\paragraph{Detecting AI use:} 

\citet{emi2024technicalreportpangramaigenerated} introduce Pangram; a robust AI-generated text detection tool. On news articles, it achieves
a false positive rate of 0.001\% \cite{emi2025falsepositives}, consistent with other studies that have also reported low FPR \cite{russell2025peoplewho, jabarian2025artificial, dugan-etal-2025-genai}.
Using Pangram's inference API\footnote{API documentation available \href{https://pangram.readthedocs.io/en/latest/api/rest.html}{here}.}, we collect (1) the likelihood (from 0-100\%) that a text is AI-generated and (2) a text label that is one of \labelhuman, \labelmixed, or \labelai.\footnote{We simplify the fine-grained labels produced by the Pangram API in the following way: \labelhuman{} $=$ \{\textit{Human}, \textit{Unlikely AI}\}; \labelmixed{} $=$ \{\textit{Mixed}, \textit{Possibly AI}, \textit{Likely AI}\}; \labelai{} $=$ \{\textit{Highly Likely AI}, \textit{AI}\}. See \S\ref{app:ai_detections} for details on how labels are combined.}  
Pangram predicts \labelmixed\ when there is a high confidence of both AI and human writing present in the document; specifically, where some segments are predicted as AI, and some segments are predicted as human. 
While prior studies have validated Pangram's high accuracy, we also experiment with another commercial detector, GPTZero~\cite{tian2023gptzero}, and observe a  high cross-detector agreement of $88.2\%$ (Cohen's $\kappa = 0.764$).\footnote{This experiment was conducted on a balanced binary held-out set of 1K news articles, 500 of which are marked by Pangram as \labelhuman\ and 500 as \labelai. Agreement on the human subset is 98.4\% while agreement on the AI subset is 78.4\%, and discrepancies on the latter label are likely due to each detector's differing treatment of \labelmixed\ text (see \S\ref{app:ai_detections} for more).} 

 \begin{figure*}[t]
        \centering
        \includegraphics[width=\linewidth]{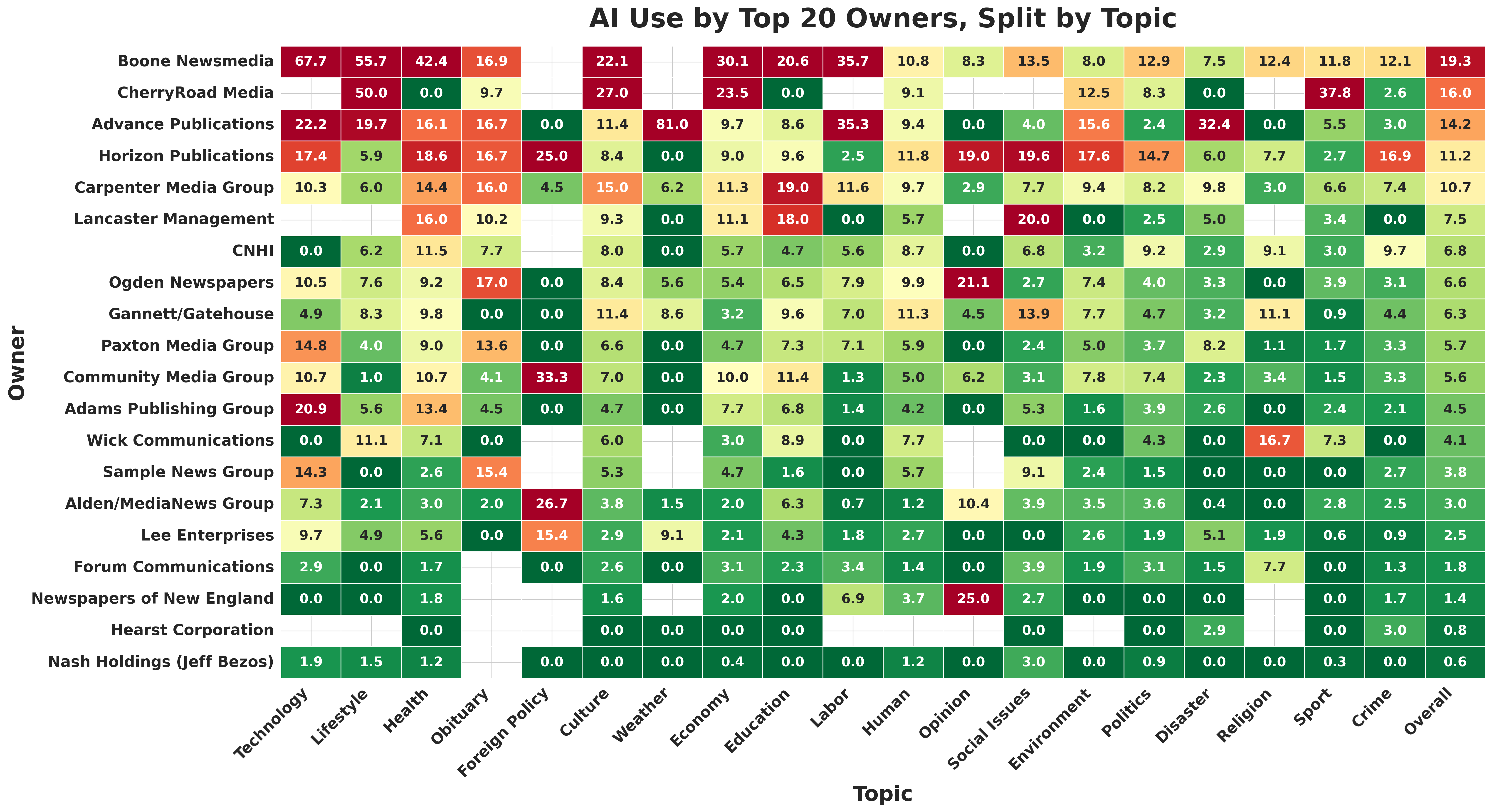}
        \caption{Heatmap of AI use by publication owner and article topic in \maindata. Some owners disclose AI use for specific content, such as \href{https://www.advance.com/}{Advance Publications} for weather reports, but the biggest adopters, such as \href{https://boonenewsmedia.com/ |}{Boone Newsmedia}, use AI broadly across many  topics.  Note that only 87\% of \maindata has ownership information, and only topics that have at least 5 articles per owner are visualized in the heatmap.}
        \label{fig:owner_topic_heatmap}
\end{figure*}

\paragraph{Topic classification:}

To analyze AI use across article topics, we further augment our datasets with topic labels for each article using the International Press Telecommunications Council \emph{Media Topics} taxonomy, which comprises 17 top-level topics~\cite{iptc2025mediatopics}. We prompt \textsc{Qwen3-8B}~\cite{qwen3-2025} in a zero-shot setting to assign a topic (full prompt and details in \S\ref{app:dataset}).\footnote{In addition to the 17 IPTC topics, we include two auxiliary categories—\emph{Other} and \emph{Obituary}—for items that fall outside the taxonomy or are obituaries.} To assess the reliability of these topic labels, two of the authors independently reclassified a random subset of 100 model outputs,\footnote{Inter-annotator agreement between the two human raters was 87\% (Cohen’s~$\kappa=0.85$).} with model–human agreement averaging 77\%, indicating moderately strong alignment between the classifier and human judgments.

\paragraph{Linking newspapers to circulation and ownership information:}
We link newspapers to print circulation data from the U.S. News Deserts Database \cite{usnewsdeserts2020} and ownership information from the Local News Initiative \cite{medill2024ai_local}. Because circulation figures are based on historical AAM audits and self-reports (primarily from 2019), we treat them as a proxy for \emph{print scale} rather than current total audience.\footnote{See \S\ref{app:circulation_definition} for details and limitations. 54.6\% of the articles and 49.7\% of newspapers in \maindata were matched to publications in the News Desert Database.} 40.3\% of articles come from small publications with daily circulation between 1-7K and another 36.3\% comes from publications with a circulation of 10-50K. Only 20.1\% of articles come from very large nationally-circulated publications with average circulations over 50K.\footnote{Circulation distribution is depicted in \autoref{fig:circulation_distribution}.}

\section{Analyzing AI use in newspapers}
\label{sec:analysis}

How much of recent news contains AI-generated content, and when and how do journalists use AI? Using our \maindata\ dataset, we investigate \textit{where} AI-generated content appears across circulation, ownership, language, and article topic. We also explore \textit{how} it is used, looking at differences between \labelmixed\ and \labelai\ articles, factual details, and disclosure. 

\subsection{How often is AI used in American newspapers?}

We find that \textbf{9.1\%} of the 186K  articles in \maindata\ are  labeled by Pangram as either \labelai\ (5.2\%) or \labelmixed\ (3.9\%), while the remaining 90.9\% of articles are classified as \labelhuman. The rest of this section goes beyond these aggregate numbers to examine AI use as a function of different fine-grained aspects like topic and ownership.

\paragraph{AI use is higher in local newspapers.}
Local outlets, often operating in "news deserts," places with little to no access to credible reporting due to limited resources \cite{medill2024state}, rely on AI  
more than national newspapers.
Only 1.7\% of articles at papers with circulation $>$100K are labeled as \labelai\ or \labelmixed, versus 9.3\% at papers below 100K (see \autoref{fig:circulation_ai_likelihood}),\footnote{At the article level, this difference is highly significant ($\chi^2(1) = 1175.6$, $p < 10^{-250}$). At the newspaper level, smaller outlets averaged 8.5\% AI content vs. 5.0\% among very large outlets (Welch's $t(\approx 23) = 2.24$, $p = 0.032$, $d = 0.22$).}
suggesting that large national newspapers enforce stricter editorial constraints on automation than local papers. 
AI use also varies geographically,
peaking in the
mid-Atlantic and southern US, Maryland (16.5\%), Tennessee (13.6\%), Alabama (13.9\%), and lowest in the Northeast, including New Hampshire (2.9\%) and Massachusetts (3.4\%)
(\autoref{fig:state_ai_heatmap}).

\paragraph{AI use varies with topic.}  
Prior work has shown that factual, data-heavy content (e.g., reports about weather, finance, or sports) is particularly amenable to automation \cite{medill2024ai_local}.
% \footnote{As journalist Tom Rosenstiel notes, “If it’s the weather report, who the hell cares? If it’s a story about Latino culture, that could be a problem.” \news{https://localnewsinitiative.northwestern.edu/assets/research/medill_ai_local_news_report_2024.pdf}} 
Consistent with this, weather articles in our dataset exhibit the highest average AI likelihood (27.7\%), as shown in \autoref{fig:topic_ai_likelihood}. However, we also observe high AI use in other topics, such as  science and technology (16.1\%) and health (11.7\%), while content on more sensitive issues such as conflict and war (4.3\%), crime, law, and justice (5.2\%), and religion (5.3\%) exhibit lower rates. 

\paragraph{AI use varies with ownership.} 
Many newspapers in \maindata\  share 
ownership: for example, Advance Publications owns
widely-read outlets like \url{pennlive.com}, \url{cleveland.com}, and \url{al.com} and often syndicates articles among them. 
While some ownership groups view AI use as a reputational risk, others emphasize cost reduction and efficiency gains \cite{medill2024ai_local}.
Boone News Media and CherryRoad Media both have detected AI use rates over 15\%, while Nash Holdings and Hearst Corporation have AI use rates under 1\% (\autoref{fig:owner_topic_heatmap}). We note the irony of several prominent media groups suing AI companies over training language models on their content (e.g., \emph{Advance Local Media v. Cohere})\footnote{Advance Local Media LLC v. Cohere Inc., No. 1:25-cv-01305, S.D.N.Y. filed Feb. 13, 2025.} even as they churn out LLM-generated articles. 

\paragraph{AI use varies by topic across ownership groups.}  
\autoref{fig:owner_topic_heatmap} shows how different ownership groups use AI. Advance Publications, for instance, relies heavily on AI for weather reporting (81\% AI Use). Boone Media exhibits the broadest adoption, with especially high rates in science and technology (67.7\%) as well as lifestyle and leisure (55.7\%). Other chains show narrower topical patterns, such as CherryRoad Media in sports (37.8\%) and Lee Enterprises in foreign policy coverage (15.4\%). These results show that AI integration is vastly uneven across both topic and ownership groups.  

 \begin{figure}[tb]
        \centering
        \includegraphics[width=\linewidth]{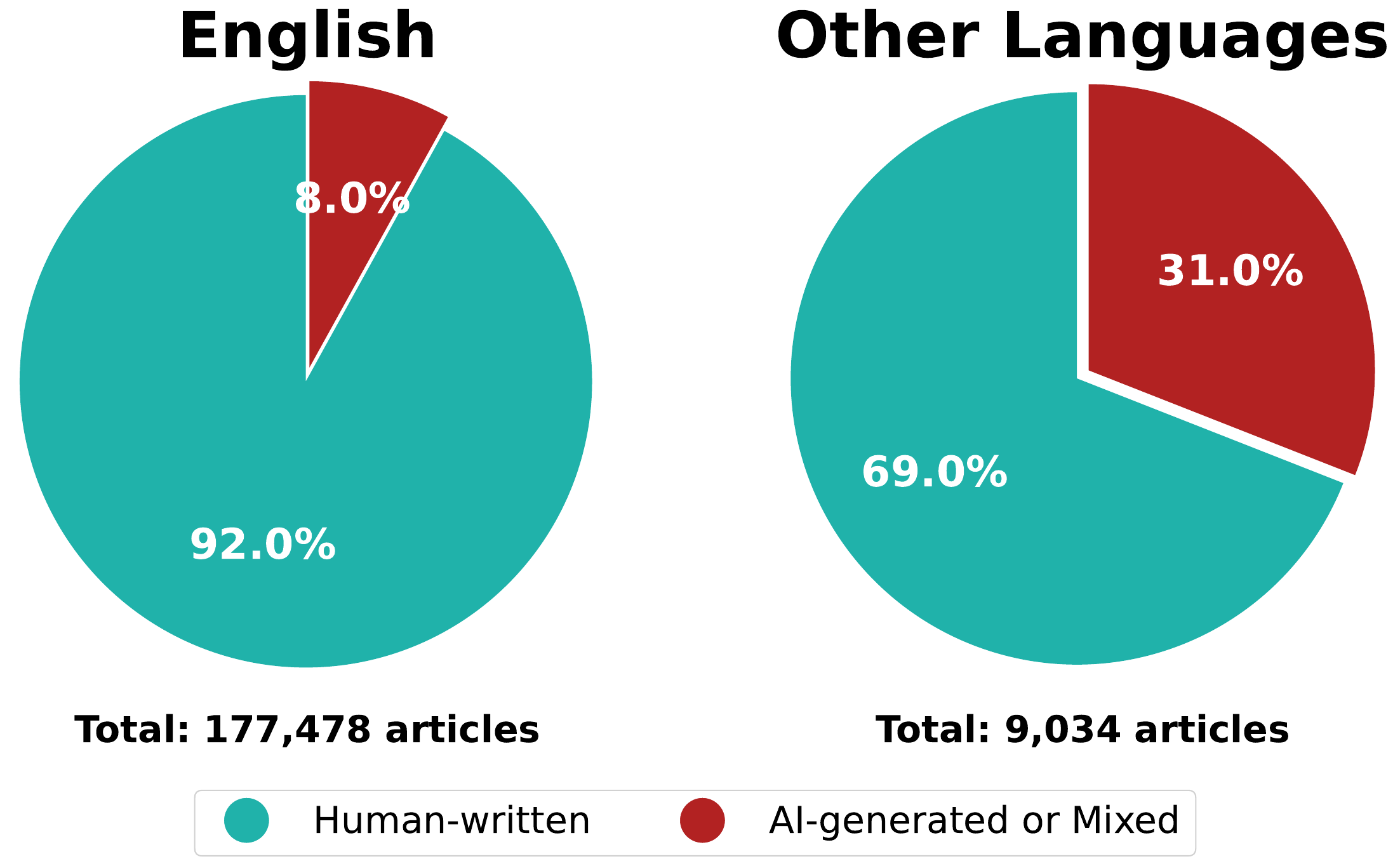}
        \caption{AI use in \maindata\ is more frequent in languages other than English. The most prominent such languages include Spanish, Portuguese, Vietnamese, French, and Polish.}
        \label{fig:eng_vs_other_langs}
\end{figure}

\paragraph{AI use is higher in languages other than English.}
AI-generated content is more prevalent in non-English news articles.\footnote{Pangram FPR's range from 0.0\% to 0.10\% across the languages in this study \cite{pangram2024multilingual}.}
As shown in \autoref{fig:eng_vs_other_langs}, only 8.0\% of English-language articles in \maindata are classified as \labelai\ or \labelmixed; this share rises to 31.0\% for articles in other languages.\footnote{Machine-translated articles may be misclassified, but our small-scale verification suggests human-written articles are still correctly identified after translation.}
Most non-English AI use ($\approx 80\%$) comes from U.S.-based Spanish-language reporting (7.2K articles), \footnote{Other major languages include: Portuguese (468), Vietnamese (403), French (343), and Polish (314).} suggesting that local bilingual outlets rely more heavily on translation and other automation.
High non-English AI use is observed across many states (\autoref{fig:non_english_state_heatmap}).

\begin{table*}[t]
\centering
\small
\setlength{\tabcolsep}{3pt}%
\resizebox{\textwidth}{!}{%
\begin{tabular}{p{0.25\textwidth}p{0.07\textwidth}p{0.25\textwidth}p{0.48\textwidth}}
\toprule
\multicolumn{1}{c}{\bf\textsc{Policy Category}} & \multicolumn{1}{c}{\bf\textsc{Freq}} & \multicolumn{1}{c}{\bf\textsc{Example Publications / Owners}} & \multicolumn{1}{c}{\bf\textsc{Illustrative Disclosure Text}} \\
\midrule
\textsc{\allowedicon\;AI Allowed} & 12 & \href{https://cm.sentinel-standard.com/ethical-conduct/}{\textit{Ionia Sentinel-Standard}}, \href{https://amsterdamnews.com/editorialpolicy/}{\textit{Amsterdam News}}, \href{https://www.pennlive.com/weather/2025/08/air-quality-alert-continues-as-hazy-conditions-persist-in-central-pa.html}{\textit{Penn Live}} &
\textit{``If AI-assisted content is approved for publication, journalists must disclose the use of AI and its limitations to their audience. AI-generated content must be verified for accuracy and factuality before being used in reporting.''} \\
\midrule
\textsc{\prohibitedicon\;AI Prohibited} & 2 & \href{https://nypost.com/editorial-standards/}{\textit{New York Post}}, \href{https://www.michigandaily.com/the-bylaws-of-the-michigan-daily-january-17-2024/}{\textit{Michigan Daily}} & 
\textit{``The use of generative artificial intelligence for content production (including written, visual and auditory content) is unacceptable in all circumstances. Any staffer found to have used generative AI to produce content for The Daily can be fired by their section editor or the Editor in Chief.''} \\
\midrule
\textsc{\shrugicon\;No Public Policy} & 186 & \textit{Daily Register, Hudson Reporter, LA Opinion} & 
\textit{No disclosure found on website.} \\
\bottomrule
\end{tabular}}%
\caption{AI disclosure policies among 200 sampled U.S. news outlets.}
\label{tab:ai_disclosures}
\end{table*}

\paragraph{Exploring the role of machine translation in AI detection}
To understand if the higher rate of detected AI use in non-English languages is an artifact of machine translation (MT), we translate a sample of 300 English articles into 12 languages and re-evaluate them using the same detection pipeline.\footnote{We sample 100 \labelai, 100 \labelmixed, and 100 \labelhuman articles. Articles are translated using GPT-4.1. We note that results may differ with other MT methods or models; see \S\ref{app:translation_details} for details.} Translated articles receive lower AI-likelihood scores than their English counterparts, with an average decrease of 13\%. Under the binary AI use versus human, translated texts agree with English in 83.2\% of cases, and disagreements are asymmetric: in our sample, translation is far more likely to suppress AI use predictions than to introduce them.\footnote{McNemar’s exact test, FDR-corrected $p<0.001$ for all but one language (Japanese).}  The magnitude of this effect varies by language: Vietnamese translations show the largest score decrease ($\Delta = -0.21$) and lowest agreement (74.7\%), while Portuguese and Spanish retain high agreement ($\geq$88\%) with minimal score shifts (see \autoref{tab:translation_binary}). While these results suggest that MT is unlikely to be the primary driver of the higher AI detection rates in non-English content, especially given the observed $\sim$31\% AI use rate, this analysis does not account for the influence of human translation or alternative MT methods. Multilingual AI detection is an evolving field; factors like language-specific writing conventions, varied editorial workflows, and regional AI tool availability likely contribute to these discrepancies.

\subsection{Characterizing AI use in American newspapers}

Our analysis establishes that (1) many published newspaper articles
today
are written partially or entirely by AI, and (2) AI use varies across factors like ownership, topic, and language. We investigate important 
questions about authorship, transparency, and public trust, revealing that journalists often use AI in conjunction with their own writing rather than as a full replacement; that prestigious outlets rely more on \labelmixed\ authorship; articles with AI use are more likely to contain hallucinations; and that disclosure of AI involvement is strikingly rare. These findings show not only how AI is changing journalistic practices, but also why it matters whether readers are informed about its role.

\paragraph{Many articles have mixed authorship.}  
While \labelmixed\ authorship articles include at least some human writing, it is difficult to tease apart the contributions of the reporter compared to AI~\cite{thai2025editlensquantifyingextentai}. Mixed authorship is common: of the 17{,}059 articles we detect as using AI, 42.7\% are predicted as \labelmixed, while 57.3\% are classified as \labelai. At the author level, 1{,}453 out of 34{,}608 writers produce at least some AI content. The majority (54.8\%) primarily publish mixed articles, while 36.1\% rely mostly on AI-generated text. These findings resonate with survey evidence from \citet{radcliffe2025journalism}, who report that over half of journalists use AI to edit their work, while only about a third employ it to generate text directly.

\paragraph{AI use is largely undisclosed.}
Disclosure of AI use (e.g., exactly how and where AI was used in the construction of the article) is important to maintain audience trust. Readers might be okay with small AI edits for style in articles labeled as \labelmixed\, but they require proper disclosure to make these judgments.cUnfortunately, in a sample of 200 AI-flagged articles from unique newspapers in \maindata, 96.5\% of authors and 94.0\% of publishers did not disclose AI use (\autoref{tab:ai_disclosures}).\footnote{95\% Wilson score CIs: 193/200 (96.5\%; [93.0\%, 98.3\%]) and 188/200 (94.0\%; [89.8\%, 96.5\%]).}

\begin{table*}[!htb]
\centering
\scriptsize
\scalebox{1}{
\begin{tabular}{
    p{0.08\textwidth}  
    p{0.12\textwidth}  
    p{0.34\textwidth}  
    p{0.34\textwidth}   
}

\toprule
\multicolumn{1}{c}{\bf\textsc{Newspaper}} & 
\multicolumn{1}{c}{\bf\textsc{Prediction}} & 
\multicolumn{1}{c}{\bf\textsc{Article Excerpt}} &
\multicolumn{1}{c}{\bf\textsc{Observations}} \\
\midrule

ARGONAUT 
& 
\labelai  & \textit{``...The forms of commerce have drastically changed, with constant evolution to meet consumer needs and \aiword{market dynamics}. From brick-and-mortar stores, e-commerce, and omnichannel strategies to pop-up shops and outdoor commerce, businesses need to remain adaptable \aiword{to thrive in} this ever-changing commercial \aiword{landscape}...''} \pangram{https://www.pangram.com/history/a00f5712-2059-4525-b886-3a120b407954} \news{https://www.argonautnewspaper.com/exploring-different-forms-of-commerce-from-brick-and-mortar-stores-to-online-marketplaces/} &  Fully AI-generated text with a very generic conclusion. No details about the author exist online, and when we further investigated the site, no details could be found about any of its staff.  
% argonaut v2: https://www.pangram.com/history/995a302c-2c9e-4adc-99f6-8dc91069622e/
% argonaut v3.2: SAME (AI) https://www.pangram.com/history/a00f5712-2059-4525-b886-3a120b407954
% GPTZero (2026-03-30-base): HUMAN_ONLY (ai=0.427, mixed=0.084, human=0.489) - low confidence; text from web scrape, not local dataset
\\

\midrule

CALEXICO CHRONICLE
& \labelhuman  & \textit{``...This grant represents an important \aiword{step forward in our} efforts to create healthier, \aiword{more sustainable} learning environments for our students. By increasing shade and greenery across our campuses, we’\aiword{re not} only improving outdoor comfort and air quality, but also setting an example of \aiword{environmental responsibility} for our students and community...''} \pangram{https://www.pangram.com/history/ffcd75ac-1cab-442f-903d-44cb97c174b2} \news{https://calexicochronicle.com/2025/07/21/iid-awards-500k-in-grants-to-support-public-green-spaces/}
& While this article is detected as human-written, it includes an AI-generated quote that was likely provided to the reporter \pangram{https://www.pangram.com/history/bc576edf-ec53-4613-ae53-97659fa2fe48}. Reporters who write their own articles may not be aware that the people they quote in their articles used AI to create their response.\\

% calexico v2: https://www.pangram.com/history/7fc6e863-fef7-44e8-9d6e-f9afd40e56a9/
% calexico v3.2: SAME, human, https://www.pangram.com/history/ffcd75ac-1cab-442f-903d-44cb97c174b2
% GPTZero (2026-03-30-base): HUMAN_ONLY (ai=0.000, mixed=0.000, human=1.000) - AGREES with Pangram

% quote v2: https://www.pangram.com/history/33d6453d-a50e-4b96-8afb-dbebd5b5356c/
% quote v3.2: SAME (AI), https://www.pangram.com/history/bc576edf-ec53-4613-ae53-97659fa2fe48

\midrule

WASHINGTON POST 
& \labelmixed  &
\textit{``...Finally, focus on who and where else you can \aiword{seek support} from. Is there even one family member or community member you can turn to and tap in for support and allyship as \aiword{you navigate} these familial struggles? }
\pangram{https://www.pangram.com/history/dbacd1d3-975b-494e-82e2-f82e8b496b5b} \news{https://www.washingtonpost.com/advice/2025/07/31/ask-sahaj-family-lonely-needs-misunderstood/} & AI use often occurs even in high circulation papers like \emph{Washington Post}. In this advice column, a person writes in feeling lonely, only to receive partially AI-generated advice.  
% wp v2: https://www.pangram.com/history/116fd457-fbb1-4af4-83d4-28d1bc8e6781/
% wp v3.2: same, mixed: https://www.pangram.com/history/dbacd1d3-975b-494e-82e2-f82e8b496b5b
% GPTZero: human
\\
\bottomrule
\end{tabular}}
\caption{Notable cases of AI use in the \maindata\ dataset. Words and phrases identified as indicative of AI use by Pangram are highlighted in red. AI use takes many forms, from completely made-up news sites to AI responses to advice columns (e.g., \emph{Dear Annie}), legitimate articles that happen to quote AI-generated text from other sources, and highly-templated topics like weather  and sports reports. 
% Additional cases can be found in \autoref{tab:notable_cases-extra}.
}
\label{tab:notable_cases}
\end{table*}

\paragraph{Notable individual cases of AI use.}
We identify several unique cases of AI-generated writing in \maindata, snippets of which are shown in \autoref{tab:notable_cases}. One outlet, \href{https://www.argonautnewspaper.com/about-argonaut/}{Argonaut}, turned out to be an entirely AI-generated newspaper with AI reporter personas who ``write'' exclusively articles labeled as \labelai\ (e.g., 
\pangram{https://www.pangram.com/history/96294f6e-82c9-4398-a74a-5784b6820608}
\news{https://www.argonautnewspaper.com/exploring-different-forms-of-commerce-from-brick-and-mortar-stores-to-online-marketplaces/}). 
% argonaut v2: https://www.pangram.com/history/995a302c-2c9e-4adc-99f6-8dc91069622e/
% argonaut v3.2: SAME (AI), https://www.pangram.com/history/96294f6e-82c9-4398-a74a-5784b6820608
% GPTZero (2026-03-30-base): HUMAN_ONLY (ai=0.427, mixed=0.084, human=0.489) - low confidence
Another reporter appears to have revisited and republished their older work, producing updated AI-assisted versions of previously \labelhuman\ stories (e.g., \textcolor{purple}{edited (2025):} \pangram{https://www.pangram.com/history/5fd93712-5085-4a8b-b3b5-d9c81b3ec2c4}, \news{https://snyderdailynews.com/lepic-games-store-vers-lauto-edition/}; \textcolor{teal}{original (2021):} \pangram{https://www.pangram.com/history/8e888b5d-a9eb-484b-aaee-4bd7e1b6a1c2}, \news{https://web.archive.org/web/20211102134405/https://snyderdailynews.com/lepic-games-store-vers-lauto-edition/}).\footnote{Earlier versions were retrieved via the Internet Archive’s Wayback Machine.}
% edited v2: https://www.pangram.com/history/b728d0a4-9d7e-4d79-a62d-4e37ebd96aeb/
% original v2: https://www.pangram.com/history/435d7c8d-2cde-4f75-9484-b5bc280a6027/

 % edited v3.2: SAME (AI), https://www.pangram.com/history/5fd93712-5085-4a8b-b3b5-d9c81b3ec2c4
% original v3.2: SAME (Human), https://www.pangram.com/history/8e888b5d-a9eb-484b-aaee-4bd7e1b6a1c2

% GPTZero: no text in local dataset for snyder_edited_2025 or snyder_original_2021
More concerningly, we also found AI-generated responses in a popular advice column, \textit{Dear Annie}, a practice that risks betraying the trust readers place in such personal guidance (e.g., \textcolor{teal}{Reader:} \pangram{https://www.pangram.com/history/0c9fc890-ef4e-4e65-9e62-09e0f5591401},
\textcolor{purple}{Annie:} \pangram{https://www.pangram.com/history/748801b9-496f-4804-b46c-0165929afe53},
\news{https://www.syracuse.com/advice/2025/10/dear-annie-the-loneliness-of-aging-feels-crushing-as-husbands-dementia-advances.html}).\footnote{We were unable to scrape the \textit{Dear Annie} column to quantify AI use because its publication across many different websites made it difficult to access archival content.}

% reader v2: https://www.pangram.com/history/f382576c-0296-4b59-84f2-09ef32173a6a/
% annie v2: https://www.pangram.com/history/07e0cade-2713-454e-91f5-c2c77ce3ae43/
% reader v3.2: SAME (human), https://www.pangram.com/history/0c9fc890-ef4e-4e65-9e62-09e0f5591401
% annie v3.2: SAME (AI), https://www.pangram.com/history/748801b9-496f-4804-b46c-0165929afe53
% GPTZero reader: 
% gptzero annie: ai 

\paragraph{AI-generated news often contains hallucinations.}

\emph{Hallucinations} are defined as outputs that are fluent and coherent but factually incorrect or unfaithful to the source material \citep{maynez2020faithfulness, ji2023survey}. The widespread hallucinations in news media risks the spread of inaccurate information and reputational damage \cite{bbc2025news_integrity_report}. To understand if news detected as AI-generated is as verifiable as human-written reporting, we manually reviewed 100 \labelai and 100 \labelhuman articles, finding that AI-labeled articles are \textbf{8.2x} more likely to contain hallucinations:\footnote{Fisher’s exact test, $p = 2.3 \times 10^{-9}$.} 41\% of AI-labeled news contained at least one hallucinated claim, whereas only 5\% of articles predicted as human-written had hallucinations.\footnote{A BBC/EBU study found around 55\% of AI-generated responses had accuracy issues \cite{bbc2025news_integrity_report}.} Typical factual errors include fabricated quotes, incorrect statistics, and misdated events (example hallucinations in \autoref{tab:hallucination_cases}). 
\section{Opinions}
\label{sec:opinions}

Opinion articles play a large role in shaping public attitudes, especially those authored by trusted figures such as Nobel laureates, elected officials, and journalists. Even a single op-ed can significantly shift beliefs \cite{coppock2018oped, Bai2025LLMPolicyPersuasion}. As LLMs are often more persuasive than humans \cite{salvi2025persuasiveness, schoenegger2025largelanguagemodelspersuasive}, AI-assisted opinion writing raises concerns of misinformation and ideological amplification \cite{weidinger2022taxonomyofrisks, nehring-etal-2024-large}. To assess these risks, we examine opinion articles written from 2022-2025 in three of the most highly circulated national newspapers (New York Times, Washington Post, and Wall Street Journal). While AI use in \opinionsdata\ remains low relative to \maindata, it has risen sharply following mainstream LLM adoption and is also almost completely undisclosed.

\begin{figure}[t!]
        \centering
        \includegraphics[width=\linewidth]{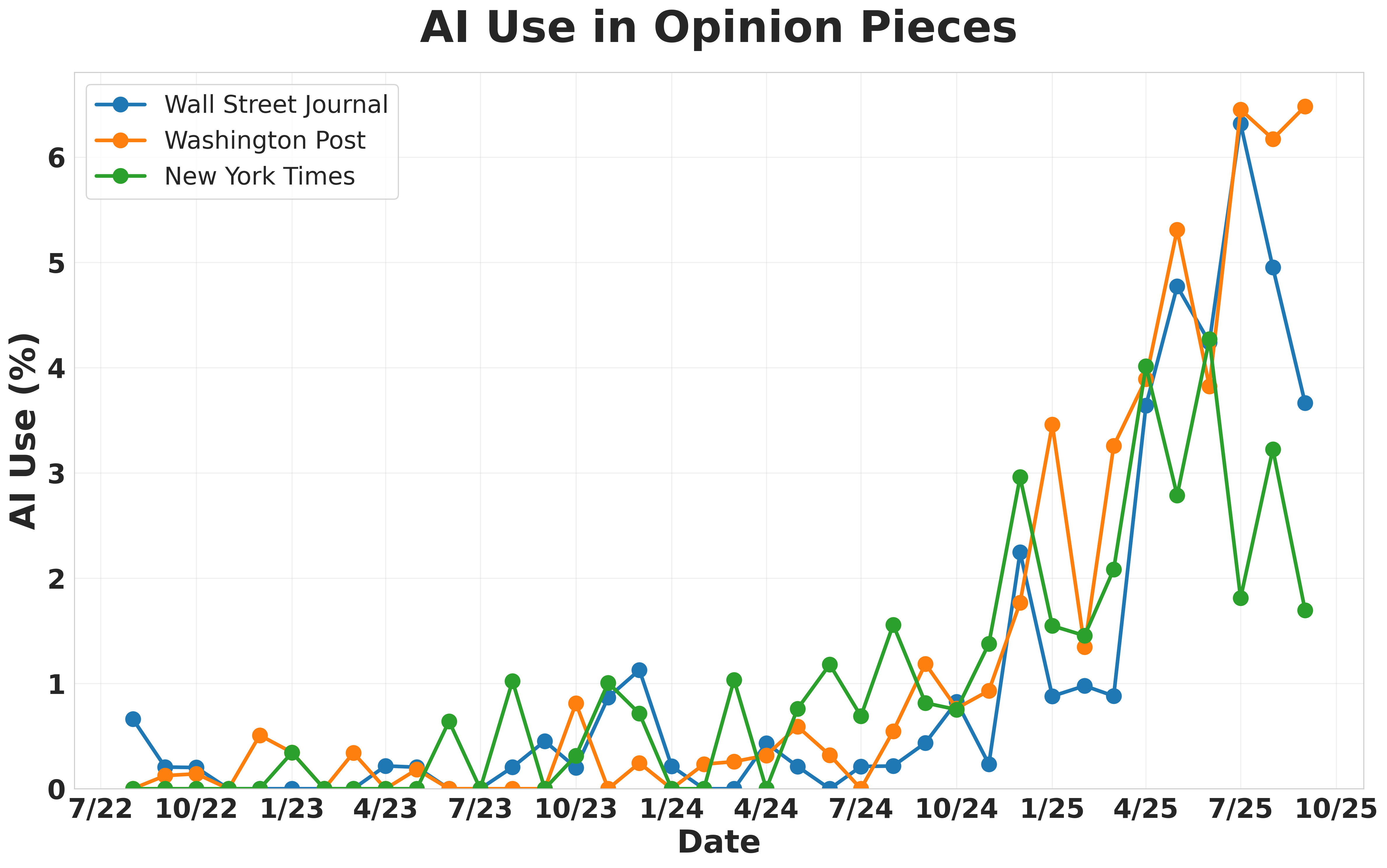}
        \caption{Monthly proportion of \opinionsdata articles flagged for AI use (Sep. 2022 -Sep. 2025). AI use in opinion pieces has dramatically increased over time.}
        \label{fig:ai_time_series}
    \end{figure}

\paragraph{AI use in opinion articles published at reputable newspapers has increased by 25x over the past three years.}
To see whether AI-generated material has risen over time, we measure the share of \opinionsdata flagged as AI between September 2022 and September 2025. AI use increases from 0.1\% in 2022 to 3.4\% in 2025, 
about a \(25\times\) rise, consistent across the three outlets.\footnote{Across all opinion pieces in our sample, 0.1\% are labeled \labelai\ and 0.8\% \labelmixed; see \autoref{fig:opinions_pie_chart}.} By outlet, the AI use share grows from 0.1\% to 3.4\% at the Wall Street Journal, 0.2\% to 4.3\% at the Washington Post, and 0.0\% to 2.6\% at the New York Times (\autoref{fig:ai_time_series}). 
As a pre-ChatGPT sanity check, only 5 of 5,029 opinion articles published before December 2022 are flagged under our AI-use definition (\labelai\ or \labelmixed), implying an empirical false-positive rate of 0.10\% (95\% Wilson CI: 0.04\%--0.23\%).

\paragraph{Opinions exhibit higher AI use than other sections (June–September 2025).}
Across the three outlets, AI use is 6.4 times more likely to occur in opinion pieces than non-opinion news articles published by those same outlets\footnote{These non-opinion articles are extracted from \maindata, which includes WSJ, WaPo, and NYT; see \autoref{tab:opinion_vs_main_ai_rates} for more details.} between June-September 2025 (4.56\% vs.\ 0.71\%; \(n{=}3{,}420\) opinions, \(n{=}10{,}129\) all articles). By outlet, the gap is largest at \textit{The Washington Post} (5.51\% vs.\ 0.55\%), followed by \textit{The Wall Street Journal} (4.99\% vs.\ 0.74\%), and smaller at \textit{The New York Times} (2.94\% vs.\ 1.80\%). Looking 
at how AI is used, \labelmixed\ dominates in both settings (86.5\% of AI use in \opinionsdata\ and 86.1\% of AI use in non-opinion articles at the same outlets). Without clear disclosure standards, readers cannot discern whether AI contributed merely to editing or to greater content generation, restricting readers' ability to judge the appropriateness of AI in specific articles.

\paragraph{Guest contributors are much more likely to use AI in opinion pieces than full-time reporters.}
We observe that 
the majority of 
the 219 unique authors in \opinionsdata\ with at least one article detected as having AI use are infrequent contributors rather than full-time journalists. Categorizing authors by occupation reveals clear occupational divides: political figures, executives, and scientists exhibit the highest AI use, with many using AI for all of their articles in \opinionsdata\. Veteran opinion columnists 
show near-zero AI incidence (<0.5\%). This pattern intuitively makes sense, as guests generally lack established editorial processes and newsroom support \cite{washingtonpost2022opedguide}.

\begin{figure}[t]
    \centering
    \includegraphics[width=\linewidth]{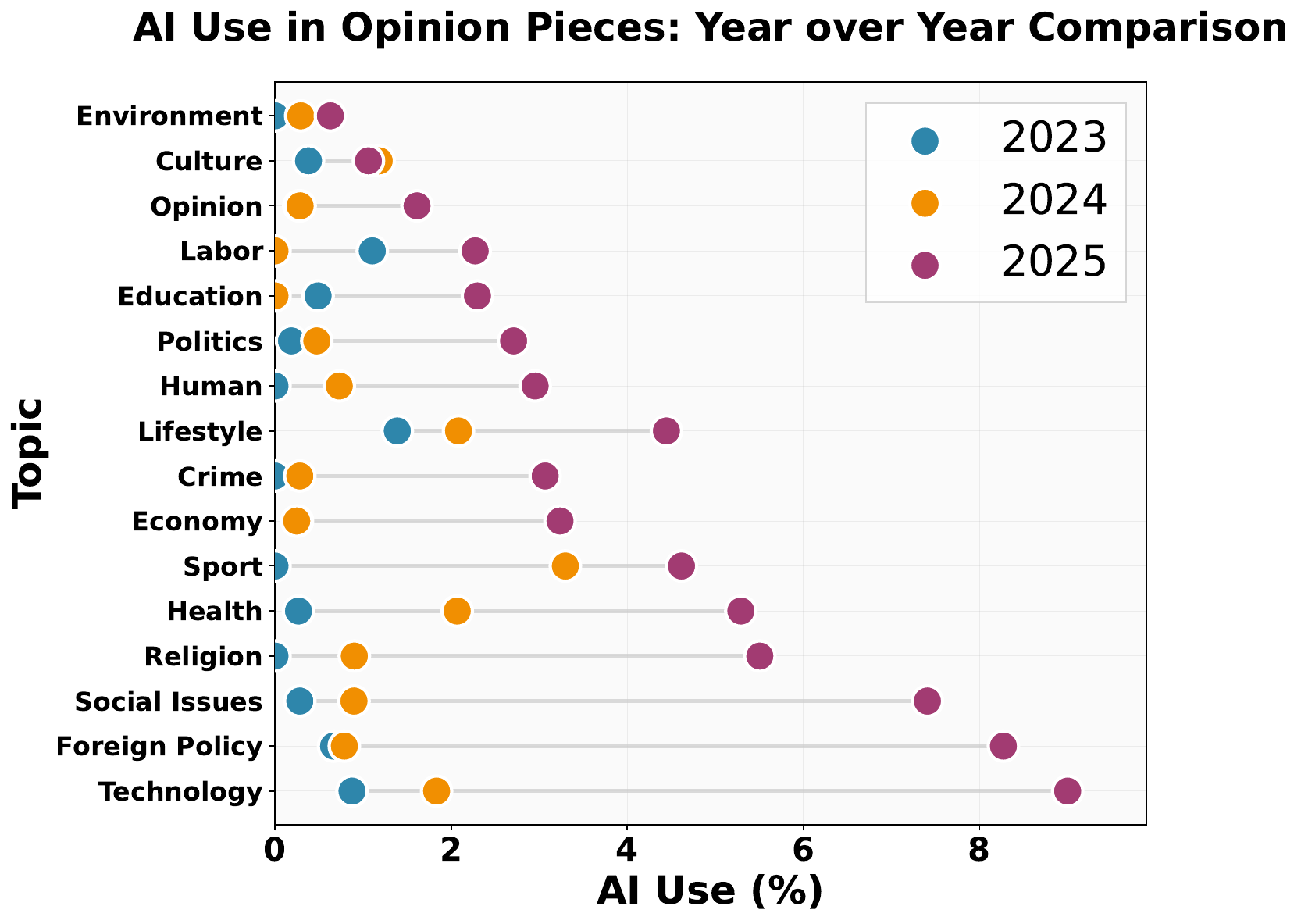}
    \caption{Changes in AI use in \opinionsdata articles year over year (2025 only includes January 1 - September 15). AI likelihood increased across all topics, with especially large gains in opinion articles about Science \& Technology, consistent with topic trends in \maindata.}
    \label{fig:dumbell}
\end{figure}

\paragraph{Topical differences in AI use between \opinionsdata\ and \maindata.}
Setting aside AI use, the majority of opinion articles in our dataset focus on politics and government (56.9\%), far more than in \maindata\ (15.0\%). In \autoref{fig:dumbell}, several categories show large increases since early 2025: \textit{crime and law} (about 31x higher),, \textit{religion} (16x), and \textit{economy and business} (14x) stand out as the fastest-growing. \textit{Science and technology} continues to lead in absolute terms, roughly 9× higher (1.2\%→9.0\%), mirroring its position in \maindata as the topic most likely to contain AI use. Substantial gains also appear in \textit{conflict, war and peace} (12×), \textit{human interest} (10x), and \textit{politics and government} (9×). These patterns indicate that AI use in opinion writing extends well beyond scientific domains and into political, economic, and broadly human-centered discourse.

\section{Related work}
\label{sec:related_work}

\paragraph{AI use in news media.}
Recent scholarship maps how generative AI is impacting newsroom practices. Recent case studies document both benefits and frictions caused by AI within news organizations~\cite{brigham2024developingstorycasestudies,jones2025bbc,ansari2025echoesautomationincreasinguse}, while industry policies study the adoption, governance, and platform–publisher dynamics~\cite{simon2024-ai-in-the-news,brown2025-journalism-zero,simon2025genai-news,ap-standards-2023}. 

\paragraph{Effects of disclosing AI in news.}
Disclosure reliably changes how readers judge identical content~\cite{longoni2022newsfrom}. In recent surveys, comfort reading AI-generated material was low (\labelai: 19\%; \labelmixed: 30\%)~\cite{reuters2025digitalnews}. Disclosing machine authorship reduces perceived credibility~\cite{toff2025disclosure,lee2025aigenerated}, yet readers rarely detect AI without cues and experience~\cite{brown_language_2020,clark_all_2021,russell2025peoplewho}. Moreover, disclosure labels often fail to reduce persuasive impact~\cite{gallegos2025labeling}. 
Despite these mixed effects on perceived credibility, transparency in AI use is important: ethically, newsroom standards emphasize being accountable to the reader \cite{ap-standards-2023, guardian2024policy} and practically, would enhance trust between readers and news outlets using AI \cite{reuters2025digitalnews}. 

\paragraph{Measuring AI content in other domains.} Prior studies audit the growing presence of LLM-generated text across academic and creative domains. In scholarly settings, the influence of AI is measured in peer reviews \cite{liang2024monitoringai, zhou2025largelanguagemodelspenetration} and academic papers \cite{liang2024mapping, luo2025llm4srsurveylargelanguage, kobak2025llmwriting}. \citet{gupta-pruthi-2025-glitters} use AI detection to identify patterns of plagiarism in generated research. Similar detection efforts in creative domains examine AI-generated media on art platforms, Wikipedia, social networks, and news websites \cite{matatov2025examiningprevalencedynamicsaigenerated,brooks-etal-2024-rise,sun-etal-2025-ai, hanley2024-machine-made-media}. Other studies trace the adoption of AI in public communication and consumer complaints \cite{liang2025widespreadadoptionlargelanguage,shin2025adoptionefficacylargelanguage}. 
\section{Conclusion}
\label{sec:conclusion}
Our analysis provides the first 
audit of AI use across \textbf{250k} articles 
of U.S. newspaper articles. We find that AI use is both \emph{widespread} and \emph{uneven}: roughly 9\% of recent articles are flagged as either partially or completely AI-generated, with particularly high rates in smaller local papers, specific topics, ownership groups, and languages other than English. In highly-reputed national newspapers, we observe AI use in opinion pieces is rapidly increasing (from 0\% in 2022 to 3-4\% in 2025), much more so than in other news articles published by the same papers. Based on our audit, we suggest some disclosure policies that newspapers could consider adopting to improve public trust regarding AI use:
\begin{enumerate}
    \item \textbf{Guidance for mixed authorship:} editorial standards should publicly outline what kind of AI use is acceptable without disclosure if any (e.g., grammar checks, style edits), acceptable with disclosure (e.g., summarization, more in-depth rewrites), and not permitted at all (e.g., full article generation). Reporters should record notes and/or log their AI use during the writing process, as these can be useful for more informative disclosure.
    \item \textbf{Explicit AI policies for external contributors:} Since opinion pieces exhibit high rates of detected AI use at top papers, particularly those written by guest contributors, we propose that author attestations about AI use are collected along with article submissions. Editors may additionally want to check submissions  (automatically or manually) for AI cues~\citep{russell2025peoplewho} and publish standards on what they deem acceptable AI use in opinion pieces. 
\end{enumerate}
\section*{Limitations}

Our study focuses primarily on AI use in the U.S. press in English-language publications. While we connect our work to prior studies about hallucination and factual errors in language model generations, we do not perform a large-scale evaluation of the factuality of \labelai\ articles in our dataset beyond studying the authenticity of quotations. Furthermore, although articles in other languages are included in our datasets, they are comparatively much fewer in number; as such, we do not claim that this sample captures the diversity of AI use in journalism globally. 

Our machine translation experiment uses a single MT model (GPT-4.1) and does not account for human translation or other MT methods, which may affect AI detection differently. Additionally, while Pangram reports low false positive rates across languages, detection performance may vary in ways not fully captured by aggregate metrics; we therefore treat our cross-lingual findings as exploratory.

\maindata\ focuses on regional and local newspapers (although national outlets like NYT and WSJ are included), which may bias results towards the journalistic practices of smaller outlets. The inclusion or exclusion of particular newspapers is partially due to data accessibility constraints. Specifically, we are unable to obtain articles from most sites without active RSS feeds, and we are not able to include print-only newspapers. Finally, the data contains inevitable noise. Despite extensive data-cleaning efforts, metadata occasionally appears within article text, which may marginally affect AI detection and topic classification results.

\section*{Ethical Considerations}
\label{sec:ethical}

Our data was collected from publicly accessible newspaper sites, either through RSS feeds or available archives. Given the sensitivity of large-scale text collection, we do not release the complete article texts, but instead provide metadata to respect the rights of content owners. We identify AI-generated text using Pangram, an AI detection model. While Pangram does exhibit very low false-positive rates in benchmark testing, it is not infallible, and all findings should be interpreted as detector outputs rather than definitive authorship attributions. We do not attribute any intent, misconduct, or ethical lapses to the individual journalists and newspapers flagged by the model. We do not draw conclusions about the intent, conduct, or practices of individual journalists, outlets, or companies. Results should not be interpreted as rankings, qualitative judgments, or accusations of wrongdoing. Rather, we purposely use large-scale data collection to understand AI use trends at an aggregate level. We aim to understand what trends in AI use appear across the industry rather than making judgments about specific cases.

\section*{Acknowledgments}

We thank the University of Maryland Computational Linguistics and Information Processing (CLIP) Lab for their feedback and support. We are grateful to the University of Maryland Howard Center for Investigative Reporting, especially Derek Willis and Daniel Trielli, for their guidance. 
This project was partially supported by awards IIS-2046248 and IIS-2312949 from the National Science Foundation (NSF) as well as an award from Open Philanthropy.

\bibliography{custom}

\begin{thebibliography}{67}
\providecommand{\natexlab}[1]{#1}

\bibitem[{Ansari et~al.(2025)Ansari, Zhang, Tripto, and Lee}]{ansari2025echoesautomationincreasinguse}
Abolfazl Ansari, Delvin~Ce Zhang, Nafis~Irtiza Tripto, and Dongwon Lee. 2025.
\newblock \href {https://doi.org/10.1007/978-3-032-07715-8_1} {Echoes of automation: The increasing use of llms in newsmaking}.
\newblock In \emph{Social, Cultural, and Behavioral Modeling: 18th International Conference, SBP-BRiMS 2025, Pittsburgh, PA, USA, October 14–17, 2025, Proceedings}, page 3–13, Berlin, Heidelberg. Springer-Verlag.

\bibitem[{Bai et~al.(2025)Bai, Voelkel, Muldowney, Eichstaedt, and Willer}]{Bai2025LLMPolicyPersuasion}
Hui Bai, Jan~G. Voelkel, Shane Muldowney, Johannes~C. Eichstaedt, and Robb Willer. 2025.
\newblock \href {https://doi.org/10.1038/s41467-025-61345-5} {Llm-generated messages can persuade humans on policy issues}.
\newblock \emph{Nature Communications}, 16:6037.

\bibitem[{Barbaresi(2021)}]{barbaresi-2021-trafilatura}
Adrien Barbaresi. 2021.
\newblock \href {https://aclanthology.org/2021.acl-demo.15} {{Trafilatura: A Web Scraping Library and Command-Line Tool for Text Discovery and Extraction}}.
\newblock In \emph{Proceedings of the Joint Conference of the 59th Annual Meeting of the Association for Computational Linguistics and the 11th International Joint Conference on Natural Language Processing: System Demonstrations}, pages 122--131. Association for Computational Linguistics.

\bibitem[{{BBC News} and {European Broadcasting Union}(2025)}]{bbc2025news_integrity_report}
{BBC News} and {European Broadcasting Union}. 2025.
\newblock \href {https://www.bbc.co.uk/mediacentre/documents/news-integrity-in-ai-assistants-report.pdf} {News integrity in ai assistants}.
\newblock Technical report, BBC Media Centre.
\newblock Accessed: 2025-10-21.

\bibitem[{Brigham et~al.(2024)Brigham, Gao, Kohno, Roesner, and Mireshghallah}]{brigham2024developingstorycasestudies}
Natalie~Grace Brigham, Chongjiu Gao, Tadayoshi Kohno, Franziska Roesner, and Niloofar Mireshghallah. 2024.
\newblock \href {https://arxiv.org/abs/2406.13706} {Developing story: Case studies of generative ai's use in journalism}.
\newblock \emph{Preprint}, arXiv:2406.13706.

\bibitem[{Brooks et~al.(2024)Brooks, Eggert, and Peskoff}]{brooks-etal-2024-rise}
Creston Brooks, Samuel Eggert, and Denis Peskoff. 2024.
\newblock \href {https://doi.org/10.18653/v1/2024.wikinlp-1.12} {The rise of {AI}-generated content in {W}ikipedia}.
\newblock In \emph{Proceedings of the First Workshop on Advancing Natural Language Processing for Wikipedia}, pages 67--79, Miami, Florida, USA. Association for Computational Linguistics.

\bibitem[{Brown and Ja{\'z}wi{\'n}ska(2025)}]{brown2025-journalism-zero}
Peter Brown and Klaudia Ja{\'z}wi{\'n}ska. 2025.
\newblock \href {https://towcenter.columbia.edu/sites/towcenter.columbia.edu/files/content/Journalism%20Zero_%20How%20Platforms%20and%20Publishers%20are%20Navigating%20AI.pdf} {Journalism zero: How platforms and publishers are navigating ai}.
\newblock Report, Tow Center for Digital Journalism, Columbia University.

\bibitem[{Brown et~al.(2020)Brown, Mann, Ryder, Subbiah, Kaplan, Dhariwal, Neelakantan, Shyam, Sastry, Askell, Agarwal, Herbert-Voss, Krueger, Henighan, Child, Ramesh, Ziegler, Wu, Winter, Hesse, Chen, Sigler, Litwin, Gray, Chess, Clark, Berner, McCandlish, Radford, Sutskever, and Amodei}]{brown_language_2020}
Tom Brown, Benjamin Mann, Nick Ryder, Melanie Subbiah, Jared~D Kaplan, Prafulla Dhariwal, Arvind Neelakantan, Pranav Shyam, Girish Sastry, Amanda Askell, Sandhini Agarwal, Ariel Herbert-Voss, Gretchen Krueger, Tom Henighan, Rewon Child, Aditya Ramesh, Daniel Ziegler, Jeffrey Wu, Clemens Winter, Chris Hesse, Mark Chen, Eric Sigler, Mateusz Litwin, Scott Gray, Benjamin Chess, Jack Clark, Christopher Berner, Sam McCandlish, Alec Radford, Ilya Sutskever, and Dario Amodei. 2020.
\newblock \href {https://proceedings.neurips.cc/paper_files/paper/2020/file/1457c0d6bfcb4967418bfb8ac142f64a-Paper.pdf} {Language models are few-shot learners}.
\newblock In \emph{Advances in Neural Information Processing Systems}, volume~33, pages 1877--1901. Curran Associates, Inc.

\bibitem[{Cavazos and Sterling(2024)}]{cavazos2024highcost}
Roberto Cavazos and Greg Sterling. 2024.
\newblock The high cost of review fraud: An economic analysis of consumer harm.
\newblock \url{https://askfortransparency.com/research/high-cost-of-review-fraud/}.
\newblock Accessed: 2025-10-17.

\bibitem[{Clark et~al.(2021)Clark, August, Serrano, Haduong, Gururangan, and Smith}]{clark_all_2021}
Elizabeth Clark, Tal August, Sofia Serrano, Nikita Haduong, Suchin Gururangan, and Noah~A. Smith. 2021.
\newblock \href {https://doi.org/10.18653/v1/2021.acl-long.565} {All that`s {\textquoteleft}human' is not gold: Evaluating human evaluation of generated text}.
\newblock In \emph{Proceedings of the 59th Annual Meeting of the Association for Computational Linguistics and the 11th International Joint Conference on Natural Language Processing (Volume 1: Long Papers)}, pages 7282--7296, Online. Association for Computational Linguistics.

\bibitem[{Conaghan(2017)}]{conaghan2017median}
Jim Conaghan. 2017.
\newblock Young, old and in-between: Newspaper platform readers ages are well-distributed.
\newblock \url{https://www.newsmediaalliance.org/age-newspaper-readers-platforms/}.
\newblock Highlights Nielsen Scarborough USA+ 2016 Release 2 finding that the median age of daily print-newspaper readers is 57.9 years.

\bibitem[{Coppock et~al.(2018)Coppock, Ekins, and Kirby}]{coppock2018oped}
Alexander Coppock, Emily Ekins, and David Kirby. 2018.
\newblock \href {https://isps.yale.edu/sites/default/files/publication/2018/04/qjps_2018_coppock_op-eds_100.00016112.pdf} {The long-lasting effects of newspaper op-eds on public opinion}.
\newblock \emph{Quarterly Journal of Political Science}, 13(1):59--87.

\bibitem[{Dugan et~al.(2025)Dugan, Zhu, Alam, Nakov, Apidianaki, and Callison-Burch}]{dugan-etal-2025-genai}
Liam Dugan, Andrew Zhu, Firoj Alam, Preslav Nakov, Marianna Apidianaki, and Chris Callison-Burch. 2025.
\newblock \href {https://aclanthology.org/2025.genaidetect-1.45/} {{G}en{AI} content detection task 3: Cross-domain machine generated text detection challenge}.
\newblock In \emph{Proceedings of the 1stWorkshop on GenAI Content Detection (GenAIDetect)}, pages 377--388, Abu Dhabi, UAE. International Conference on Computational Linguistics.

\bibitem[{Emi(2025)}]{emi2025falsepositives}
Bradley Emi. 2025.
\newblock All about false positives in ai detectors.
\newblock \url{https://www.pangram.com/blog/all-about-false-positives-in-ai-detectors}.
\newblock Pangram Labs blog.

\bibitem[{Emi and Spero(2024)}]{emi2024technicalreportpangramaigenerated}
Bradley Emi and Max Spero. 2024.
\newblock \href {https://arxiv.org/abs/2402.14873} {Technical {R}eport on the {P}angram {AI}-{G}enerated {T}ext {C}lassifier}.
\newblock \emph{Preprint}, arXiv:2402.14873.

\bibitem[{Evanko and Natale(2025)}]{evanko2025quantifying}
Daniel~S. Evanko and Michael~Di Natale. 2025.
\newblock \href {https://peerreviewcongress.org/abstract/quantifying-and-assessing-the-use-of-generative-ai-by-authors-and-reviewers-in-the-cancer-research-field/} {Quantifying and assessing the use of generative ai by authors and reviewers in the cancer research field}.
\newblock In \emph{Peer Review Congress}.
\newblock Accessed 2025-10-17.

\bibitem[{Gallegos et~al.(2024)Gallegos, Rossi, Barrow, Tanjim, Kim, Dernoncourt, Yu, Zhang, and Ahmed}]{gallegos-etal-2024-bias}
Isabel~O. Gallegos, Ryan~A. Rossi, Joe Barrow, Md~Mehrab Tanjim, Sungchul Kim, Franck Dernoncourt, Tong Yu, Ruiyi Zhang, and Nesreen~K. Ahmed. 2024.
\newblock \href {https://doi.org/10.1162/coli_a_00524} {Bias and fairness in large language models: A survey}.
\newblock \emph{Computational Linguistics}, 50(3):1097--1179.

\bibitem[{Gallegos et~al.(2025)Gallegos, Shani, Shi, Bianchi, Willer, and Jurafsky}]{gallegos2025labeling}
Isabel~O. Gallegos, Chen Shani, Weiyan Shi, Federico Bianchi, Robb Willer, and Dan Jurafsky. 2025.
\newblock \href {https://openreview.net/forum?id=R7FHdsZv1w} {{AI}-generated content and public persuasion: The limited effect of {AI} authorship labels}.
\newblock In \emph{NeurIPS 2024 Workshop on Regulatable ML}.

\bibitem[{Gupta and Pruthi(2025)}]{gupta-pruthi-2025-glitters}
Tarun Gupta and Danish Pruthi. 2025.
\newblock \href {https://doi.org/10.18653/v1/2025.acl-long.1249} {All that glitters is not novel: Plagiarism in {AI} generated research}.
\newblock In \emph{Proceedings of the 63rd Annual Meeting of the Association for Computational Linguistics (Volume 1: Long Papers)}, pages 25721--25738, Vienna, Austria. Association for Computational Linguistics.

\bibitem[{Hanley and Durumeric(2024)}]{hanley2024-machine-made-media}
Hans W.~A. Hanley and Zakir Durumeric. 2024.
\newblock \href {https://doi.org/10.1609/icwsm.v18i1.31333} {Machine-made media: Monitoring the mobilization of machine-generated articles on misinformation and mainstream news websites}.
\newblock \emph{Proceedings of the International AAAI Conference on Web and Social Media}, 18(1):542--556.
\newblock ICWSM 2024.

\bibitem[{Hu et~al.(2025)Hu, Kyrychenko, Rathje et~al.}]{hu2025identitybias}
Tony Hu, Yuliya Kyrychenko, Steve Rathje, et~al. 2025.
\newblock \href {https://doi.org/10.1038/s43588-024-00741-1} {Generative language models exhibit social identity biases}.
\newblock \emph{Nature Computational Science}, 5:65--75.

\bibitem[{{IPTC}(2025)}]{iptc2025mediatopics}
{IPTC}. 2025.
\newblock Media topics.
\newblock \url{https://iptc.org/standards/media-topics/}.
\newblock Accessed: 2025-09-29.

\bibitem[{Jabarian and Imas(2025)}]{jabarian2025artificial}
Brian Jabarian and Alex Imas. 2025.
\newblock \href {https://www.nber.org/papers/w34223} {Artificial writing and automated detection}.

\bibitem[{Ji et~al.(2023{\natexlab{a}})Ji, Lee, Frieske, Yu, Su, Xu, Ishii, Bang, Madotto, and Fung}]{ji2023surveyofhallucination}
Ziwei Ji, Nayeon Lee, Rita Frieske, Tiezheng Yu, Dan Su, Yan Xu, Etsuko Ishii, Ye~Jin Bang, Andrea Madotto, and Pascale Fung. 2023{\natexlab{a}}.
\newblock \href {https://doi.org/10.1145/3571730} {Survey of hallucination in natural language generation}.
\newblock \emph{ACM Comput. Surv.}, 55(12).

\bibitem[{Ji et~al.(2023{\natexlab{b}})Ji, Lee, Frieske, Yu, Su, Xu, Ishii, Bang, Madotto, and Fung}]{ji2023survey}
Ziwei Ji, Nayeon Lee, Rita Frieske, Tiezheng Yu, Dan Su, Yan Xu, Etsuko Ishii, Ye~Jin Bang, Andrea Madotto, and Pascale Fung. 2023{\natexlab{b}}.
\newblock \href {https://doi.org/10.1145/3571730} {Survey of hallucination in natural language generation}.
\newblock \emph{ACM Comput. Surv.}, 55(12).

\bibitem[{Jones and Jones(2025)}]{jones2025bbc}
Bethany Jones and Richard Jones. 2025.
\newblock \href {https://doi.org/10.1177/14648849251317150} {Action research at the {BBC}: Interrogating artificial intelligence with journalists to generate actionable insights for the newsroom}.
\newblock \emph{Journalism}, 26(8):1708--1725.
\newblock Original work published 2025.

\bibitem[{Kennedy et~al.(2025)Kennedy, Yam, Kikuchi, Pula, and Fuentes}]{pew2025-ai-views}
Brian Kennedy, Eileen Yam, Emma Kikuchi, Isabelle Pula, and Javier Fuentes. 2025.
\newblock \href {https://www.pewresearch.org/science/2025/09/17/how-americans-view-ai-and-its-impact-on-people-and-society/} {How americans view {AI} and its impact on people and society}.
\newblock Pew Research Center, Report.

\bibitem[{Knibbs(2024)}]{knibbs2024aiSlop}
Kate Knibbs. 2024.
\newblock \href {https://www.wired.com/story/ai-generated-medium-posts-content-moderation/} {Ai slop is flooding medium}.
\newblock \emph{Wired}.
\newblock Accessed: 2025-10-17.

\bibitem[{Kobak et~al.(2025)Kobak, Gonz{\'a}lez-M{\'a}rquez, Horv{\'a}t, and Lause}]{kobak2025llmwriting}
Dmitry Kobak, Roc{\'i}o Gonz{\'a}lez-M{\'a}rquez, {\'E}va~A. Horv{\'a}t, and Jana Lause. 2025.
\newblock \href {https://doi.org/10.1126/sciadv.adt3813} {Delving into llm-assisted writing in biomedical publications through excess vocabulary}.
\newblock \emph{Science Advances}, 11(27):eadt3813.
\newblock PMID: 40601754; PMCID: PMC12219543.

\bibitem[{Lee et~al.(2025)Lee, Jhang, and Baek}]{lee2025aigenerated}
Daniel~Chaein Lee, Jihoon Jhang, and Tae~Hyun Baek. 2025.
\newblock \href {https://doi.org/10.1080/10447318.2025.2477739} {Ai-generated news content: The impact of {AI} writer identity and perceived {AI} human-likeness}.
\newblock \emph{International Journal of Human-Computer Interaction}, pages 1--13.
\newblock Published online 2025-03-24; Taylor \& Francis.

\bibitem[{Liang et~al.(2024{\natexlab{a}})Liang, Izzo, Zhang, Lepp, Cao, Zhao, Chen, Ye, Liu, Huang, McFarland, and Zou}]{liang2024monitoringai}
Weixin Liang, Zachary Izzo, Yaohui Zhang, Haley Lepp, Hancheng Cao, Xuandong Zhao, Lingjiao Chen, Haotian Ye, Sheng Liu, Zhi Huang, Daniel~A. McFarland, and James~Y. Zou. 2024{\natexlab{a}}.
\newblock \href {https://dl.acm.org/doi/10.5555/3692070.3693262} {Monitoring ai-modified content at scale: a case study on the impact of chatgpt on ai conference peer reviews}.
\newblock In \emph{Proceedings of the 41st International Conference on Machine Learning}, ICML'24. JMLR.org.

\bibitem[{Liang et~al.(2025)Liang, Zhang, Codreanu, Wang, Cao, and Zou}]{liang2025widespreadadoptionlargelanguage}
Weixin Liang, Yaohui Zhang, Mihai Codreanu, Jiayu Wang, Hancheng Cao, and James Zou. 2025.
\newblock \href {https://arxiv.org/abs/2502.09747} {The widespread adoption of large language model-assisted writing across society}.
\newblock \emph{Preprint}, arXiv:2502.09747.

\bibitem[{Liang et~al.(2024{\natexlab{b}})Liang, Zhang, Wu, Lepp, Ji, Zhao, Cao, Liu, He, Huang, Yang, Potts, Manning, and Zou}]{liang2024mapping}
Weixin Liang, Yaohui Zhang, Zhengxuan Wu, Haley Lepp, Wenlong Ji, Xuandong Zhao, Hancheng Cao, Sheng Liu, Siyu He, Zhi Huang, Diyi Yang, Christopher Potts, Christopher~D Manning, and James~Y. Zou. 2024{\natexlab{b}}.
\newblock \href {https://openreview.net/forum?id=YX7QnhxESU} {Mapping the increasing use of {LLM}s in scientific papers}.
\newblock In \emph{First Conference on Language Modeling}.

\bibitem[{Lipka and Eddy(2025)}]{pew2025-chatbots-news}
Michael Lipka and Kirsten Eddy. 2025.
\newblock \href {https://www.pewresearch.org/short-reads/2025/10/01/relatively-few-americans-are-getting-news-from-ai-chatbots-like-chatgpt/} {Relatively few americans are getting news from {AI} chatbots like {ChatGPT}}.
\newblock Pew Research Center, Short Reads.

\bibitem[{Longoni et~al.(2022)Longoni, Fradkin, Cian, and Pennycook}]{longoni2022newsfrom}
Chiara Longoni, Andrey Fradkin, Luca Cian, and Gordon Pennycook. 2022.
\newblock \href {https://doi.org/10.1145/3531146.3533077} {News from generative artificial intelligence is believed less}.
\newblock In \emph{Proceedings of the 2022 ACM Conference on Fairness, Accountability, and Transparency}, FAccT '22, page 97–106, New York, NY, USA. Association for Computing Machinery.

\bibitem[{Luo et~al.(2025)Luo, Yang, Xu, Yang, and Du}]{luo2025llm4srsurveylargelanguage}
Ziming Luo, Zonglin Yang, Zexin Xu, Wei Yang, and Xinya Du. 2025.
\newblock \href {https://arxiv.org/abs/2501.04306} {Llm4sr: A survey on large language models for scientific research}.
\newblock \emph{Preprint}, arXiv:2501.04306.

\bibitem[{Matatov et~al.(2025)Matatov, Quéré, Amir, and Naaman}]{matatov2025examiningprevalencedynamicsaigenerated}
Hana Matatov, Marianne Aubin~Le Quéré, Ofra Amir, and Mor Naaman. 2025.
\newblock \href {https://arxiv.org/abs/2410.07302} {Examining the prevalence and dynamics of ai-generated media in art subreddits}.
\newblock \emph{Preprint}, arXiv:2410.07302.

\bibitem[{Maynez et~al.(2020{\natexlab{a}})Maynez, Narayan, Bohnet, and McDonald}]{maynez-etal-2020-faithfulness}
Joshua Maynez, Shashi Narayan, Bernd Bohnet, and Ryan McDonald. 2020{\natexlab{a}}.
\newblock \href {https://doi.org/10.18653/v1/2020.acl-main.173} {On faithfulness and factuality in abstractive summarization}.
\newblock In \emph{Proceedings of the 58th Annual Meeting of the Association for Computational Linguistics}, pages 1906--1919, Online. Association for Computational Linguistics.

\bibitem[{Maynez et~al.(2020{\natexlab{b}})Maynez, Narayan, Bohnet, and McDonald}]{maynez2020faithfulness}
Joshua Maynez, Shashi Narayan, Bernd Bohnet, and Ryan McDonald. 2020{\natexlab{b}}.
\newblock \href {https://doi.org/10.18653/v1/2020.acl-main.173} {On faithfulness and factuality in abstractive summarization}.
\newblock In \emph{Proceedings of the 58th Annual Meeting of the Association for Computational Linguistics}, pages 1906--1919, Online. Association for Computational Linguistics.

\bibitem[{{Medill Local News Initiative}(2024)}]{medill2024ai_local}
{Medill Local News Initiative}. 2024.
\newblock Ai and local news: Report 2024.
\newblock Technical report, Northwestern University.
\newblock Accessed: 2025-10-01.

\bibitem[{Meir(2023)}]{ap-standards-2023}
Nicole Meir. 2023.
\newblock \href {https://www.ap.org/the-definitive-source/behind-the-news/standards-around-generative-ai/} {Standards around generative ai}.
\newblock The Definitive Source: Behind the News, The Associated Press.
\newblock Accessed on 2025-10-15.

\bibitem[{Metzger(2024)}]{medill2024state}
Zachary Metzger. 2024.
\newblock The state of local news 2024.
\newblock Technical report, Northwestern University.
\newblock Accessed: 2025-10-01.

\bibitem[{Nehring et~al.(2024)Nehring, Gabryszak, J{\"u}rgens, Burchardt, Schaffer, Spielkamp, and Stark}]{nehring-etal-2024-large}
Jan Nehring, Aleksandra Gabryszak, Pascal J{\"u}rgens, Aljoscha Burchardt, Stefan Schaffer, Matthias Spielkamp, and Birgit Stark. 2024.
\newblock \href {https://aclanthology.org/2024.lrec-main.884/} {Large language models are echo chambers}.
\newblock In \emph{Proceedings of the 2024 Joint International Conference on Computational Linguistics, Language Resources and Evaluation (LREC-COLING 2024)}, pages 10117--10123, Torino, Italia. ELRA and ICCL.

\bibitem[{Newman et~al.(2025)Newman, Arguedas, Robertson, Nielsen, and Fletcher}]{reuters2025digitalnews}
Nic Newman, Amy~Ross Arguedas, Craig~T. Robertson, Rasmus~Kleis Nielsen, and Richard Fletcher. 2025.
\newblock Digital news report 2025.
\newblock \url{https://reutersinstitute.politics.ox.ac.uk/sites/default/files/2025-06/Digital_News-Report_2025.pdf}.
\newblock Accessed: 2025-09-30.

\bibitem[{OpenAI(2022)}]{openai2022chatgpt}
OpenAI. 2022.
\newblock Introducing {ChatGPT}.
\newblock \url{https://openai.com/blog/chatgpt}.
\newblock Accessed: 2025-10-06.

\bibitem[{{OpenAI}(2024)}]{openai2024gpt41}
{OpenAI}. 2024.
\newblock Gpt-4.1.
\newblock \url{https://platform.openai.com/docs/models/gpt-4.1}.
\newblock Accessed 2025.

\bibitem[{{Pangram}(2024)}]{pangram2024multilingual}
{Pangram}. 2024.
\newblock Pangram multilingual v2: Expansion of ai content detection into multiple languages.
\newblock \url{https://www.pangram.com/blog/pangram-multilingual-v2}.
\newblock Pangram Labs blog.

\bibitem[{{Pew Research Center}(2025)}]{pew2025news-print-read}
{Pew Research Center}. 2025.
\newblock News platforms (news \& information) – american trends panel wave 177: Topline results.
\newblock \url{https://www.pewresearch.org/wp-content/uploads/sites/20/2025/09/PJ_2025.09.25_news-information-fact-sheets_topline.pdf}.

\bibitem[{{Qwen Team}(2025)}]{qwen3-2025}
{Qwen Team}. 2025.
\newblock \href {https://arxiv.org/abs/2505.09388} {Qwen3 technical report}.
\newblock \emph{Preprint}, arXiv:2505.09388.

\bibitem[{Radcliffe(2025)}]{radcliffe2025journalism}
Damian Radcliffe. 2025.
\newblock \href {https://www.trust.org/wp-content/uploads/2025/01/TRF-Insights-Journalism-in-the-AI-Era.pdf} {Journalism in the {AI} era: Opportunities and challenges in the global south and emerging economies}.
\newblock Trf insights report, Thomson Reuters Foundation.
\newblock Survey of 200+ journalists from 70+ countries.

\bibitem[{Russell et~al.(2025)Russell, Karpinska, and Iyyer}]{russell2025peoplewho}
Jenna Russell, Marzena Karpinska, and Mohit Iyyer. 2025.
\newblock \href {https://doi.org/10.18653/v1/2025.acl-long.267} {People who frequently use {C}hat{GPT} for writing tasks are accurate and robust detectors of {AI}-generated text}.
\newblock In \emph{Proceedings of the 63rd Annual Meeting of the Association for Computational Linguistics (Volume 1: Long Papers)}, pages 5342--5373, Vienna, Austria. Association for Computational Linguistics.

\bibitem[{Salvi et~al.(2025)Salvi, Horta~Ribeiro, Gallotti et~al.}]{salvi2025persuasiveness}
Fabio Salvi, Manoel Horta~Ribeiro, Riccardo Gallotti, et~al. 2025.
\newblock \href {https://doi.org/10.1038/s41562-025-02194-6} {On the conversational persuasiveness of gpt-4}.
\newblock \emph{Nature Human Behaviour}, 9(10):1645--1653.

\bibitem[{Schoenegger et~al.(2025)Schoenegger, Salvi, Liu, Nan, Debnath, Fasolo, Leivada, Recchia, Günther, Zarifhonarvar, Kwon, Islam, Dehnert, Lee, Reinecke, Kamper, Kobaş, Sandford, Kgomo, Hewitt, Kapoor, Oktar, Kucuk, Feng, Jones, Gainsburg, Olschewski, Heinzelmann, Cruz, Tappin, Ma, Park, Onyonka, Hjorth, Slattery, Zeng, Finke, Grossmann, Salatiello, and Karger}]{schoenegger2025largelanguagemodelspersuasive}
Philipp Schoenegger, Francesco Salvi, Jiacheng Liu, Xiaoli Nan, Ramit Debnath, Barbara Fasolo, Evelina Leivada, Gabriel Recchia, Fritz Günther, Ali Zarifhonarvar, Joe Kwon, Zahoor~Ul Islam, Marco Dehnert, Daryl Y.~H. Lee, Madeline~G. Reinecke, David~G. Kamper, Mert Kobaş, Adam Sandford, Jonas Kgomo, Luke Hewitt, Shreya Kapoor, Kerem Oktar, Eyup~Engin Kucuk, Bo~Feng, Cameron~R. Jones, Izzy Gainsburg, Sebastian Olschewski, Nora Heinzelmann, Francisco Cruz, Ben~M. Tappin, Tao Ma, Peter~S. Park, Rayan Onyonka, Arthur Hjorth, Peter Slattery, Qingcheng Zeng, Lennart Finke, Igor Grossmann, Alessandro Salatiello, and Ezra Karger. 2025.
\newblock \href {https://arxiv.org/abs/2505.09662} {Large language models are more persuasive than incentivized human persuaders}.
\newblock \emph{Preprint}, arXiv:2505.09662.

\bibitem[{Shin et~al.(2025)Shin, Kim, and Shin}]{shin2025adoptionefficacylargelanguage}
Minkyu Shin, Jin Kim, and Jiwoong Shin. 2025.
\newblock \href {https://arxiv.org/abs/2311.16466} {The adoption and efficacy of large language models: Evidence from consumer complaints in the financial industry}.
\newblock \emph{Preprint}, arXiv:2311.16466.

\bibitem[{Simon(2024)}]{simon2024-ai-in-the-news}
Felix~M. Simon. 2024.
\newblock \href {https://doi.org/10.7916/ncm5-3v06} {Artificial intelligence in the news: How ai retools, rationalizes, and reshapes journalism and the public arena}.
\newblock Report, Tow Center for Digital Journalism, Columbia University.

\bibitem[{Simon et~al.(2025)Simon, Nielsen, and Fletcher}]{simon2025genai-news}
Felix~M. Simon, Rasmus~Kleis Nielsen, and Richard Fletcher. 2025.
\newblock \href {https://doi.org/10.60625/risj-5bjv-yt69} {Generative ai and news report 2025: How people think about ai's role in journalism and society}.
\newblock Report, Reuters Institute for the Study of Journalism, University of Oxford.

\bibitem[{Su et~al.(2024)Su, Cardie, and Nakov}]{su-etal-2024-adapting}
Jinyan Su, Claire Cardie, and Preslav Nakov. 2024.
\newblock \href {https://doi.org/10.18653/v1/2024.findings-naacl.95} {Adapting fake news detection to the era of large language models}.
\newblock In \emph{Findings of the Association for Computational Linguistics: NAACL 2024}, pages 1473--1490, Mexico City, Mexico. Association for Computational Linguistics.

\bibitem[{Sun et~al.(2025)Sun, Zhang, Shen, Zhang, Liu, Backes, Zhang, and He}]{sun-etal-2025-ai}
Zhen Sun, Zongmin Zhang, Xinyue Shen, Ziyi Zhang, Yule Liu, Michael Backes, Yang Zhang, and Xinlei He. 2025.
\newblock \href {https://doi.org/10.18653/v1/2025.acl-long.1120} {Are we in the {AI}-generated text world already? quantifying and monitoring {AIGT} on social media}.
\newblock In \emph{Proceedings of the 63rd Annual Meeting of the Association for Computational Linguistics (Volume 1: Long Papers)}, pages 22975--23005, Vienna, Austria. Association for Computational Linguistics.

\bibitem[{Thai et~al.(2025)Thai, Emi, Masrour, and Iyyer}]{thai2025editlensquantifyingextentai}
Katherine Thai, Bradley Emi, Elyas Masrour, and Mohit Iyyer. 2025.
\newblock \href {https://arxiv.org/abs/2510.03154} {Editlens: Quantifying the extent of ai editing in text}.
\newblock \emph{Preprint}, arXiv:2510.03154.

\bibitem[{Tian and Cui(2023)}]{tian2023gptzero}
Edward Tian and Alexander Cui. 2023.
\newblock {GPTZero}: Towards detection of {AI}-generated text using zero-shot and supervised methods.
\newblock \url{https://gptzero.me}.
\newblock Accessed: 2025-10-09.

\bibitem[{Toff and Simon(2025)}]{toff2025disclosure}
Benjamin Toff and Felix~M. Simon. 2025.
\newblock \href {https://doi.org/10.1177/19401612241308697} {{``Or They Could Just Not Use It?'': The Dilemma of {AI} Disclosure for Audience Trust in News}}.
\newblock \emph{The International Journal of Press/Politics}, 30(4):881--903.

\bibitem[{{UNC Center for Innovation \& Sustainability in Local Media}(2020)}]{usnewsdeserts2020}
{UNC Center for Innovation \& Sustainability in Local Media}. 2020.
\newblock Newspapers 2020 — active u.s. newspapers as of 2020.
\newblock \url{https://usnewsdeserts.cislm.org/}.
\newblock Accessed: 2025-09-29.

\bibitem[{Viner and Bateson(2023)}]{guardian2024policy}
Katharine Viner and Anna Bateson. 2023.
\newblock \href {https://www.theguardian.com/help/insideguardian/2023/jun/16/the-guardians-approach-to-generative-ai} {The guardian’s approach to generative ai}.
\newblock Accessed: 2025-10-17.

\bibitem[{{Washington Post Staff}(2022)}]{washingtonpost2022opedguide}
{Washington Post Staff}. 2022.
\newblock The washington post guide to writing an opinion article.
\newblock \url{https://www.washingtonpost.com/opinions/2022/op-ed-writing-guide-washington-post-examples/}.
\newblock Updated June 23, 2022; accessed 2025-10-16.

\bibitem[{Weidinger et~al.(2022)Weidinger, Uesato, Rauh, Griffin, Huang, Mellor, Glaese, Cheng, Balle, Kasirzadeh, Biles, Brown, Kenton, Hawkins, Stepleton, Birhane, Hendricks, Rimell, Isaac, Haas, Legassick, Irving, and Gabriel}]{weidinger2022taxonomyofrisks}
Laura Weidinger, Jonathan Uesato, Maribeth Rauh, Conor Griffin, Po-Sen Huang, John Mellor, Amelia Glaese, Myra Cheng, Borja Balle, Atoosa Kasirzadeh, Courtney Biles, Sasha Brown, Zac Kenton, Will Hawkins, Tom Stepleton, Abeba Birhane, Lisa~Anne Hendricks, Laura Rimell, William Isaac, Julia Haas, Sean Legassick, Geoffrey Irving, and Iason Gabriel. 2022.
\newblock \href {https://doi.org/10.1145/3531146.3533088} {Taxonomy of risks posed by language models}.
\newblock In \emph{Proceedings of the 2022 ACM Conference on Fairness, Accountability, and Transparency}, FAccT '22, page 214–229, New York, NY, USA. Association for Computing Machinery.

\bibitem[{Yam and Kennedy(2025)}]{pew2025-ai-reaction}
Eileen Yam and Brian Kennedy. 2025.
\newblock \href {https://www.pewresearch.org/short-reads/2025/09/17/from-political-speeches-to-songs-how-would-americans-react-if-they-found-out-ai-was-involved/} {From political speeches to songs, how would americans react if they found out {AI} was involved?}
\newblock Pew Research Center, Short Reads.

\bibitem[{Zhou et~al.(2025)Zhou, Zhang, Dai, Hershcovich, and Li}]{zhou2025largelanguagemodelspenetration}
Li~Zhou, Ruijie Zhang, Xunlian Dai, Daniel Hershcovich, and Haizhou Li. 2025.
\newblock \href {https://arxiv.org/abs/2502.11193} {Large language models penetration in scholarly writing and peer review}.
\newblock \emph{Preprint}, arXiv:2502.11193.

\end{thebibliography}

\appendix
\section{Dataset}
\label{app:dataset}

In this section of the appendix we provide more details on our data collection process.

\paragraph{\maindata additional collection details.}
To facilitate automatic data collection, we first obtain a list of 6,175 URLs for American newspapers\footnote{\url{https://onlinenewspapers.com/usstate/usatable.shtml}} and filter out those that are unreachable and/or do not have active RSS feeds, which yielded 1,528 URLs. Roughly twice a week, from June 15 to September 15, we automatically accessed each RSS feed and downloaded the full text and metadata for up to 50 recently-published articles from each paper.\footnote{While the majority of \maindata\ is obtained from RSS feeds, articles from some newspapers (e.g., Washington Post, Wall Street Journal) were accessed via \emph{ProQuest Recent Newspapers}.} 
Each full text article was then preprocessed using the Trafilatura~\citep{barbaresi-2021-trafilatura} and Newspaper4K\footnote{\url{https://github.com/AndyTheFactory/newspaper4k}} libraries to strip headers/footers, advertisements, and HTML artifacts from the text.   

\paragraph{Topic classification.} 

Each instance was presented to the \textsc{Qwen3-8B} model with the full article text and an instruction to select the single most semantically appropriate topic from the IPTC taxonomy. The complete prompt is shown in \autoref{prompt:topics}. To verify, we randomly sampled 100 articles from the \maindata\ set. Two of the authors independently re-labeled these samples according to the same IPTC taxonomy. Inter-annotator agreement between the two human raters was 87\% (Cohen’s~$\kappa=0.85$), reflecting strong consistency in human judgments. Agreement between the \textsc{Qwen3-8B} predictions and the majority human label averaged 77\%, indicating moderately strong alignment between the classifier and human annotations. 

\begin{figure*}[!tb]
\centering
\begin{tcolorbox}[colback=gray!5!white, colframe=teal, title=Prompt for classifying topic of articles]
\lstset{
    basicstyle=\ttfamily\footnotesize,
    breaklines=true,
    frame=none,
    xleftmargin=0pt,
    framexleftmargin=0pt,
    columns=fullflexible,
    tabsize=1,
    breakindent=0pt,
    breakautoindent=false,
    postbreak=\space,
    showstringspaces=false,
}
\begin{lstlisting}[language=Markdown]
Classify the following article into the most appropriate primary topic from the IPTC taxonomy.

IPTC Primary Topics:
{taxonomy_text}

Article to classify:
Title: {article.title}
Text: {article_text}

Instructions:
1. Carefully read the article content and identify the main focus
2. Choose the most appropriate primary topic from the taxonomy above
3. Base your decision on the primary subject matter and content focus
4. Consider the description of each category when making your choice
5. If the article truly doesn't fit well into any of the listed categories, choose "Other"
6. Be specific and accurate - don't force a category if it's not a good fit
7. Respond ONLY with the classification in this exact format:

<classification>
Primary Topic: [exact primary topic name]
</classification>
\end{lstlisting}
\end{tcolorbox}
\caption{Prompt for classifying topic of articles}
\label{prompt:topics}
\end{figure*}

\paragraph{Topics present in datasets.}
The classification scheme in this work builds on the 17 top-level categories defined in the \href{https://iptc.org/standards/media-topics/}{International Press Telecommunications Council (IPTC) “Media Topics”} taxonomy. We map each article to one of the 17 top-level topics. In addition to those categories, we include two supplementary labels — “Obituary” and “Other” — to capture content that does not cleanly fall into the predefined classes. Definitions of topics are displayed in \autoref{tab:topic_defs} and the distribution of topics across the three datasets are shown in \autoref{tab:topics-comparison}.

\begin{table*}[!tb]
\centering
\small
\begin{tabularx}{\linewidth}{@{}l X@{}}
\toprule
\textbf{Topic} & \textbf{Definition (IPTC summary)} \\
\midrule
arts, culture, entertainment and media & All forms of arts, entertainment, cultural heritage and media \\
conflict, war and peace & Acts of socially or politically motivated protest or violence, military activities, geopolitical conflicts, as well as resolution efforts \\
crime, law and justice & The establishment and/or statement of the rules of behavior in society, the enforcement of these rules, breaches of the rules, the punishment of offenders and the organizations and bodies involved in these activities \\
disaster, accident and emergency incident & Man made or natural event resulting in loss of life or injury to living creatures and/or damage to inanimate objects or property \\
economy, business and finance & All matters concerning the planning, production and exchange of wealth. \\
education & All aspects of furthering knowledge, formally or informally \\
environment & The protection, damage, and condition of the ecosystem of the planet Earth and its surroundings \\
health & All aspects of physical and mental well-being \\
human interest & Item that discusses individuals, groups, animals, plants or other objects in an emotional way \\
labor & Social aspects, organizations, rules and conditions affecting the employment of human effort for the generation of wealth or provision of services and the economic support of the unemployed. \\
lifestyle and leisure & Activities undertaken for pleasure, relaxation or recreation outside paid employment, including eating and travel. \\
politics and government & Local, regional, national and international exercise of power, the day-to-day running of government, and the relationships between governing bodies and states. \\
religion & Belief systems, institutions and people who provide moral guidance to followers \\
science and technology & All aspects pertaining to human understanding of, as well as methodical study and research of natural, formal and social sciences, such as astronomy, linguistics or economics \\
society & The concerns, issues, affairs and institutions relevant to human social interactions, problems and welfare, such as poverty, human rights and family planning \\
sport & Competitive activity or skill that involves physical and/or mental effort and organizations and bodies involved in these activities \\
weather & The study, prediction and reporting of meteorological phenomena \\
\addlinespace
\multicolumn{2}{@{}l}{\textit{Additional labels used in our classification}}\\
Obituary & Memorial and death-notice content about individuals. \\
Other & Articles that do not clearly align with any IPTC top-level topic. \\
\bottomrule
\end{tabularx}
\caption{Top-level topics from the IPTC Media Topics taxonomy and definitions \cite{iptc2025mediatopics}.}
\label{tab:topic_defs}
\end{table*}

\begin{table*}[!htbp]
\centering
\small
\begin{tabular}{lrrr}
\hline
\textbf{Topic} & \textbf{\maindata (\%)} & \textbf{\opinionsdata(\%)} & \textbf{\reporterdata (\%)} \\
\hline
Politics and government               & 15.02 & 56.86 & 19.55 \\
Economy, business and finance         & 10.53 & 9.77  & 2.30 \\
Arts, culture, entertainment \& media & 20.55 & 4.87  & 6.55 \\
Opinion                               & 0.40  & 4.65  & 0.75 \\
Health                                & 3.75  & 4.47  & 12.92 \\
Human interest                        & 3.61  & 2.91  & 6.09 \\
Crime, law and justice                & 10.91 & 2.80  & 7.41 \\
Environment                           & 4.22  & 2.77  & 4.36 \\
Education                             & 3.72  & 2.06  & 10.27 \\
Science and technology                & 0.98  & 1.97  & 0.72 \\
Society                               & 1.29  & 1.86  & 3.71 \\
Conflict, war and peace               & 0.54  & 1.72  & 0.35 \\
Labor                                 & 1.54  & 0.96  & 1.45 \\
Religion                              & 0.74  & 0.94  & 0.69 \\
Sport                                 & 11.24 & 0.69  & 5.28 \\
Lifestyle and leisure                 & 1.70  & 0.42  & 0.50 \\
Disaster, accident, emergency         & 3.69  & 0.23  & 1.25 \\
Obituary                              & 2.48  & 0.04  & 1.49 \\
Weather                               & 0.76  & 0.01  & 0.37 \\
Other                                 & --    & --    & 14.00 \\
\hline
\end{tabular}
\caption{Comparison of topic distributions between \maindata, \opinionsdata, and \reporterdata.}
\label{tab:topics-comparison}
\end{table*}

\begin{figure}[htbp]
    \centering
    \includegraphics[width=0.9\linewidth]{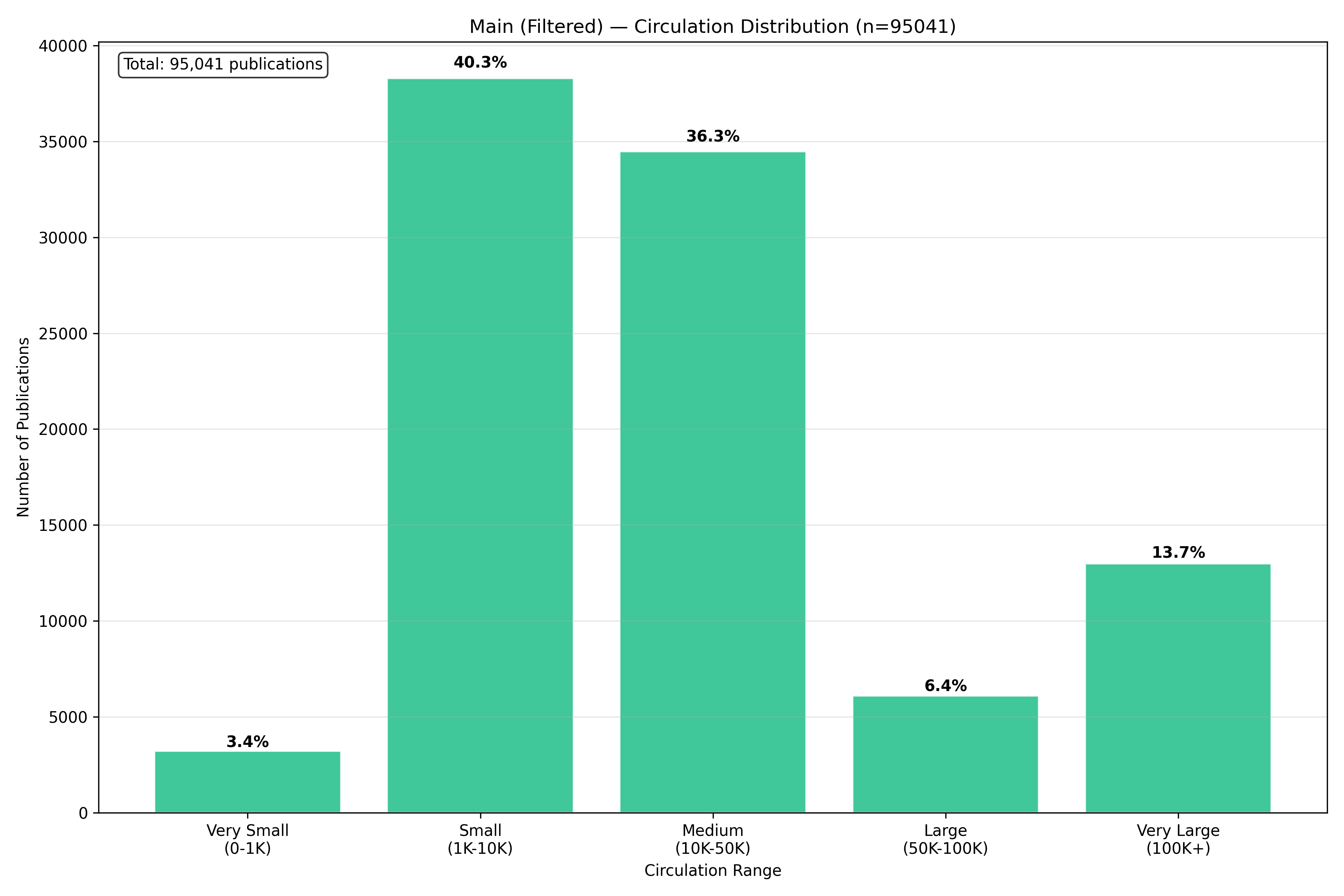}
    \caption{Distribution of circulations of articles in the \maindata dataset.}
    \label{fig:circulation_distribution}
\end{figure}

\paragraph{Circulation definition and data source.}
\label{app:circulation_definition}
In the \emph{U.S. News Deserts Database}, \textbf{circulation} refers exclusively to the average number of printed copies distributed per publishing day, excluding any digital readership metrics. These figures are compiled primarily from \emph{Alliance for Audited Media} (AAM) audits when available, and supplemented with publisher self-reports, state press association directories, and industry trade sources for unaudited outlets. As noted in the database’s section \emph{``Dealing with Circulation Limitations''} on \href{https://www.usnewsdeserts.com/methodology/}{https://www.usnewsdeserts.com/methodology/}, many small newspapers rely on self-reported or infrequently updated numbers, which are treated as approximate indicators of print scale rather than precise measures of audience size. The distribution of newspapers by circulation amounts is depicted in \autoref{fig:circulation_distribution}.

\paragraph{Data coverage and overlap.}
Our correlation analysis uses the subset of articles for which both AI-likelihood and circulation are observed. Out of \num{186507} articles in total print circulation is available for \num{101799} (\SI{54.58}{\percent}). At the outlet level, the main dataset contains \num{1560} unique newspapers, of which \num{776} (\SI{49.74}{\percent}) report any circulation figure. Unless otherwise noted, all circulation–AI likelihood correlations are computed on this overlapping set to avoid missing-data biases.

\paragraph{Ownership Details}
\label{app:ownershipdetails}
Our ownership information comes from the \href{https://localnewsinitiative.northwestern.edu/}{Northwestern Local News Initiative} \cite{medill2024ai_local}. We collect ownership information for 162,342 articles (87.04\% of \maindata) and 1,259 unique newspapers (82.40\% of unique newspapers in \maindata). Since ownership of newspapers is transferred frequently, this database shows ownership information current as of November 2025. 

\begin{figure}[htbp]
    \centering
    \includegraphics[width=0.9\linewidth]{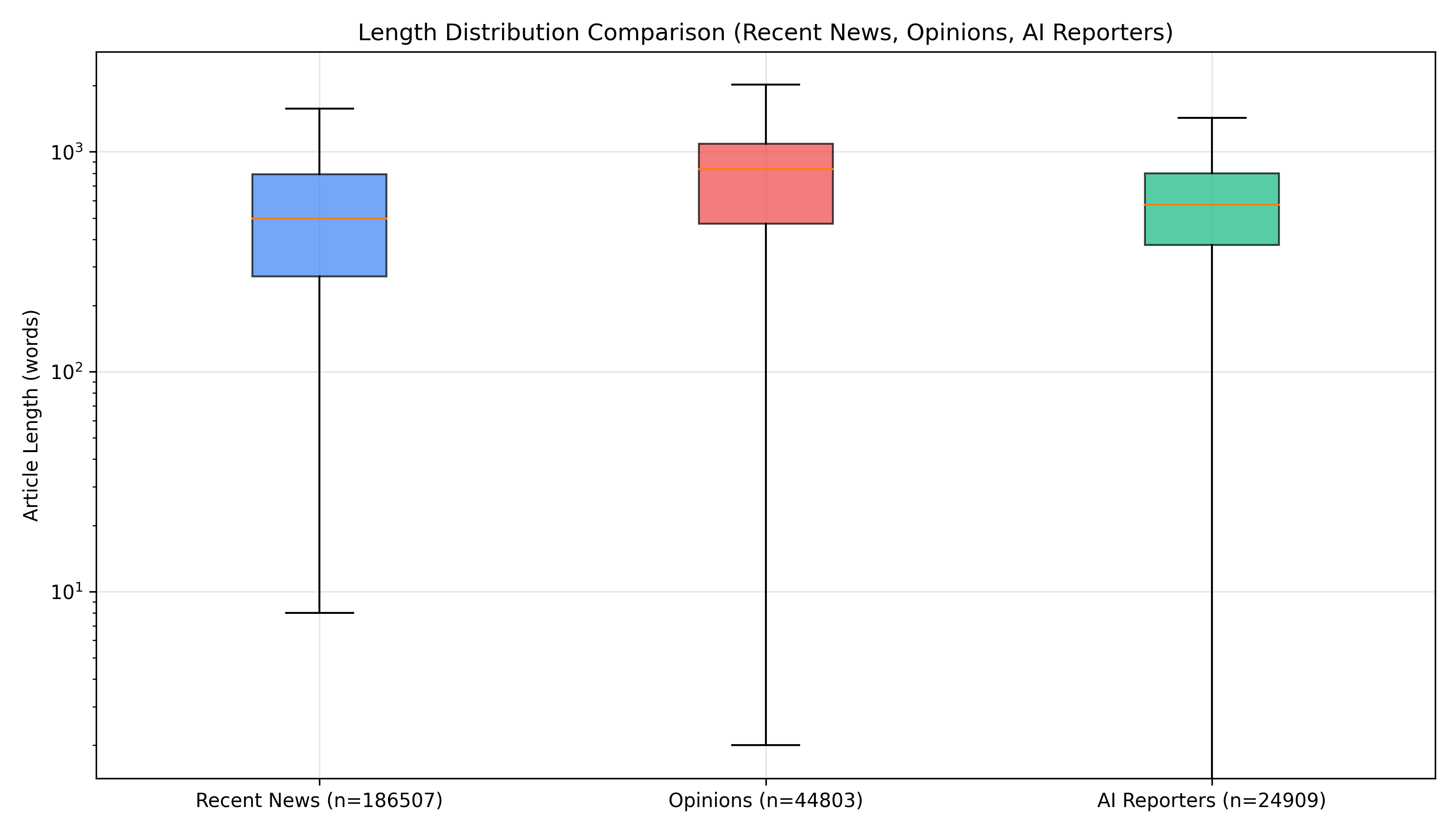}
    \caption{Comparison of the lengths of articles in \maindata, \opinionsdata, and 
    \reporterdata}
    \label{fig:data_lengths}
\end{figure}

\paragraph{Lengths of datasets. }
We see that all datasets are similar lengths, with \opinionsdata being slightly longer at an average of 1078.4 tokens, compared to recent news which has an average of 787.4 tokens. Comparison of lengths shown in \autoref{fig:data_lengths}.

\section{AI Detection}
\label{app:ai_detections}

In this section of the appendix we discuss details of how we used AI Detection models.

\paragraph{Pangram prediction API details. }
Pangram is a highly accurate AI detection language model \cite{emi2024technicalreportpangramaigenerated}. To detect AI, Pangram divides the text into segments, with each segment assigned an AI probability score before an overall confidence score and final label are produced We use Pangram v2 for all results in this study. However, examples may reflect newer versions of Pangram likelihood.

\paragraph{Condensing the Pangram API labeling scope.}
\label{app:condensing_labels}
To standardize categories across Pangram’s short-text (single pass) and long-text (sliding-window) endpoints, we collapse the vendor’s granular labels into three meta-labels used throughout our analysis: \labelhuman = \{\textit{Human}, \textit{Unlikely AI}\}; \labelmixed = \{\textit{Mixed}, \textit{Possibly AI}, \textit{Likely AI}\}; and \labelai = \{\textit{Highly Likely AI}, \textit{AI}\}. 

\paragraph{Comparison to GPTZero.}
We drew a balanced sample of 1{,}000 articles from Pangram: 500 labeled \emph{AI / Highly Likely AI} and 500 labeled \emph{Human / Unlikely AI}. We then ran GPTZero on the same texts; GPTZero returns \texttt{human}, \texttt{mixed}, or \texttt{ai}. For evaluation, we binarized the GPTZero outputs by counting \texttt{mixed} as AI and compared the two detectors on this binary task. Under this protocol, we observed 88.2\% raw agreement and Cohen’s $\kappa = 0.764$ (118/1{,}000 disagreements).
\[
\begin{array}{c|cc}
\text{Pangram}\backslash\text{GPTZero} & Human & AI \\\hline
Human & 490 & 10\\
AI & 108 & 392
\end{array}
\]

\section{Tracking reporter adoption of AI}
\label{sec:reporters}

 \begin{itemize}
     \item[\faUser] \reporterdata\ is a historical dataset of \textbf{20K} articles published by a subset of \textbf{10} veteran reporters who “authored” multiple AI articles in \maindata. Each reporter in this dataset published articles written both before and after the release of ChatGPT (November 2022), enabling longitudinal analysis.
 \end{itemize}

Additionally, we analyze \reporterdata, a longitudinal dataset for 10 veteran reporters (5 male and 5 female) who published articles both before and after ChatGPT's release (November 2022) and had at least three articles in \maindata\ labeled as either \labelai\ or \labelmixed.\footnote{All but one reporter are veteran reporters, often with decades of experience. However, due to technical difficulties associated with scraping specific websites, we could not obtain \emph{all} of their older articles. The one early-career reporter started their career in 2021, and thus had less exposure to the profession prior to the release of ChatGPT. The reporters collectively cover a variety of topics ranging from local government, public safety, and environmental justice to national politics, racial equity, LGBTQ+ rights, Caribbean-American culture, economic development, and community sports.} For each reporter, we scrape all of their available online articles and run the same detection pipeline detailed in \S\ref{sec:data_methodology} .
We note that since the reporters selected for this analysis have multiple articles flagged for AI use in \maindata, they are not representative of all reporters. Nevertheless, they give us valuable insight into some of the heaviest adopters of AI in modern American journalism. 

\paragraph{\reporterdata}
\label{app:reporterdataset}
Several reporters in \maindata\ published articles both before and after the release of ChatGPT, making them good candidates for a longitudinal analysis to explore when and how they started using AI. We identify a set of 10 veteran reporters from \maindata\ who meet two criteria: (1) they have published articles prior to November 2022,\footnote{This cutoff was selected because the public release of ChatGPT in November 2022 \cite{openai2022chatgpt} made AI-assisted writing tools widely accessible.} and (2) at least three of their articles were identified as \labelmixed\ or \labelai. We then write ten custom scrapers, one for each reporter, to collect a corpus of \emph{all} of their published articles that are available online, resulting in a final dataset of 20,132 articles from 14 newspapers.\footnote{One journalist published in multiple newspapers.}

\begin{figure}[t]
    \centering
    \includegraphics[width=\linewidth]{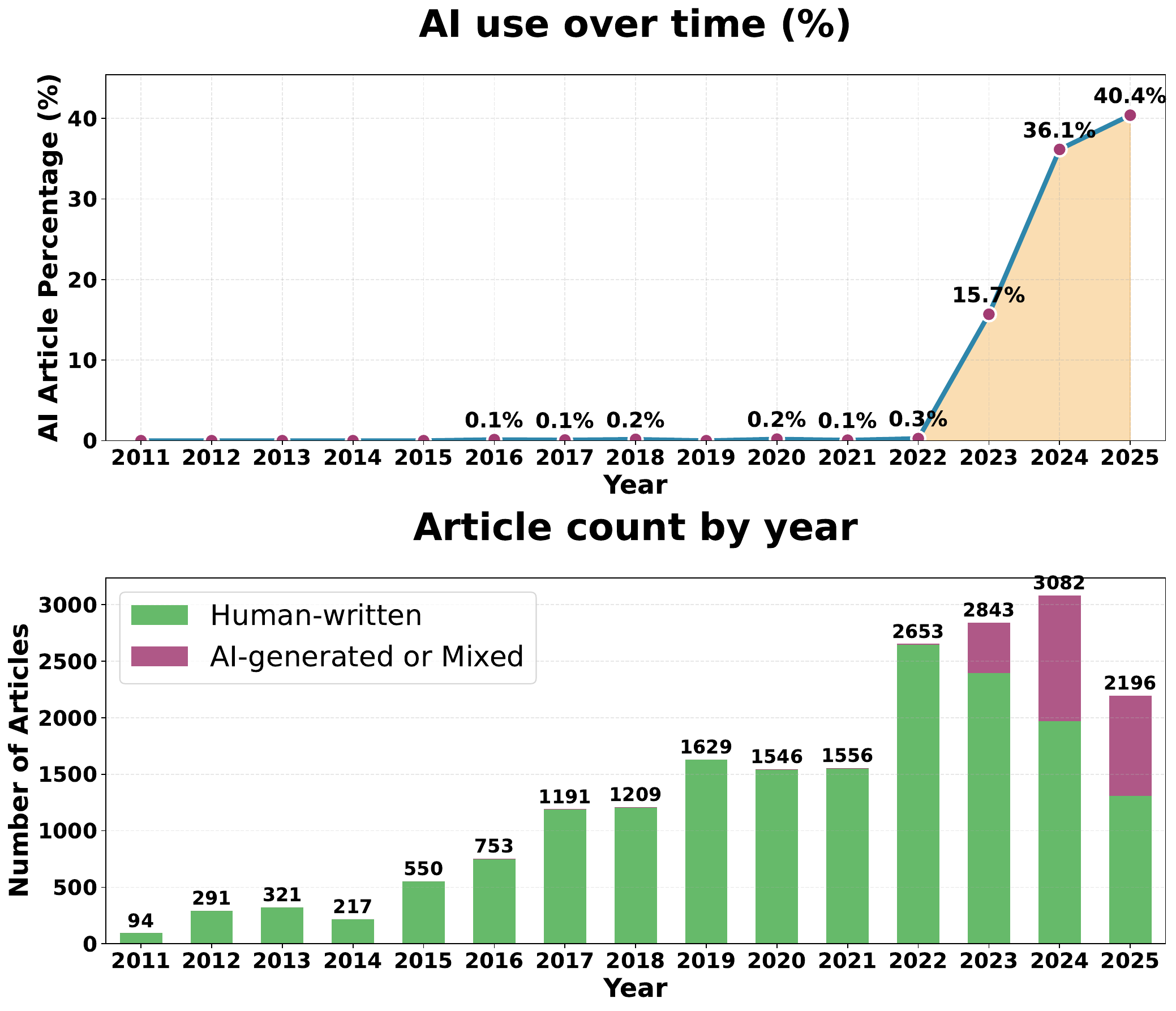}
    \caption{AI adoption takes off after ChatGPT's release in late 2022 with the ten reporters in our \reporterdata\ sample.}
    \label{fig:ai_reporters}
\end{figure}

\paragraph{AI use by these reporters rises from 0\% pre-ChatGPT to 40\% in 2025.} \autoref{fig:ai_reporters} shows a near-absence in AI use prior to November 2022 before rising sharply to 15.7\% in 2023, 36.1\% in 2024, and then 40.4\% in 2025. This result also serves as a sanity check on Pangram's reliability: as all of these reporters published articles pre-ChatGPT that were correctly detected as human-written, it is unlikely that their unique writing styles are a source of false positives.

\paragraph{While AI use varies across this cohort of reporters, none discloses their use.} We observe high variance in AI use across these ten reporters. While some show a negligible increase in articles flagged with AI use (from 0\% in 2022 to about 2.3\% in 2025), others are far more receptive. For instance, 90.1\% of the most prolific author's 2025 articles were classified as \labelmixed\ or \labelai, up from 0\% in 2022 (see individual plots in  \S\ref{app:ai_reporters}). Importantly, none of the ten reporters disclose use of AI to their readership.\footnote{We reached out to each reporter in this cohort with a publicly-available email address in an effort to learn more about their AI use, but (perhaps unsurprisingly) we received no responses.}

\paragraph{AI traces are visible in the articles.} Perhaps not surprisingly, we observe qualitative differences between the articles marked for AI use and the human-written ones. For instance, AI-assisted articles in this subset have up to 11.53× more em dashes than those authored entirely by humans (\autoref{tab:reporter-ai-shifts}, \faUser\ Reporter 3). Furthermore, human-written articles tend to be more concrete, naming specific people and locations (\faUser\ Reporter 1) and focus more on factual knowledge than vague statements (\faUser\ Reporter  2).

\begin{table*}[t]
\centering
\scriptsize
\setlength{\tabcolsep}{4pt}
\renewcommand{\arraystretch}{1}

\begin{tabularx}{\textwidth}{@{} l p{0.29\textwidth} p{0.29\textwidth} p{0.29\textwidth} @{}}
\toprule \textsc{Reporter} &  \labelhuman\ & \labelai\ or \labelmixed\ & \textsc{Observation} \\
\midrule

 \faUser\ \textbf{1} &
\emph{``Problems associated with the past California administration seem to have dissipated as the IID and Newsome government officials are on the verge of agreeing to a plan restoring the lower part of the Salton Sea as reported by Water Manager Tina Shields during the April 30 meeting. ...''}\par
{(2019-05-06)} \pangram{https://www.pangram.com/history/de8fe24e-43c5-43c7-871a-63d8ee005559} 
% v2: human, https://www.pangram.com/history/616b3d32-2230-4631-a168-b0233d3bf249/
% v3.2: SAME, https://www.pangram.com/history/de8fe24e-43c5-43c7-871a-63d8ee005559
\news{https://www.thedesertreview.com/news/a-new-state-administration-brings-hope-for-the-battered-sea/article_fcc09122-704c-11e9-b741-83e0895c15e0.html}&
\emph{``The State of California has established the Salton Sea Conservancy under the Salton Sea Conservancy Act, a \aiword{comprehensive} initiative to reverse decades of \aiword{ecological} damage and \aiword{promote sustainable} development in the Salton Sea region. This new state agency, housed within the Natural Resources Agency, \aiword{is poised to} lead the charge ..."}\par
{(2024-11-18)} \pangram{https://www.pangram.com/history/32c55d19-1e2e-4e39-830d-e3d591db5390} 
% v2: AI, https://www.pangram.com/history/7bc59327-9594-434f-861e-fa815039412e/
% v3.2: SAME, https://www.pangram.com/history/32c55d19-1e2e-4e39-830d-e3d591db5390
\news{https://www.thedesertreview.com/news/state/salton-sea-conservancy-act-a-new-chapter-for-environmental-restoration/article_033eda84-6b12-11ef-b766-6b091ca39cd3.html} &The human-written article names specific relevant people, times, and pinpoints the main point of the article, the restoration of the Salton Sea. In the AI article, the conservancy act is tied to lofty statements such as \textit{'decades of ecological damage'}. 
\\
\midrule

 \faUser\ \textbf{2} &
\emph{``One shouldn’t need a reason to buy Black, it should be a way of life. Such is the thought process for several Black business owners and individuals who hang their hat on the idea of supporting within to keep the Black dollar circulating longer in local neighborhoods, mom-and-pop shops, and places that need it most...''}  \par
{(2023-02-15)} \pangram{https://www.pangram.com/history/892a7e90-cd3b-4f69-9ce3-84c88a8575ed} 
% v2: human, https://www.pangram.com/history/561d3f37-662d-49df-9c5a-9a2e65a75318/
% v3.2: same, https://www.pangram.com/history/892a7e90-cd3b-4f69-9ce3-84c88a8575ed
\news{https://sacobserver.com/2023/02/the-power-of-the-black-dollar-is-still-strong/} &
\emph{``The spending power of the Black dollar stands at a staggering \$1.7 trillion, reflecting \aiword{immense potential} for community growth and economic empowerment. This financial strength, however, is underutilized, prompting a \aiword{crucial} need for increased and unconditional support within the African American community...''}  \par
{(2023-11-29)} \pangram{https://www.pangram.com/history/067cee50-f04e-4bb9-9597-ad00d341b39b} 
% v2: ai, https://www.pangram.com/history/5214e365-b51a-49eb-9ee8-1f7846803846/
% v3.2: same, https://www.pangram.com/history/067cee50-f04e-4bb9-9597-ad00d341b39b
\news{https://sacobserver.com/2023/11/property-is-power-the-power-of-the-black-dollar/} &
In the human-written article, the reporter uses unique phrases like 'hang their hat' and 'mom-and-pop shops', focusing on specific ways the reader could buy from Black businesses. In the latter, the ``author'' makes sweeping claims about the financial state of Black businesses in a more general manner. \\
\midrule

 \faUser\ \textbf{3} &
\emph{``The UN General Assembly met on Tuesday afternoon in Emergency Special Session on the decades long Israel-Palestine conflict and as the ongoing crisis in Gaza shows no signs of abating. Member States adopted a resolution, demanding an ``immediate humanitarian ceasefire”, the immediate and unconditional release of all hostages ...''}\par
{(2024-06-20)} \pangram{https://www.pangram.com/history/172dcbf9-fcca-435f-bd30-5f7b49c8ad40} 
% v2: human, https://www.pangram.com/history/60959095-de2d-4399-b43a-818bc7f46634/
% v3.2: same, https://www.pangram.com/history/172dcbf9-fcca-435f-bd30-5f7b49c8ad40
\news{https://nycaribnews.com/un-general-assembly-voted-for-immediate-ceasefire-in-gaza/}&
\emph{``The Government of Guyana has expressed its support for the United States’ call for a ceasefire \aiword{in the ongoing} conflict between Israel and Gaza. \aiword{In a recent statement}, the Guyanese government acknowledged the three-phase plan proposed by U.S. President Joseph Biden on May 31, 2024, aimed at ending Israel’s war on Gaza...''}\par
{ (2025-03-21)} \pangram{https://www.pangram.com/history/a036f169-78d0-4dbc-bff4-040f294c0247} 
% v2: ai, https://www.pangram.com/history/d8b18db4-bc91-40fb-8435-5d9a83705030/
% v3.2: ai assisted?, https://www.pangram.com/history/a036f169-78d0-4dbc-bff4-040f294c0247
\news{https://nycaribnews.com/guyana-backs-call-for-immediate-ceasefire-in-gaza/}&
In the human-written article, the author references the time (Tuesday) and setting (Emergency Special Setting). In the AI-generated article, the author uses more fluff words and more vague time placement (i.e. ongoing, recent). \\

\bottomrule
\end{tabularx}
\caption{Excerpts from passages of newspaper articles written by reporters in the \reporterdata\ dataset. Words and phrases identified as indicative of AI use by Pangram are highlighted in red. In the left \labelhuman\ column, excerpts of older, human-written articles are displayed, while the \labelai\ or \labelmixed\ column shows newer articles by the same author detected as AI-generated or assisted. When AI use is present, articles by these reporters include fewer specific details, broader time markers, and loftier language.}
\label{tab:reporter-ai-shifts}
\end{table*}

\section{Additional Results}

\subsection{\maindata}

In this section, we report more results about \maindata, a collection of over 185k articles collected between June 15th and September 15th, 2025.

\begin{figure}[htbp]
        \centering
        \includegraphics[width=0.95\linewidth]{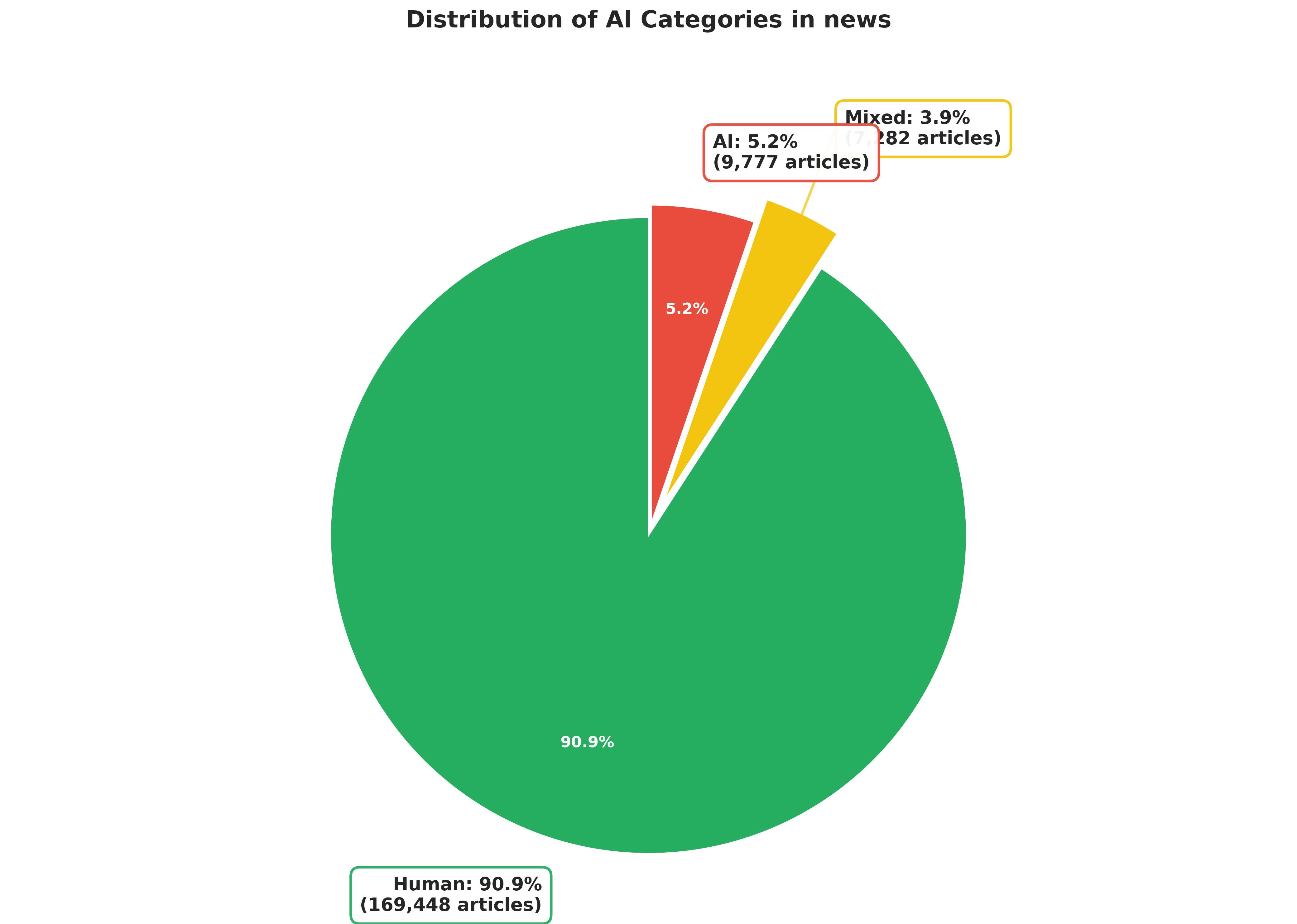}
        \caption{Distribution of AI Use predictions in \maindata.}
        \label{fig:recent_news_pie_chart}
    \end{figure}

\begin{figure}[htbp]
        \centering
        \includegraphics[width=0.95\linewidth]{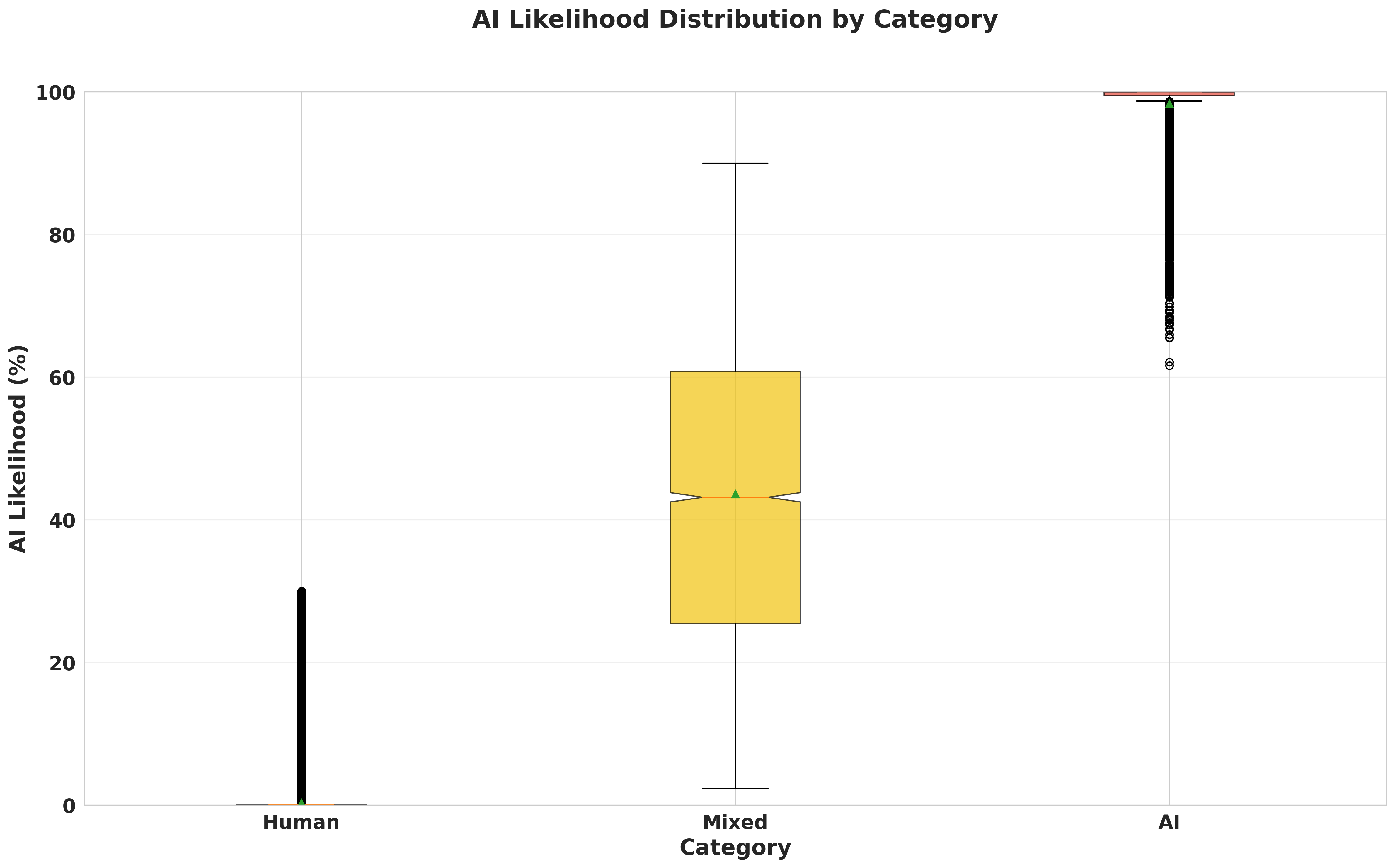}
        \caption{Distribution of AI Likelihoods per each AI Use category in \maindata. }
        \label{fig:recent_news_box_plot}
    \end{figure}

\begin{figure*}[t]
        \centering
        \includegraphics[width=0.95\linewidth]{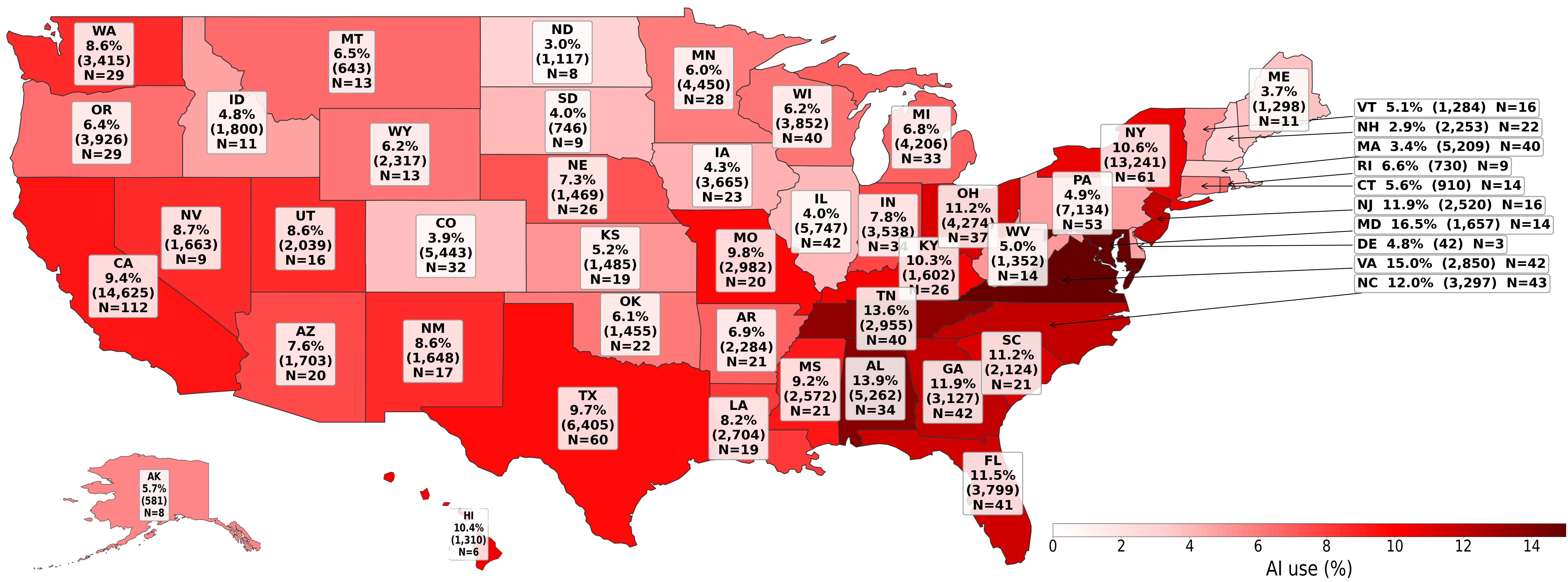}
        \caption{Map of the United States showing the AI use (\%), number of articles, and number of unique newspapers for each state.  States in the mid-Atlantic and southern US exhibit markedly higher AI use than other states. Note that this plot only considers articles written in English (see \autoref{fig:non_english_state_heatmap} for others). 
        }
        \label{fig:state_ai_heatmap}
    \end{figure*}

\begin{figure*}[t]
  \centering
  \includegraphics[width=0.95\linewidth]{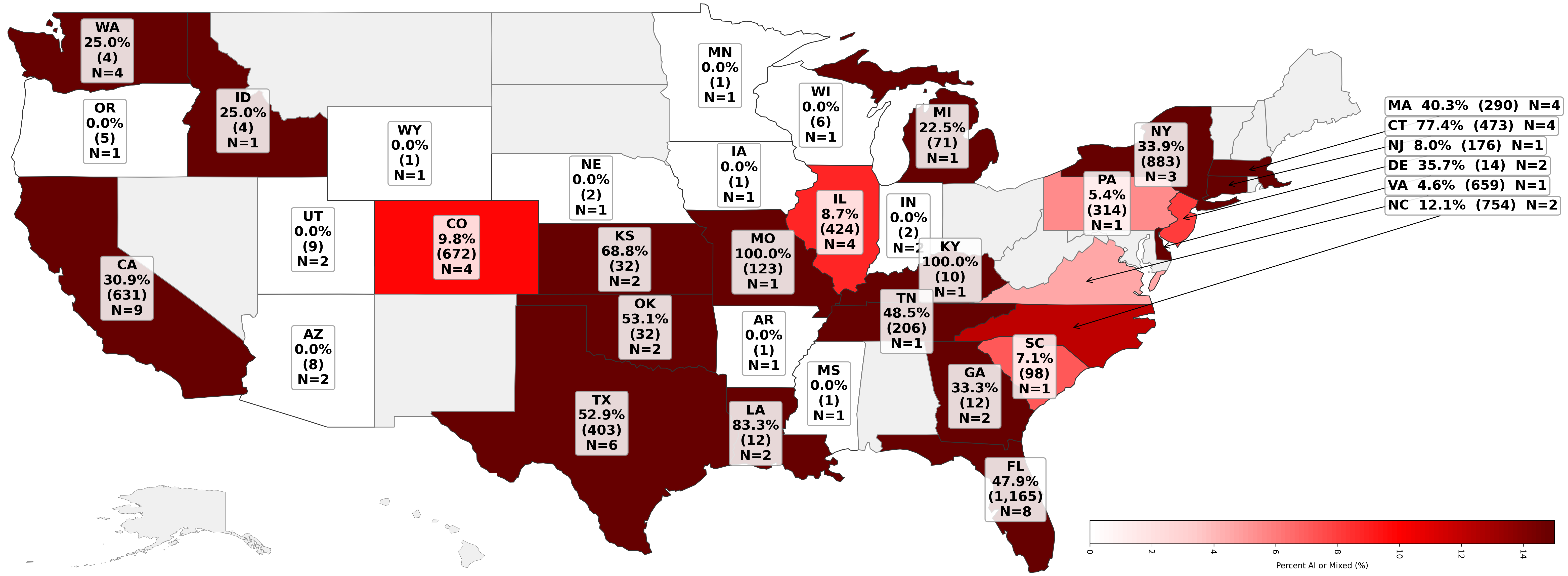}
  \caption{AI use in articles written in languages other than English. The most common language are Spanish, Portuguese, and Vietnamese.}
  \label{fig:non_english_state_heatmap}
\end{figure*}

\begin{figure}[t]
    \centering
    \includegraphics[width=0.95\linewidth]{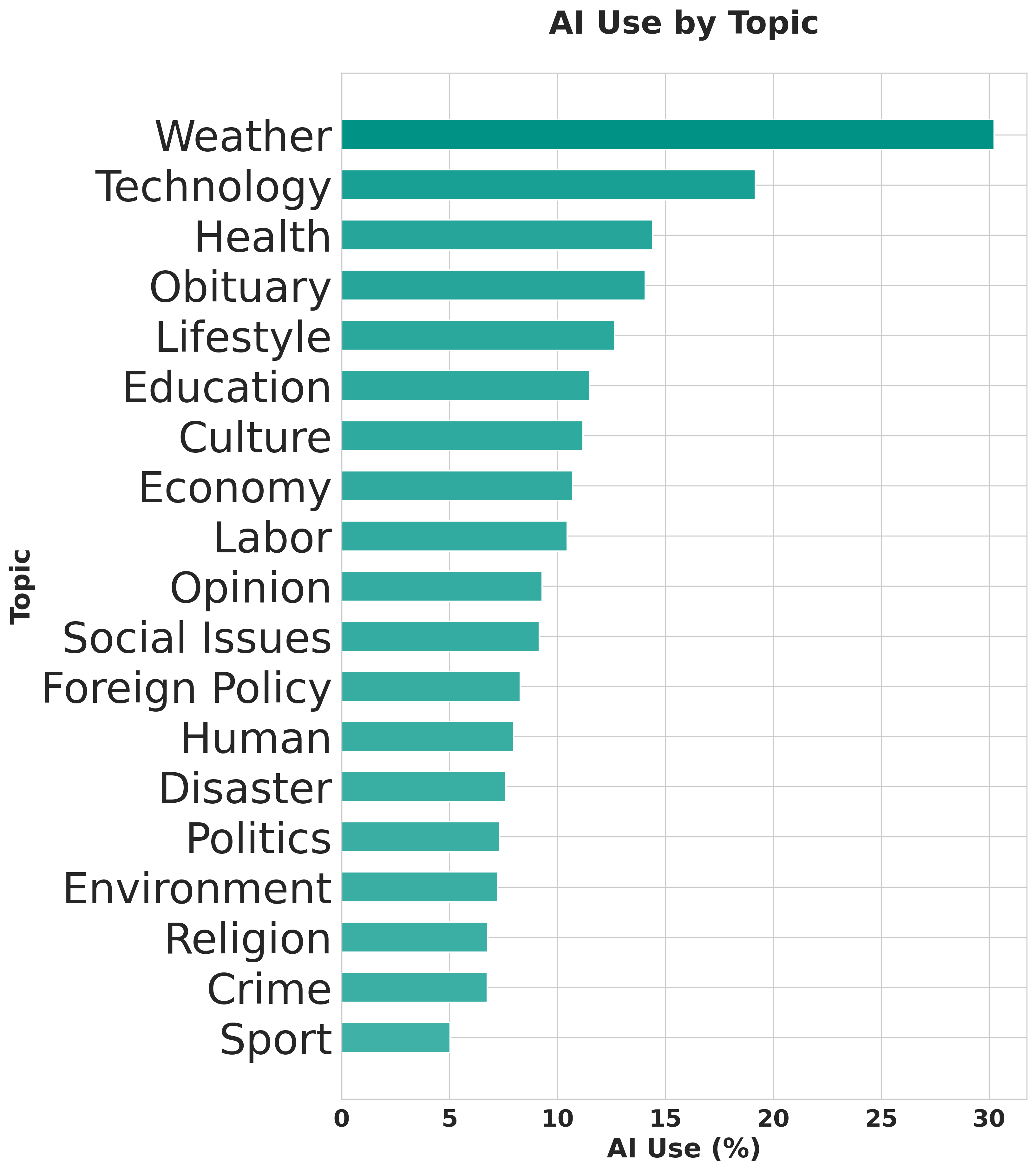}
    \caption{AI use by topic in \maindata. Weather, science/technology, and health exhibit higher AI use than topics like conflict, crime/justice, and religion. }
    \label{fig:topic_ai_likelihood}
\end{figure}

\paragraph{Overall AI usage in \maindata.} \maindata has 90.9\% \labelhuman\ articles, with a total of 9.1\% having AI Use. 5.2\% is detected as \labelai\ and 3.9\% is \labelmixed\, as shown in 
\autoref{fig:recent_news_pie_chart}. The distribution of AI likelihoods per each of the three labels is shown in \autoref{fig:recent_news_box_plot}. Data by state is displayed in \autoref{fig:state_ai_heatmap} and \autoref{fig:non_english_state_heatmap} Band by topic in \autoref{fig:topic_ai_likelihood}.

 \begin{figure*}[t]
        \centering
        \includegraphics[width=\linewidth]{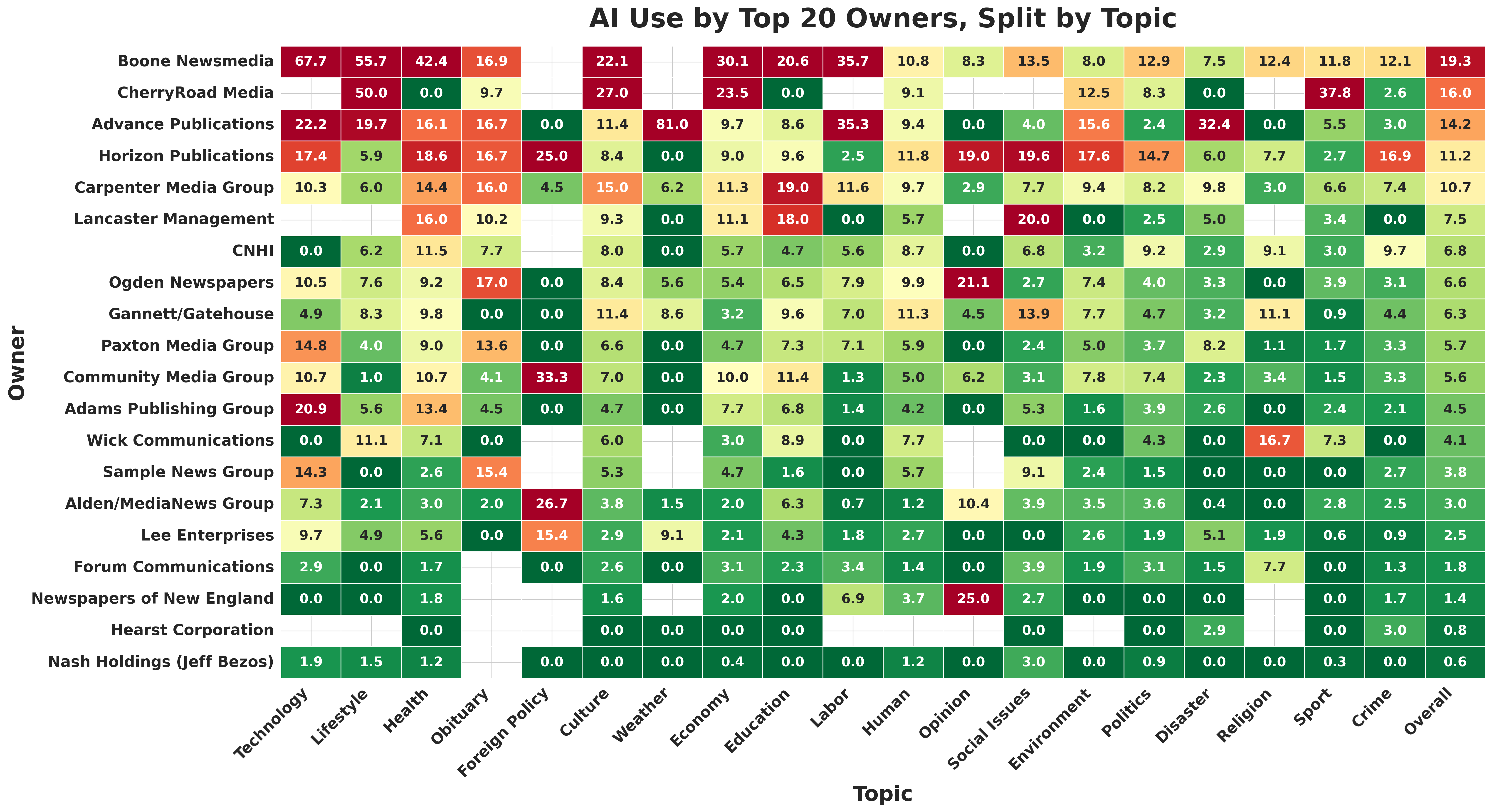}
        \caption{Heatmap of AI use by publication owner and article topic in \maindata. Some owners disclose AI use for specific content, such as \href{https://www.advance.com/}{Advance Publications} for weather reports, but the biggest adopters, such as \href{https://boonenewsmedia.com/ |}{Boone Newsmedia}, use AI broadly across many  topics.  Note that only 87\% of \maindata has ownership information, and only topics that have at least 5 articles per owner are visualized in the heatmap.}
        \label{fig:owner_topic_heatmap}
\end{figure*}

\begin{figure}[htbp]
        \centering
        \includegraphics[width=0.95\linewidth]{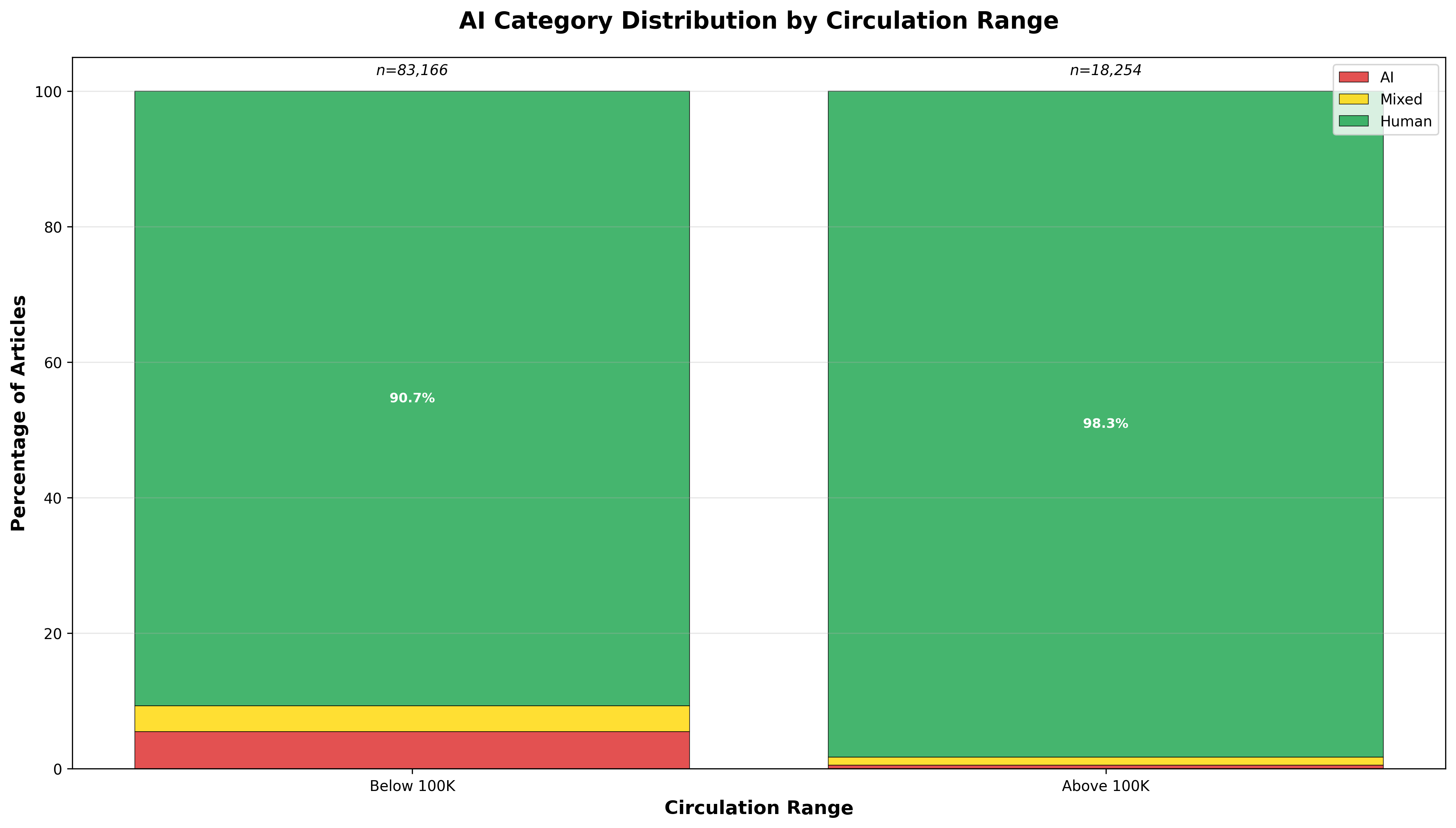}
        \caption{Distribution of AI Categories in \maindata articles between papers with circulations below 100k (left) and above 100k (right).}
        \label{fig:circulation_ai_likelihood}
    \end{figure}

\paragraph{Newspaper-level test (below 100K vs.\ 100K+ circulation).}
We collapse to one observation per outlet (share of AI-labeled articles) and compare newspapers below 100K circulation to those at 100K+. The below-100K group (n=750) averages 8.47\% AI articles (median 2.93\%, SD 16.08, range 0–100), while the 100K+ group (n=24) averages 4.95\% (median 1.75\%, SD 7.14, range 0–24.74). The difference in means is +3.52 percentage points (Cohen’s $d=0.22$). A Welch $t$-test indicates a statistically significant gap ($t=2.24$, $p=0.032$), whereas a Mann–Whitney $U$ test does not ($U=9506$, $p=0.634$). The divergence reflects heavy right-skew and zero-inflation in the outlet-level shares: mean-based tests are more sensitive to a few high-AI smaller outlets, while rank-based tests emphasize the bulk of the distribution. In other words, higher AI usage is \emph{not uniform} across smaller outlets; it is concentrated in a subset of them, while many smaller outlets have low (often zero) AI shares. Overall, we read this as modest evidence that smaller-circulation newspapers exhibit higher AI usage on average. In \autoref{fig:circulation_ai_likelihood}, the rate of newspapers labeled as \labelai or \labelmixed is much lower in papers with cirulation above 100k.

\begin{table}[t]
  \centering
  \small
  \caption{Language distribution}
  \label{tab:language-dist}
  \begin{tabular}{@{}lr@{}}
    \toprule
    Language & \% (n) \\
    \midrule
    English     & 95.16\% (177{,}478) \\
    Spanish     & 3.88\% (7{,}235) \\
    Portuguese  & 0.25\% (468) \\
    Vietnamese  & 0.22\% (403) \\
    French      & 0.18\% (343) \\
    Polish      & 0.17\% (314) \\
    \addlinespace[2pt]
    \textbf{Other} & \textbf{0.15\% (271)} \\
    \bottomrule
  \end{tabular}

  \medskip
  \footnotesize\emph{Other includes: Russian (163), German (37), Yiddish (27), Indonesian (25), Turkish (10), Dutch (5), Latin (2), Ukrainian (1), Chinese (1). Total: 186{,}512 articles.}
\end{table}

\paragraph{AI use in print.}
Many \labelai\ articles in \maindata\ appear in print in addition to the online editions.
According to a 2025 Pew study, the print audience still includes an estimated 65 million Americans who read physical newspapers
``often'' or ``sometimes'' \cite{pew2025news-print-read}. Critically, this group is heavily skewed towards an older demographic less likely to be technologically savvy, with a median age of 57.9~\cite{conaghan2017median}. We further note that readers of print newspapers face higher barriers to verifying content, in contrast to digital readers who can simply copy and paste text into search engines. Thus, newspapers that \emph{only} appear in print are much harder to audit using our methodology.

\paragraph{Detecting Hallucinated Content}
We manually reviewed 200 articles: 100 articles predicted as \labelai and 100 predicted as \labelhuman. 41\% of \labelai and 5\% of \labelhuman contained factual inaccuracies refered to as hallucinations. Some examples of hallucinations found in \labelai articles are depicted in \autoref{tab:hallucination_cases}.

\begin{table*}[!htb]
\centering
\scriptsize
\scalebox{1}{
\begin{tabular}{
    p{0.10\textwidth}  
    % p{0.12\textwidth}  
    p{0.39\textwidth}  
    p{0.39\textwidth}   
}

\toprule
\multicolumn{1}{c}{\bf\textsc{Newspaper}} & 
% \multicolumn{1}{c}{\bf\textsc{Prediction}} & 
\multicolumn{1}{c}{\bf\textsc{Article Excerpt}} &
\multicolumn{1}{c}{\bf\textsc{Hallucination}} \\
\midrule

LA SENTINEL &
% GPTZero (2026-03-30-base): AI_ONLY (ai=0.797, mixed=0.203, human=0.000) - AGREES with Pangram
\textit{``We know that 52 percent of adult women across this country are on Medicaid, and that number is exponentially higher in Los Angeles''}  \news{https://lasentinel.net/kamlager-dove-l-a-leaders-warn-defunding-planned-parenthood-would-spark-public-health-crisis.html}
& The percent of woman on medicaid is 19\%, not 52\% (\href{https://ccf.georgetown.edu/2025/06/24/women-depend-on-medicaid-across-the-lifespan/}{Source}). The article also contains other hallucinations such as referring to 'former President Trump' in 2025, when he is the current U.S. president. \\

\midrule

SANTA FE REPORTER &
% \labelai &
% GPTZero (2026-03-30-base): HUMAN_ONLY (ai=0.220, mixed=0.000, human=0.780) 
\textit{``The program is currently set to expire in June 2030 under the reconciliation bill, though advocates continue pushing for longer extensions and inclusion of additional affected communities.''}  \news{https://sfreporter.com/news/new-mexico-radiation-victims-could-soon-receive-long-awaited/}
& The program is set to expire in 2027, not 2030 as reported in the article (\href{https://www.justice.gov/civil/reca}{Source}).\\

\midrule

SUNCOAST NEWS &
% \labelmixed &
% GPTZero (2026-03-30-base): MIXED (ai=0.000, mixed=0.927, human=0.073) - AGREES with Pangram
\textit{``”We’re not trying to silence anyone,” Koulianos said. “We’re trying to conduct the people’s business in a reasonable timeframe.” ‘’ }
\news{https://www.suncoastnews.com/news/tarpon-springs-commission-reduces-time-for-public-comment/article_ce0fa2bf-d7e0-4ab5-961e-18383095c7ec.html} & Most of the quotes in this article, including the excerpt here, cannot be found in the transcript of the Board of Commissioners meeting (\href{https://www.youtube.com/watch?v=3dLZMYMEtE8\&list=PLZqIRUhB7O-ZfDStEaVV0KsyrdoTb1r1o}{Source}). Most quotes are paraphrases or similar quotes, but do not have the same wording.  \\

\midrule

THE ADVOCATE-MESSENGER&
% \labelmixed &
% GPTZero (2026-03-30-base): AI_ONLY (ai=0.617, mixed=0.118, human=0.265) - AGREES with Pangram
\textit{``It still shows up, five days a week in print and daily online, to deliver what no outside media outlet ever could: the truth of this town, in this moment, for these people.'' }
\news{https://amnews.com/2025/06/24/still-standing-still-serving-160-years-of-the-advocate-messenger/} & In an article detailing the history of their own paper, they incorrectly state the paper is printed 5 times a week, when in reality it is printed twice a week. (\href{https://web.archive.org/web/20200413222457/https://www.amnews.com/2020/04/10/newspaper-invests-more-in-online-coverage-as-covid-19-economic-crisis-worsens/}{Source}). \\

\bottomrule
\end{tabular}}
\caption{Examples of hallucinations found in articles predicted as \labelai.}
\label{tab:hallucination_cases}
\end{table*}

\paragraph{Detecting AI in quoted speech}
\label{app:quotes}

We analyze 30,462 articles with available quote-level predictions. Each quote is evaluated with an AI-likelihood classifier, restricted to spans of at least 50 tokens to reduce false positives on very short fragments. We treat quotes labeled as \emph{Unlikely AI} or \emph{Human} as human-written, and all others (Highly Likely AI, AI, Likely AI, Mixed, Possibly AI) as AI-generated. At the article level, we compare the incidence of AI quotes in AI-flagged versus human-flagged articles, yielding $P(\text{AI quote}|\text{AI article}) = 0.239$ and $P(\text{AI quote}|\text{human article}) = 0.034$. Conversely, the probability that an article is AI-generated given the presence of an AI quote is $P(\text{AI article}|\text{AI quote}) = 0.293$. In total, 1,376 articles contain at least one AI-generated quote, and 403 articles contain both AI-generated narrative and AI-generated quotes. To test whether length influences detection, we also compare long quotes ($\geq 120$ words, $n=870$) to short quotes, finding no significant difference in AI likelihood (two-sample $t$-test, $p = 0.96$). 

\paragraph{Many AI-generated articles contain authentic quotes.}  
To examine whether articles labeled as \labelmixed\ and \labelai\ include fabricated information, we analyze the authorship of quotations in these articles. Specifically, we extract all quotes in the dataset longer than 50 words, 
and run each of them through Pangram individually. Note that Pangram's reliability degrades on shorter texts, and so we are unable to perform this analysis on all quotes. Within the subset of articles that include at least one quote $>$50 words long, 76.1\% of articles flagged with AI use contain at least one human-written quote. This suggests that many stories written with AI use rely on authentically sourced material. However, it remains unclear whether journalists are choosing these quotes and feeding them into a prompt for AI generation, or if the AI is also doing quote selection (see \autoref{tab:notable_cases} for examples).

\begin{figure}[htbp]
  \centering
  \includegraphics[width=0.9\linewidth]{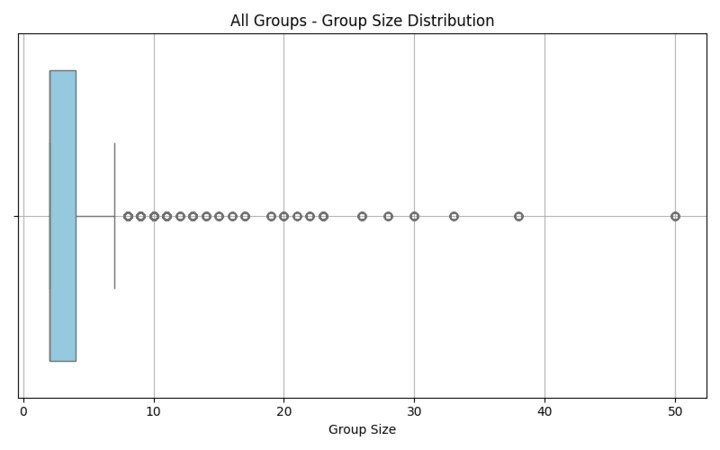}
  \caption{Distribution of duplicate-cluster sizes in \maindata. Most clusters are small, with a minority of large clusters driving redistribution volume.}
  \label{fig:group_boxplot}
\end{figure}

\paragraph{Effects of republishing articles.}
\label{app:republishing}
We examine how AI-generated content propagates via redistribution. Using \maindata, we cluster articles by cosine similarity and define \emph{exact duplicates} for scores $\geq 0.95$ and \emph{semantically similar} for $0.85\leq\text{score}<0.95$. In total, \num{16580} articles meet these criteria, forming \num{6413} clusters (mean size $=\num{4.16}$). The \textit{Associated Press} appears most frequently, in \num{1664} clusters and \num{4030} redistributed articles, consistent with syndication as the dominant driver of redistribution. Across all duplicates, \SI{6.7}{\percent} are labeled AI-generated, with a lower rate for exact duplicates (\SI{3.1}{\percent}) and a markedly higher rate for semantically similar duplicates (\SI{14.3}{\percent}). Figure~\ref{fig:group_boxplot} summarize the resulting cluster-size distribution; Table~\ref{tab:top10-authors-duplicates} lists the top content providers present in duplicate coverage.

\begin{table}[htbp]
\centering
\scriptsize
\begin{tabular}{r l r}
\hline
Rank & Author & Count\\
\hline
1  & Assoicated Press                        & 4,011 \\
2  & Unknown           & 1,204 \\
3  & Staff Reports        & 701 \\
4  & Advance Local Weather Alerts       &  404 \\
5  & USA Today Network           & 264 \\
6  & Myedmondnews             &  104 \\
7  & Teresa Wippel       & 101 \\
8  & WP Block Co-Authors                & 98 \\
9  & Cascade PBS Staff            & 82 \\
10 & Grace Gilson             & 81 \\
\hline
\end{tabular}
\caption{Top 10 authors by frequency of appearances in duplicate articles.}
\label{tab:top10-authors-duplicates}
\end{table}

\paragraph{Effects of machine translation on AI-text detection.}
\label{app:translation_details}
As depicted in \autoref{fig:eng_vs_other_langs}, AI-use is more frequent in languages other than English. To better understand if this is due to Pangram picking up on texts that were automatically translated into other languages from English or written in-part or fully using AI, we take a sample of 300 English detects with the following distribution: 100 \labelhuman, 100 \labelmixed, and 100 \labelai. The articles are translated using GPT4.1 \cite{openai2024gpt41} in chunks of 2000 tokens. The translation prompt is displayed in \autoref{prompt:translate}. If the article is less than 2000 tokens, the whole article is translated at once. If it is longer, the article is chunked into chunks of 2000 tokens or less, respecting paragraph boundaries, and then translated. We translate the articles into the 12 languages: Arabic, Chinese (Simplified), Dutch, French, German, Italian, Japanese, Polish, Portuguese, Russian, Spanish, and Vietnamese. After translation, we find the average AI-likelihood decreases by an average of 0.13, with Japanese having the smallest decrease of 0.08 (\autoref{tab:translation_binary}). We see that agreement using Pangram is high but imperfect, ranging between 74.7\% (Vietnamese) to 88.7\% (Portuguese). All languages have a McNemar P lower than 0.01 except Japanese, indicating that translation suppresses some AI-use. We note that this experiment uses a single MT model (GPT-4.1); results may differ with other translation methods or models, and the effects of human translation on AI detection remain an open question.

\begin{table*}[htbp]
\centering
\small
\begin{tabular}{lccccc}
\toprule
Language & 
$\Delta$ AI Likelihood &
Binary Match (\%) &
FP &
FN &
McNemar $p$ \\
\midrule
Arabic            & $-0.12$ & 82.0 & 11 & 43 & $<\!0.001$ \\
Chinese (Simpl.)  & $-0.12$ & 82.0 & 14 & 37 & $0.002$ \\
Dutch             & $-0.17$ & 82.0 &  4 & 50 & $<\!0.001$ \\
French            & $-0.11$ & 83.0 &  9 & 35 & $<\!0.001$ \\
German            & $-0.16$ & 82.3 &  5 & 48 & $<\!0.001$ \\
Italian           & $-0.13$ & 83.3 &  4 & 36 & $<\!0.001$ \\
Japanese          & $-0.08$ & 82.3 & 28 & 25 & $0.78$ \\
Polish            & $-0.19$ & 82.0 &  9 & 49 & $<\!0.001$ \\
Portuguese        & $-0.10$ & 88.7 &  7 & 27 & $0.001$ \\
Russian           & $-0.14$ & 82.3 &  6 & 45 & $<\!0.001$ \\
Spanish           & $-0.09$ & 88.0 &  4 & 32 & $<\!0.001$ \\
Vietnamese        & $-0.21$ & 74.7 & 10 & 66 & $<\!0.001$ \\
\midrule
\textbf{Mean}     & \textbf{$-0.13$} & \textbf{83.2} &  &  &  \\
\bottomrule
\end{tabular}
\caption{
Binary effects of machine translation on AI-use detection.
$\Delta$ AI likelihood denotes the mean change in predicted AI likelihood (translated minus English).
Binary match reports agreement with English labels under a human vs.\ AI-use formulation.
FP and FN denote false positives and false negatives relative to English.
McNemar’s exact test evaluates directional asymmetry in disagreements (FDR-corrected).
}
\label{tab:translation_binary}
\end{table*}

\begin{figure}[!tb]
\centering
\begin{tcolorbox}[colback=gray!5!white, colframe=teal, title=Prompt for translating langauge of articles]
\lstset{
    basicstyle=\ttfamily\footnotesize,
    breaklines=true,
    frame=none,
    xleftmargin=0pt,
    framexleftmargin=0pt,
    columns=fullflexible,
    tabsize=1,
    breakindent=0pt,
    breakautoindent=false,
    postbreak=\space,
    showstringspaces=false,
}
\begin{lstlisting}[language=Markdown]
Translate the following English text to {TARGET LANGUAGE}. 
Preserve the original meaning, tone, and style as much as possible.
Only return the translated text, without any explanations or additional comments.

Text to translate:
{TEXT}
\end{lstlisting}
\end{tcolorbox}
\caption{Prompt for translating articles to other languages}
\label{prompt:translate}
\end{figure}

\paragraph{Topic-level variation is unlikely to be driven by detector bias alone.}
One concern is that the uneven distribution of AI labels across topics could reflect differences in detector performance rather than differences in AI adoption. We use \opinionsdata\ to test whether topic-level uniformity differs in 2022, before widespread LLM-assisted writing, versus 2025. If detector bias were the main driver of topic-level differences, we would expect to observe similar topic variance in both years. In 2022, across 4{,}344 articles, only 5 were flagged (0.12\%). A permutation chi-squared test with 10{,}000 samples yielded a non-significant result ($p = 0.1101$), indicating no significant topic-level divergence. In 2025, across 7{,}932 articles, 267 were flagged (3.37\%). The same permutation chi-squared test was significant ($p = 0.0002$), showing strong topic-level divergence. We use a permutation chi-squared test because expected flagged counts in most topics are near zero in 2022, violating the large-sample assumptions of the standard asymptotic test. Taken together, topic uniformity in 2022 and topic divergence in 2025 support the interpretation that the detector is capturing real differential AI adoption across topics rather than fixed topic-specific detector artifacts.

\subsection{\opinionsdata}

\begin{figure}[htbp]
        \centering
        \includegraphics[width=0.95\linewidth]{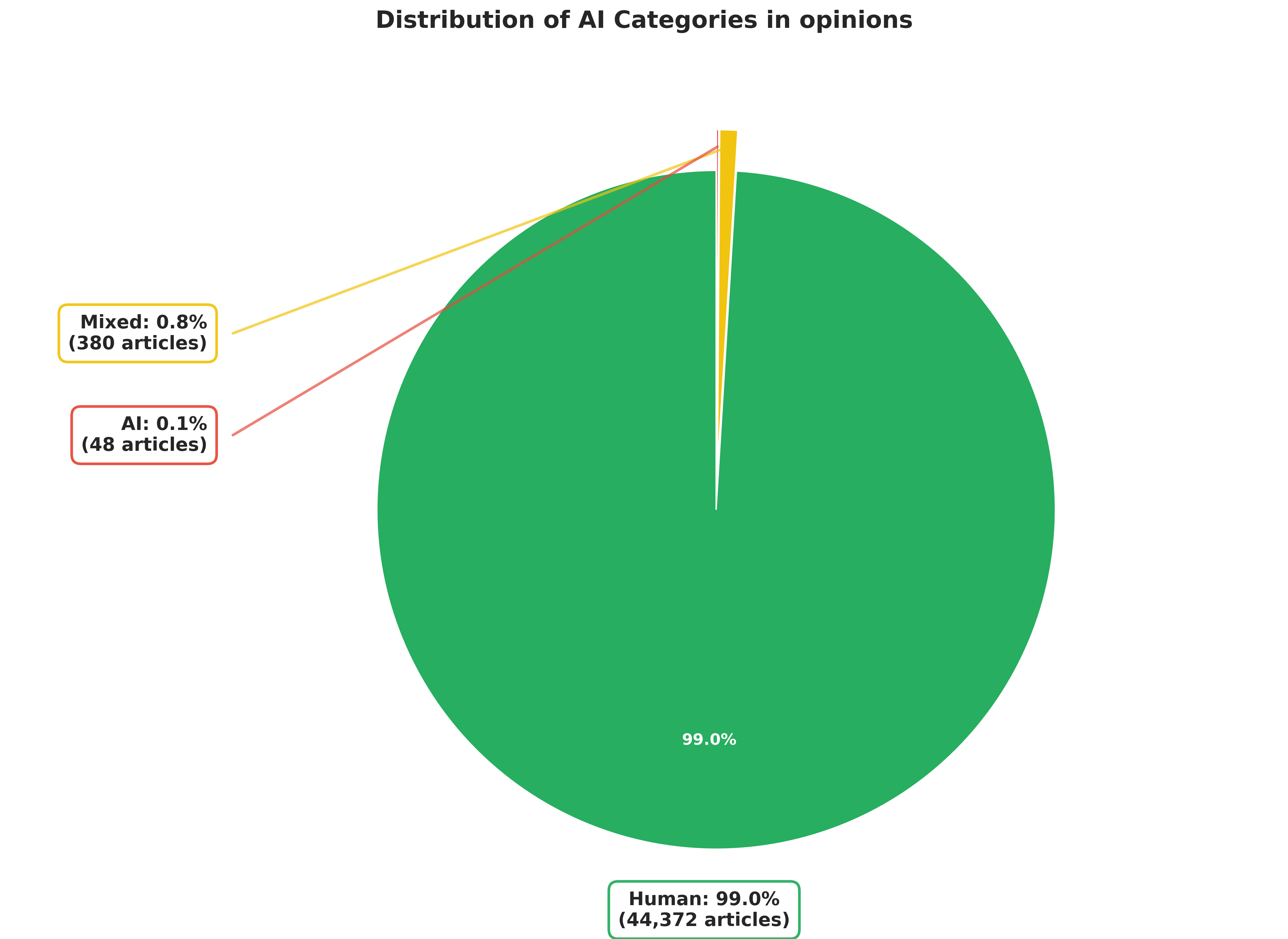}
        \caption{Distribution of AI Use predictions in \opinionsdata.}
        \label{fig:opinions_pie_chart}
    \end{figure}

\begin{figure}[htbp]
        \centering
        \includegraphics[width=0.95\linewidth]{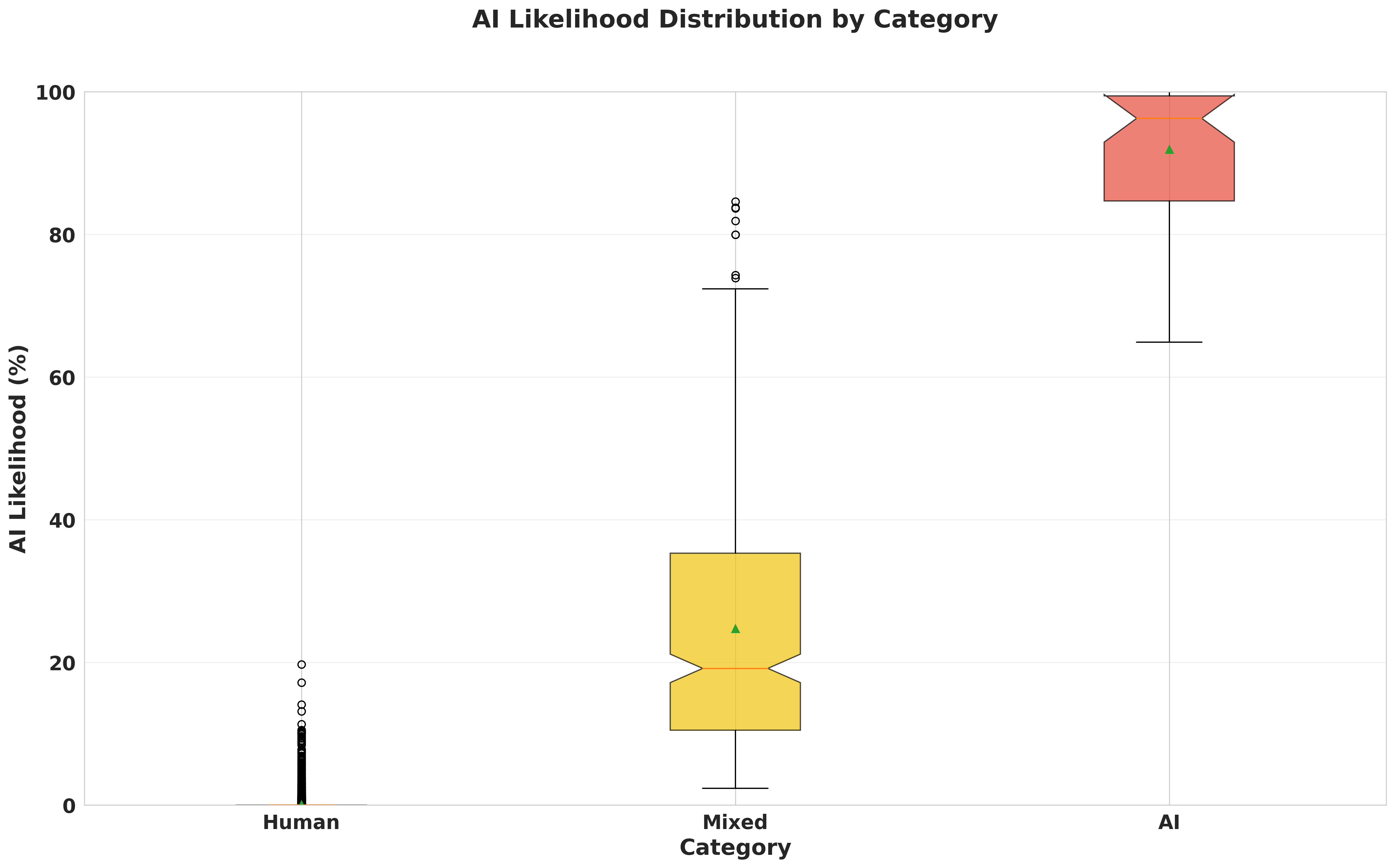}
        \caption{Distribution of AI Likelihoods per each AI Use category in \opinionsdata. }
        \label{fig:opinions_box_plot}
    \end{figure}

\paragraph{Overall AI usage in \opinionsdata.} 
\opinionsdata has 99\% \labelhuman\ articles, with a total of 1\% having AI Use. 0.11\% is detected as \labelai\ and 0.85\% is \labelmixed\, as shown in 
\autoref{fig:opinions_pie_chart}. The distribution of AI likelihoods per each of the three labels is shown in \autoref{fig:opinions_box_plot}.

\begin{table*}[t]
\centering
\small
\begin{tabular}{lccc}
\toprule
Outlet & Opinions (AI{+}mixed) & \maindata (AI{+}mixed) & Ratio \\
\midrule
WSJ  & 4.99\% & 0.74\% & 6.8$\times$ \\
WaPo & 5.51\% & 0.55\% & 10.1$\times$ \\
NYT  & 2.94\% & 1.80\% & 1.6$\times$ \\
\midrule
\textbf{Pooled} & \textbf{4.56\%} & \textbf{0.71\%} & \textbf{6.4$\times$} \\
\bottomrule
\end{tabular}
\caption{Opinion vs.\ \maindata AI rates, 2025-06-01–2025-09-15. }
\label{tab:opinion_vs_main_ai_rates}
\end{table*}

\begin{table*}[htbp]
\centering
\footnotesize
\setlength{\tabcolsep}{3pt}
\renewcommand{\arraystretch}{1}
\begin{tabularx}{\textwidth}{@{} p{0.10\textwidth} p{0.15\textwidth}  p{0.25\textwidth} Y @{}}
\toprule
\textit{Publication} & \textit{Label} & \textit{Author role} & \textit{Title} \\
\midrule

\multicolumn{4}{l}{\textbf{U.S. Politics \& Governance}} \\
\cmidrule(lr){1-4}

\textit{WaPo}  & \labelai  & Political writer and former congressional speechwriter & MAGA Maoism is spreading through the populist right. \pangram{https://www.pangram.com/history/ff4de517-d7cb-445d-a3f9-4e2e0f77473c} \news{https://www.washingtonpost.com/opinions/2025/04/08/maga-maoism-tariffs-trump/} \\
% old/V2 label: AI; new/V3 label: AI
% v3.2 dashboard: https://www.pangram.com/history/ff4de517-d7cb-445d-a3f9-4e2e0f77473c
% GPTZero (2026-03-30-base): AI_ONLY (ai=0.899, mixed=0.101, human=0.000) - AGREES with Pangram

\textit{WaPo} & \labelmixed & Former U.S. Senator & My fellow Republicans, the responsibility to speak out rests with you. \pangram{https://www.pangram.com/history/119b96aa-2707-4502-a967-1a355b5f2b7a} \news{https://www.washingtonpost.com/opinions/2025/05/29/jeff-flake-trump-republican-congress/} \\
% old/V2 label: Mixed; new/V3 label: Mixed
% v3.2 dashboard: https://www.pangram.com/history/119b96aa-2707-4502-a967-1a355b5f2b7a
% GPTZero (2026-03-30-base): HUMAN_ONLY (ai=0.070, mixed=0.128, human=0.802) 

\textit{WaPo} & \labelmixed & Urban warfare scholar & As the border wall goes up, the underground threat will grow. \pangram{https://www.pangram.com/history/8620b841-594a-4e9d-9976-2c8f784d7ca2} \news{https://www.washingtonpost.com/opinions/2025/01/22/us-mexico-border-wall-tunneling/} \\
% old/V2 label: Mixed; new/V3 label: Mixed
% v3.2 dashboard: https://www.pangram.com/history/8620b841-594a-4e9d-9976-2c8f784d7ca2
% GPTZero (2026-03-30-base): HUMAN_ONLY (ai=0.425, mixed=0.072, human=0.502)

\textit{WSJ} & \labelmixed & Legal activist and think tank fellow & The Gates Foundation Put Its Tax Exemption at Risk. \pangram{https://www.pangram.com/history/f7c36d54-535b-4970-8feb-b0ecd39db413} \news{https://www.wsj.com/opinion/the-gates-foundation-puts-its-tax-exemption-at-risk-race-based-discrimination-scholarship-0235d189} \\
% old/V2 label: Mixed; new/V3 label: Mixed
% v3.2 dashboard: https://www.pangram.com/history/f7c36d54-535b-4970-8feb-b0ecd39db413
% GPTZero (2026-03-30-base): HUMAN_ONLY (ai=0.015, mixed=0.066, human=0.919) 

\addlinespace[2pt]
\multicolumn{4}{l}{\textbf{Public Health}} \\
\cmidrule(lr){1-4}

\textit{NYT}  & \labelmixed & Immunologist & America Is Abandoning One of the Greatest Medical Breakthroughs. \pangram{https://www.pangram.com/history/595e38bd-2df1-4744-8350-c9b73955b464} \news{https://www.nytimes.com/2025/08/18/opinion/mrna-vaccines.html}\\
% v2: mixed, https://www.pangram.com/history/f4222362-36c2-4cb8-b193-c8de437d7b57/
% v3.2: mixed, https://www.pangram.com/history/595e38bd-2df1-4744-8350-c9b73955b464
% GPTZero (2026-03-30-base): MIXED (ai=0.038, mixed=0.742, human=0.220) - AGREES with Pangram

\textit{WSJ}  & \labelmixed & Health-longevity company founder & A Chatbot Can Be Part of Your Medical Team. \pangram{https://www.pangram.com/history/e9ea6aef-8a6c-44e9-80a6-70570bd42a62} \news{https://www.wsj.com/opinion/a-chatbot-can-be-part-of-your-medical-team-9dc11ed7}\\
% old/V2 label: Mixed; new/V3 label: Mixed
% v3.2 dashboard: https://www.pangram.com/history/e9ea6aef-8a6c-44e9-80a6-70570bd42a62
% GPTZero (2026-03-30-base): HUMAN_ONLY (ai=0.154, mixed=0.212, human=0.634) - DISAGREES with Pangram (Mixed)

\textit{WSJ}  & \labelai & Physician, Fox News commentator & The Madness in RFK Jr.'s Autism Method \pangram{https://www.pangram.com/history/028cb47d-06cc-45c2-8363-b08cea3bc486} \news{https://www.wsj.com/opinion/the-madness-in-rfk-jr-s-autism-method-hhs-secretary-science-health-2087ffe8} \\
% v2: ai, https://www.pangram.com/history/012d30db-f2b8-4e07-9367-2fa044c1aa6f/
% v3.2: ai, https://www.pangram.com/history/028cb47d-06cc-45c2-8363-b08cea3bc486
% GPTZero (2026-03-30-base): HUMAN_ONLY (ai=0.001, mixed=0.061, human=0.938) 

\addlinespace[2pt]
\multicolumn{4}{l}{\textbf{War \& National Security}} \\
\cmidrule(lr){1-4}

\textit{WSJ}  & \labelai  & Former Canadian international trade minister & Canada Is the Best Friend America's Got. \pangram{https://www.pangram.com/history/5a91e9b7-6a50-4974-936f-34bd141489b4} \news{https://www.wsj.com/opinion/canada-is-the-best-friend-americas-got-trade-border-jobs-9ebb6600} \\
% old/V2 label: AI; new/V3 label: AI
% v3.2 dashboard: https://www.pangram.com/history/5a91e9b7-6a50-4974-936f-34bd141489b4
% GPTZero (2026-03-30-base): AI_ONLY (ai=1.000, mixed=0.000, human=0.000) - AGREES with Pangram

\textit{NYT}  & \labelai    & Retired U.S. Army general & Be Not Afraid. \pangram{https://www.pangram.com/history/8c7dca0e-450a-46d3-96c2-2e69c0100db5} \news{https://www.nytimes.com/2025/04/13/opinion/mcchrystal-fear-america.html}\\
% v2: ai, https://www.pangram.com/history/189228c0-22fb-4ca1-957b-9603f3f9eae7/
% v3.2: ai, https://www.pangram.com/history/8c7dca0e-450a-46d3-96c2-2e69c0100db5
% GPTZero (2026-03-30-base): AI_ONLY (ai=0.800, mixed=0.200, human=0.000) - AGREES with Pangram

\textit{WaPo}  & \labelmixed & Space research CEO & Nuclear-powered rockets will win the new space race. \pangram{https://www.pangram.com/history/13a0b26d-6250-4802-99a8-81a1145e116d} \news{https://www.washingtonpost.com/opinions/2025/08/20/nasa-nuclear-power-rockets-space/} \\
% old/V2 label: Mixed; new/V3 label: Mixed
% v3.2 dashboard: https://www.pangram.com/history/13a0b26d-6250-4802-99a8-81a1145e116d
% GPTZero (2026-03-30-base): HUMAN_ONLY (ai=0.077, mixed=0.000, human=0.923) 

\addlinespace[2pt]
\multicolumn{4}{l}{\textbf{Cybersecurity}} \\
\cmidrule(lr){1-4}

\textit{WaPo} & \labelmixed & U.S. Representative \& CEO of TFG Asset Management & Cyber warfare has arrived. Here’s the United States’ best defense. \pangram{https://www.pangram.com/history/c033bc92-e64f-4045-a984-fa3a30fbd9ab} \news{https://www.washingtonpost.com/opinions/2025/07/22/cyberwar-ai-us-tech-adademy/} \\
% v2: mixed, https://www.pangram.com/history/6be37047-0913-47b2-bfc9-ebd87554d5d9/
% v3.2: mixed, https://www.pangram.com/history/c033bc92-e64f-4045-a984-fa3a30fbd9ab
% GPTZero (2026-03-30-base): HUMAN_ONLY (ai=0.011, mixed=0.000, human=0.989) 

\textit{WaPo}  & \labelmixed & AI data-center CEO & AI extremists are peddling science fiction. \pangram{https://www.pangram.com/history/672bcf63-8efc-4563-a519-91ddbd87906e} \news{https://www.washingtonpost.com/opinions/2025/09/12/ai-realism-tool-doomers-zealots/} \\
% old/V2 label: Mixed; new/V3 label: Mixed
% v3.2 dashboard: https://www.pangram.com/history/672bcf63-8efc-4563-a519-91ddbd87906e
% GPTZero (2026-03-30-base): HUMAN_ONLY (ai=0.142, mixed=0.000, human=0.858) 

\textit{NYT}  & \labelmixed & Principal Product Manager & Ye and the Limits of Free Speech \pangram{https://www.pangram.com/history/2c5db03d-c8ba-46ca-b238-eec43b574210/} \news{https://www.nytimes.com/2025/02/19/opinion/kanye-west-x-antisemitism.html} \\
% v2: mixed, https://www.pangram.com/history/30d37985-9d95-4cbf-92fc-7cc4d88af728/
% v3.2: human, https://www.pangram.com/history/30d37985-9d95-4cbf-92fc-7cc4d88af728
% GPTZero (2026-03-30-base): HUMAN_ONLY (ai=0.052, mixed=0.140, human=0.808) - AGREES with Pangram v3.2 (Human)

\addlinespace[2pt]
\multicolumn{4}{l}{\textbf{Technology}} \\
\cmidrule(lr){1-4}

\textit{WSJ}  & \labelmixed    & Former Intel CEO & A Sovereign-Wealth Fund to Keep America's Technological Edge; \pangram{https://www.pangram.com/history/9f62ef3c-974b-4755-b914-6fc6fc4ebbef} \news{https://www.wsj.com/opinion/a-sovereign-wealth-fund-to-keep-americas-technological-edge-ai-china-4760dc09}\\
% v2: mixed, https://www.pangram.com/history/359b77eb-ea04-40cb-8618-9900ba5c53e9/
% v3.2: mixed, https://www.pangram.com/history/9f62ef3c-974b-4755-b914-6fc6fc4ebbef
% GPTZero (2026-03-30-base): HUMAN_ONLY (ai=0.023, mixed=0.025, human=0.952)

\textit{NYT} & \labelmixed & Therapist & I’m a Therapist. ChatGPT Is Eerily Effective. \pangram{https://www.pangram.com/history/ab4415d3-1547-455e-a72b-d660a80c7a76} \news{https://www.nytimes.com/2025/08/01/opinion/chatgpt-therapist-journal-ai.html} \\
% old/V2 label: Mixed; new/V3 label: Mixed
% v3.2 dashboard: https://www.pangram.com/history/ab4415d3-1547-455e-a72b-d660a80c7a76
% GPTZero (2026-03-30-base): MIXED (ai=0.083, mixed=0.793, human=0.124) - AGREES with Pangram

\textit{WSJ}  & \labelmixed & Investment firm founder & The AI Boom May Be Too Good to Be True. \pangram{https://www.pangram.com/history/5fa71aea-0a0e-458e-8694-dfbae8cdc5bd} \news{https://www.wsj.com/opinion/the-ai-boom-may-be-too-good-to-be-true-copyright-ip-lawsuits-could-derail-econ-potential-fd514ea3} \\
% old/V2 label: Mixed; new/V3 label: Mixed
% v3.2 dashboard: https://www.pangram.com/history/5fa71aea-0a0e-458e-8694-dfbae8cdc5bd
% GPTZero (2026-03-30-base): HUMAN_ONLY (ai=0.133, mixed=0.000, human=0.867) 

\addlinespace[2pt]
\multicolumn{4}{l}{\textbf{Law \& Justice}} \\
\cmidrule(lr){1-4}

\textit{NYT}  & \labelai    & Journalist, ex–Navy SEAL & What the ‘Rust’ Shooting Case Is Really About. \pangram{https://www.pangram.com/history/04cf7775-06c9-4f0c-bacc-07cfaaa744ec} \news{https://www.nytimes.com/2024/03/13/opinion/rust-hollywood-gun-safety.html} \\
% v2: ai, https://www.pangram.com/history/2e178c61-8dd8-4db5-bf95-f93b564bf05a/
% v3.2: ai, https://www.pangram.com/history/04cf7775-06c9-4f0c-bacc-07cfaaa744ec
% GPTZero (2026-03-30-base): AI_ONLY (ai=0.886, mixed=0.114, human=0.000) - AGREES with Pangram

\textit{WSJ}  & \labelai & Governor and sociology professor & Utah Hands Parents the Keys to the App Store. \pangram{https://www.pangram.com/history/3d2a45f4-d091-4a08-a60e-f785621b2fc7} \news{https://www.wsj.com/opinion/utah-hands-parents-the-keys-to-the-app-store-content-ratings-harm-social-media-youth-ff52106a} \\
% old/V2 label: AI; new/V3 label: AI
% v3.2 dashboard: https://www.pangram.com/history/3d2a45f4-d091-4a08-a60e-f785621b2fc7
% GPTZero (2026-03-30-base): MIXED (ai=0.040, mixed=0.960, human=0.000) - AGREES with Pangram (AI-involved)

\bottomrule
\end{tabularx}
\caption{A sample of op-eds labeled as either \labelmixed or \labelai from WSJ, WaPo, and NYT. Author names are omitted and replaced with professional roles. These op-eds often address polarizing topics such as politics, war, and public health, making the disclosure of AI imperative. All examples included produced the same labels on Pangram v2 and v3.2.}
\label{tab:opinion-ai-controversial}
\end{table*}

\subsection{\reporterdata}
\label{app:ai_reporters}

In section \S\ref{sec:reporters}, we talk about the aggregated trends for the 10 reporters identified as using AI in our data set, whose data we analyzed longitudinally. Here we present more details about the reporters, label distribution, and patterns observed in individual reporters.

\paragraph{Reporter Profile.} \autoref{fig:ai_reporters} shows reporter statistics. We have anonymized the data, as we want to show only the diversity of expertise the reporters in \reporterdata have rather than highlight specific reporters. Nine out of ten reporters in this data had at least 10 years of experience. 

\begin{table*}[t]
\centering
\footnotesize
\resizebox{\linewidth}{!}{%
\begin{tabular}{@{\hskip 4pt}llllp{8cm}@{\hskip 4pt}}
\toprule
\textsc{Reporter} & \textsc{Years active} & \textsc{AI use in 2025 (\%)} & \textsc{Topic coverage} \\
\midrule
\faUser\ 1  & 25+ years        & 90.1\%\textsubscript{519/576}   & National politics, civil rights, racial justice, Black affairs \\[2pt]
\faUser\ 2  & 20+ years        & 84.6\%\textsubscript{11/13}     & Regional news, agriculture, water policy, environment \\[2pt]
\faUser\ 3  & 10+ years        & 78.3\%\textsubscript{90/115}    & Caribbean-American news, culture, health, diaspora \\[2pt]
\faUser\ 4  & 40+ years        & 72.7\%\textsubscript{8/11}      & Opinion on policy, education, creativity, technology \\[2pt]
\faUser\ 5  & 20+ years        & 38.1\%\textsubscript{48/126}    & Local news, public safety, city government, environment \\[2pt]
\faUser\ 6  & 30+ years        & 30.4\%\textsubscript{157/517}   & Environmental justice, racial equity, misinformation (opinion) \\[2pt]
\faUser\ 7  & 50+ years        & 11.7\%\textsubscript{19/162}    & Hyper-local news, history, community affairs \\[2pt]
\faUser\ 8  & 10+ years        & 6.9\%\textsubscript{27/393}     & LGBTQ+ news, politics, legal issues, culture \\[2pt]
\faUser\ 9  & \textless5 years & 3.3\%\textsubscript{5/153}      & Business, Black entrepreneurship, economic development \\[2pt]
\faUser\ 10 & \textasciitilde10 years & 2.3\%\textsubscript{3/130}     & Local sports, high school athletics, youth leagues \\
\bottomrule
\end{tabular}%
}
\caption{Reporter Profiles: Years active, aggregate AI use in 2025 (\%\textsubscript{$ai\_flag/total$}), and topic coverage. Reporters are arranged from most to least prolific AI user.}
\label{tab:reporters_profiles}
\end{table*}

\begin{figure}[t]
        \centering
        \includegraphics[width=\linewidth]{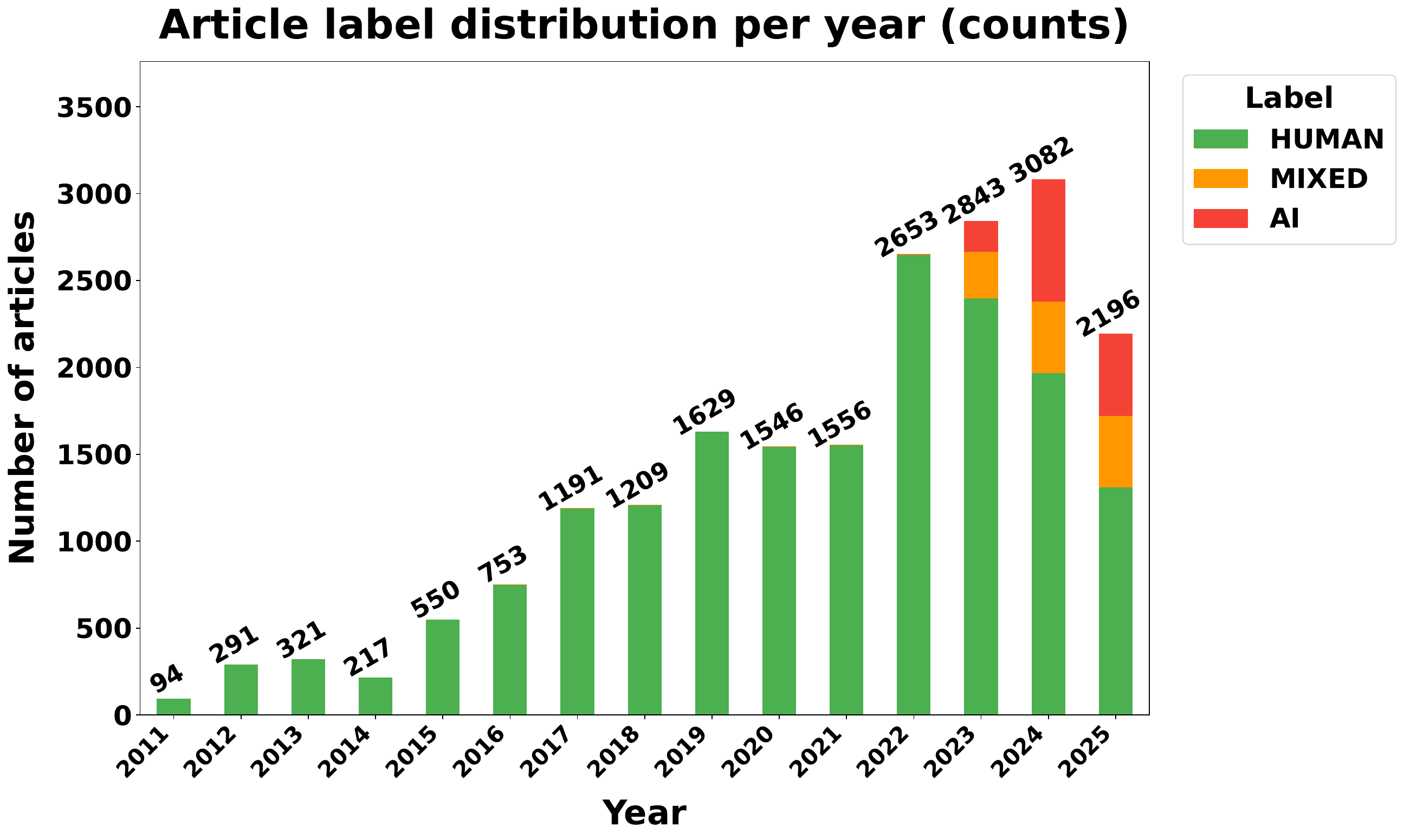}
        \caption{Distribution of AI use predictions in \reporterdata\ by year.}
        \label{fig:reporter_label_distr_year}
    \end{figure}

\begin{figure}[t]
        \centering
        \includegraphics[width=0.95\linewidth]{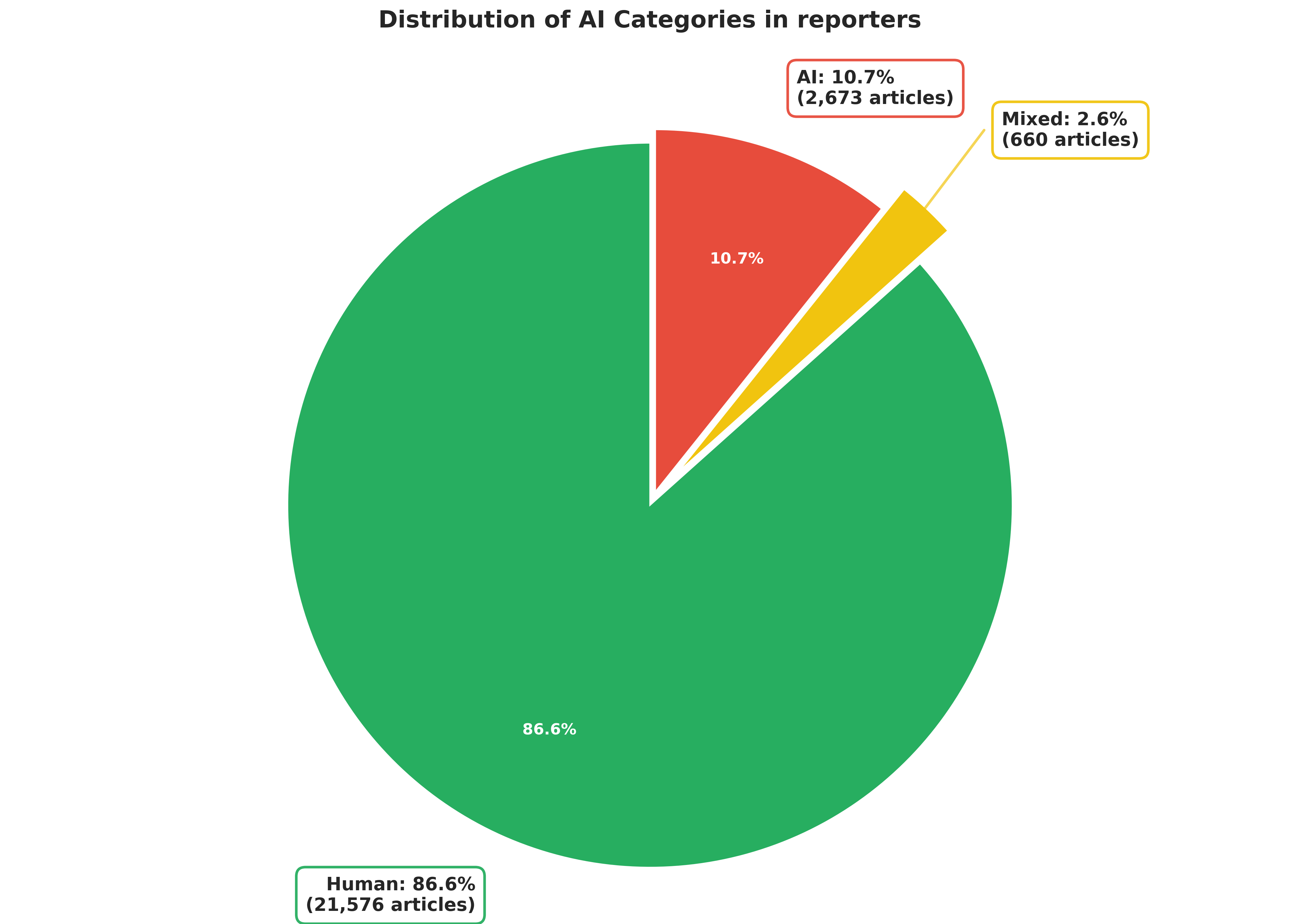}
        \caption{Distribution of AI Use predictions in \reporterdata.}
        \label{fig:reporter_pie_chart}
    \end{figure}

\begin{figure}[t]
        \centering
        \includegraphics[width=0.95\linewidth]{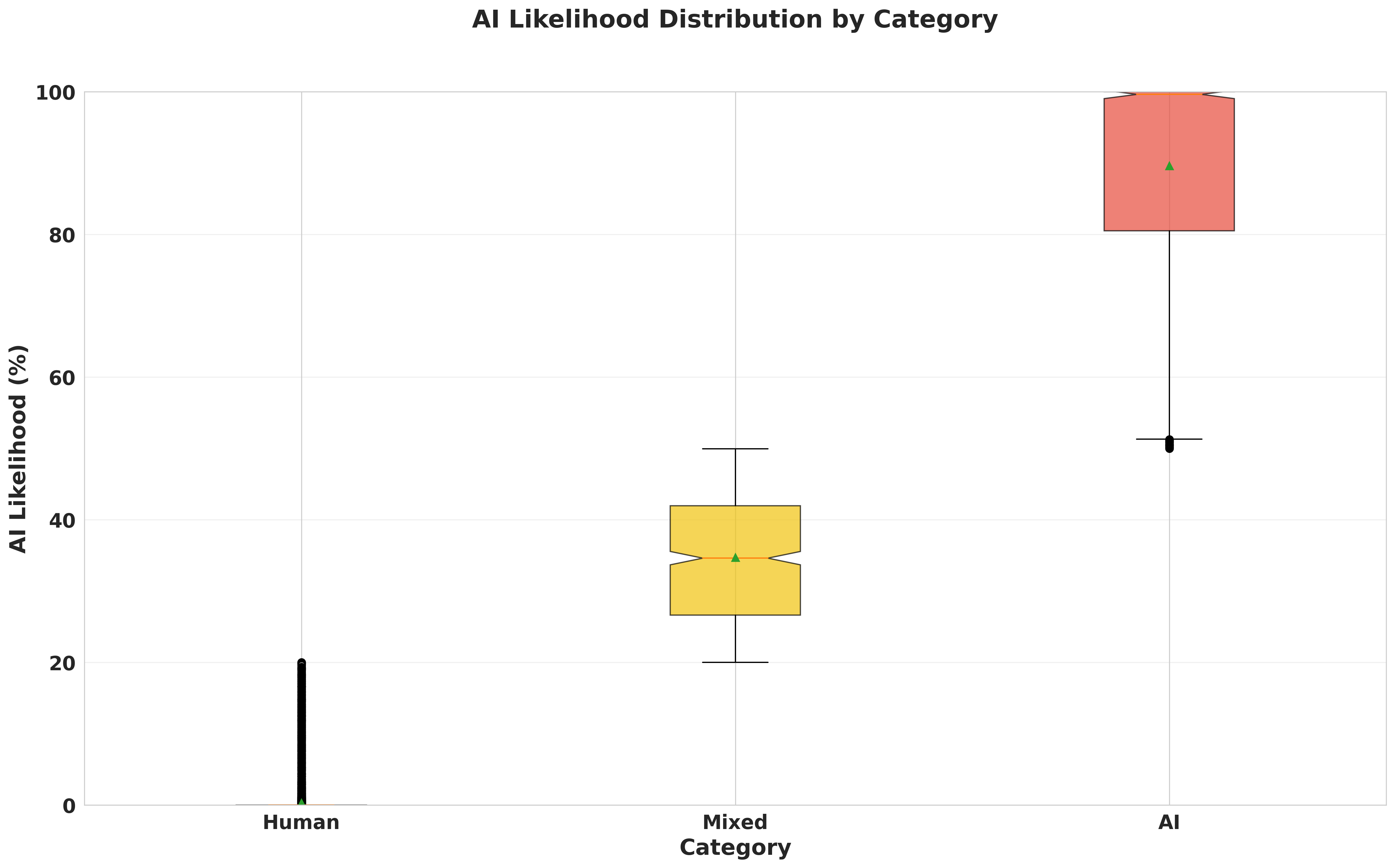}
        \caption{Distribution of AI Likelihoods per each AI Use category in \reporterdata. }
        \label{fig:reporter_box_plot}
    \end{figure}

\paragraph{Label distribution in \reporterdata.} 
\reporterdata has 86.6\% \labelhuman\ articles, with a total of 13.3\% having AI Use. 10.7\% is detected as \labelai\ and 2.6\% is \labelmixed\, as shown in 
\autoref{fig:reporter_pie_chart}. The distribution of AI likelihoods per each of the three labels is shown in \autoref{fig:reporter_box_plot}. More importantly, all \labelai\ labels and almost all \labelmixed\ are concentrated in the last three years (i.e., 2023--2025; see \autoref{fig:reporter_label_distr_year})

\begin{figure*}[t]
  \centering
  \begin{subfigure}[b]{0.45\textwidth}
    \centering
    \includegraphics[width=\linewidth]{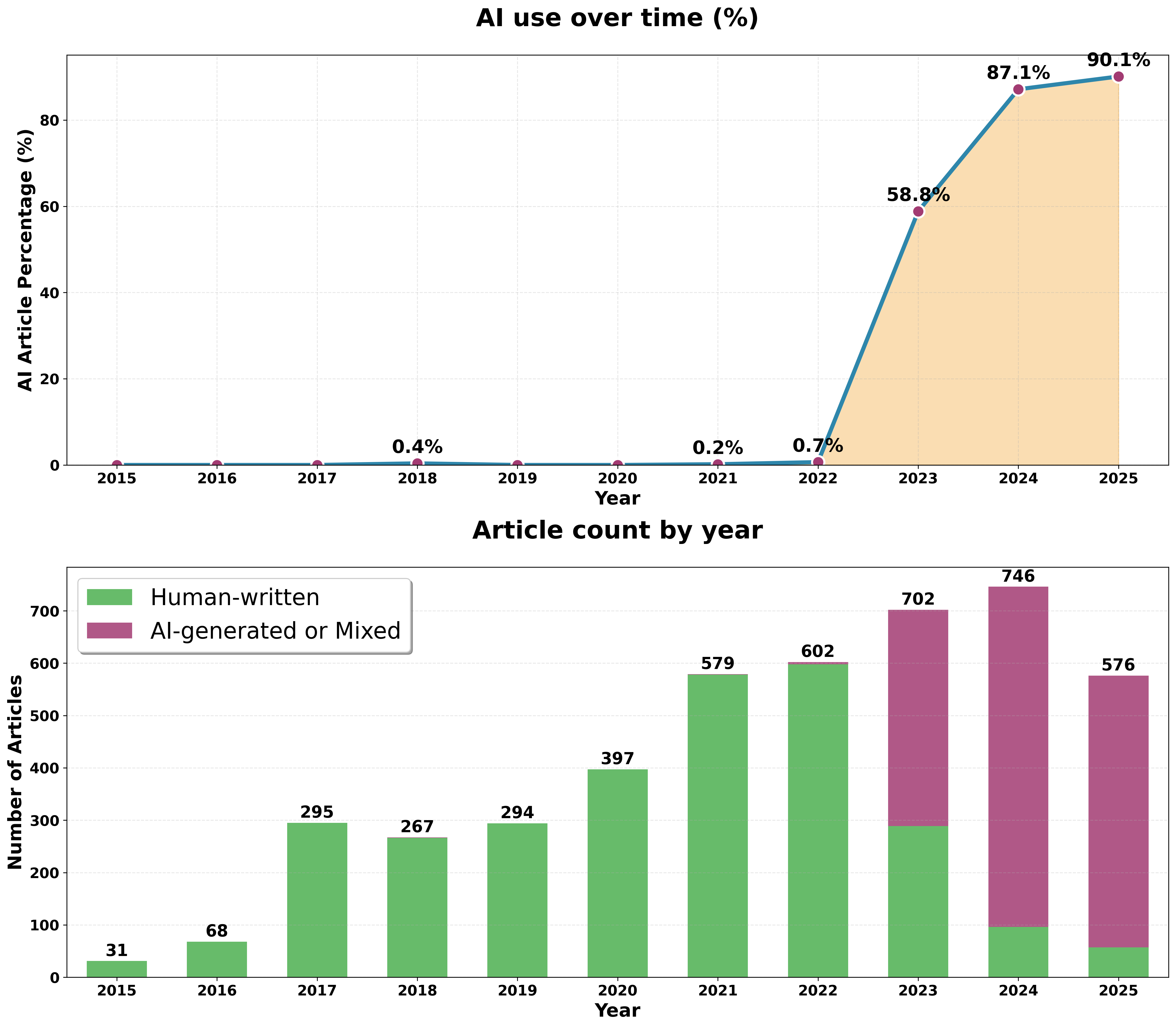}
    \caption{Reporter \faUser\ 1}
  \end{subfigure}
  \hfill
  \begin{subfigure}[b]{0.45\textwidth}
    \centering
    \includegraphics[width=\linewidth]{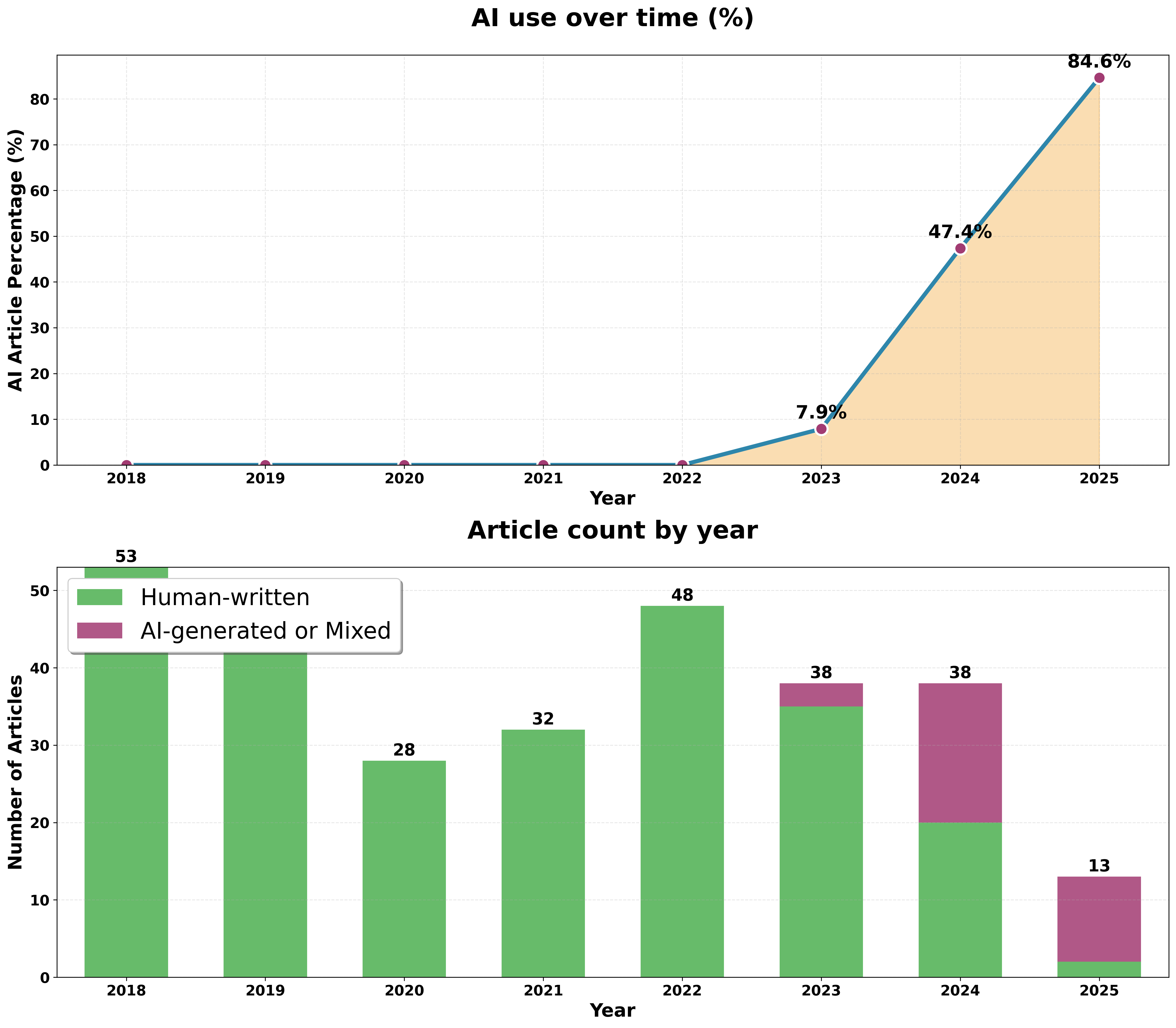}
    \caption{Reporter \faUser\ 2}
  \end{subfigure}

  \vspace{1em}
  \begin{subfigure}[b]{0.45\textwidth}
    \centering
    \includegraphics[width=\linewidth]{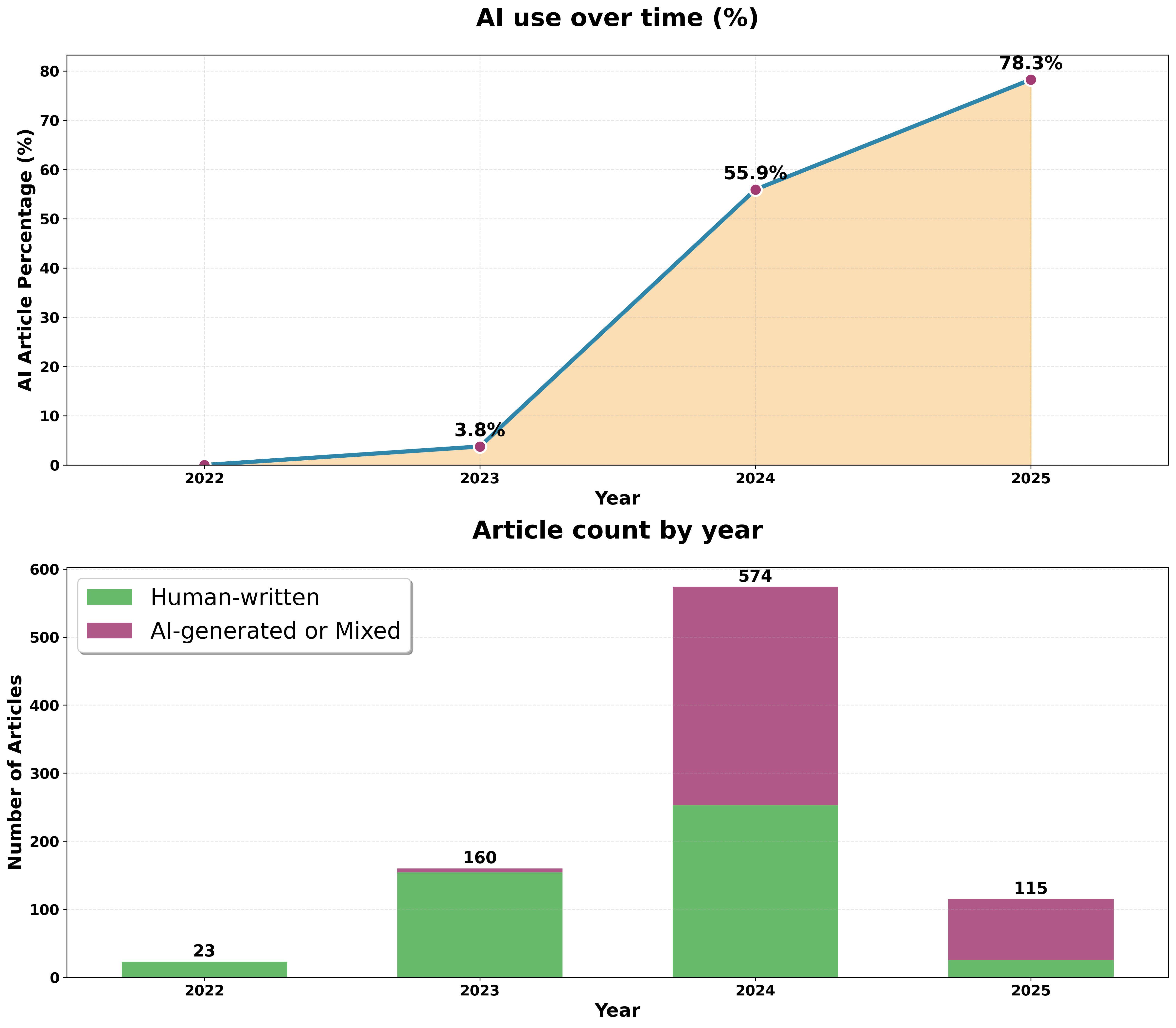}
    \caption{Reporter \faUser\ 3}
  \end{subfigure}
  \hfill
  \begin{subfigure}[b]{0.45\textwidth}
    \centering
    \includegraphics[width=\linewidth]{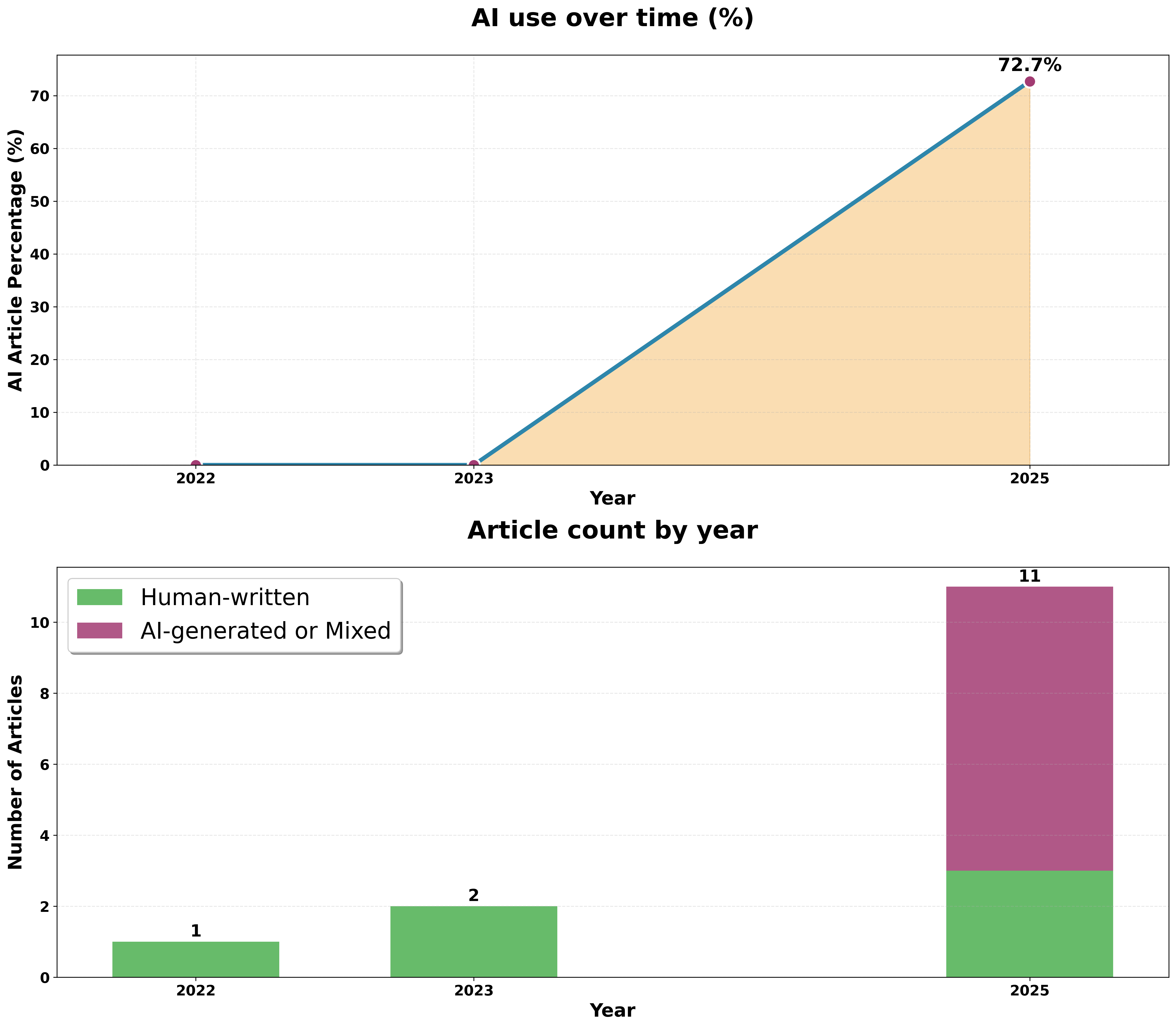}
    \caption{Reporter \faUser\ 4}
  \end{subfigure}

  \vspace{1em}
  \begin{subfigure}[b]{0.45\textwidth}
    \centering
    \includegraphics[width=\linewidth]{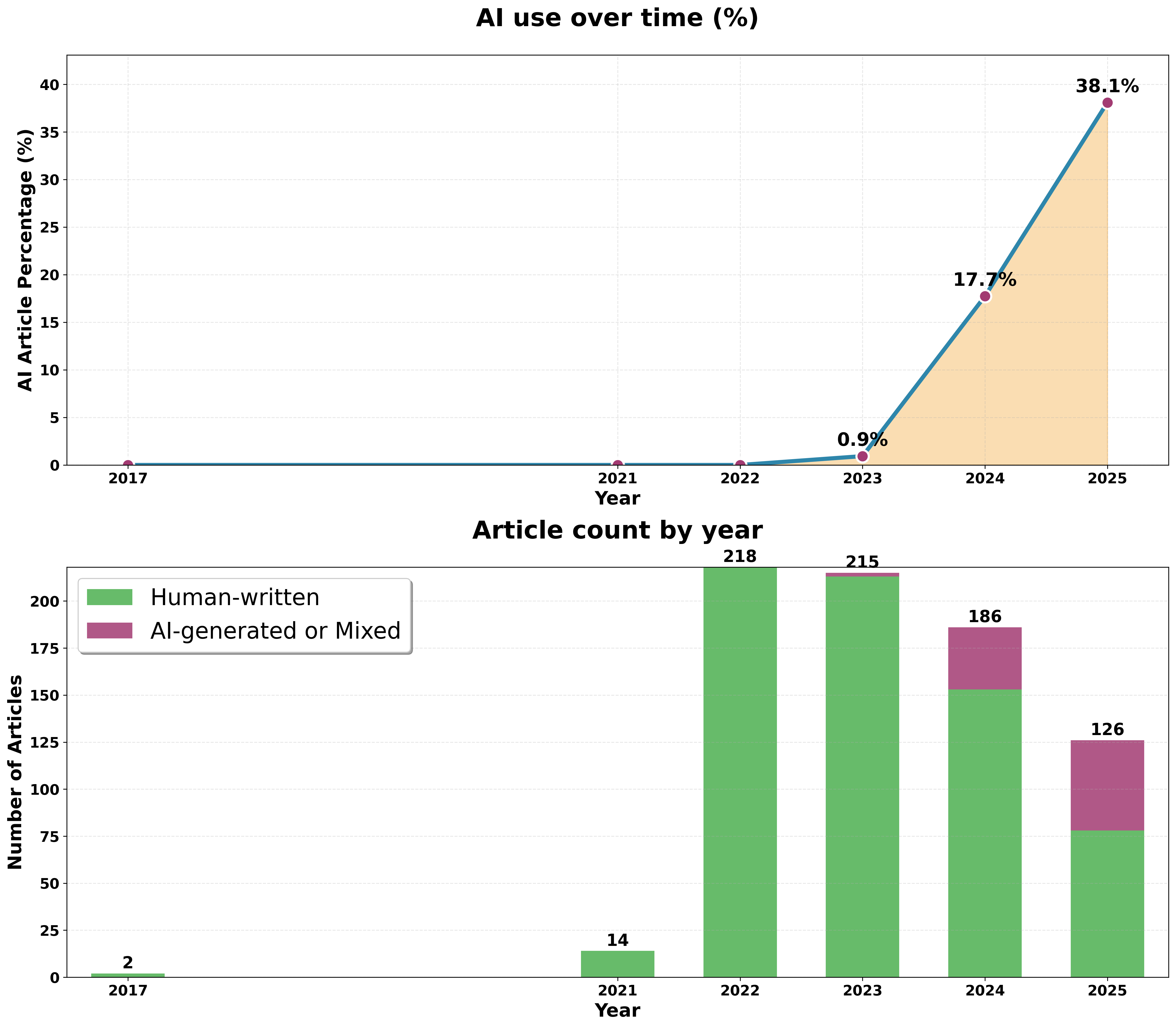}
    \caption{Reporter \faUser\ 5}
  \end{subfigure}
  \hfill
  \begin{subfigure}[b]{0.45\textwidth}
    \centering
    \includegraphics[width=\linewidth]{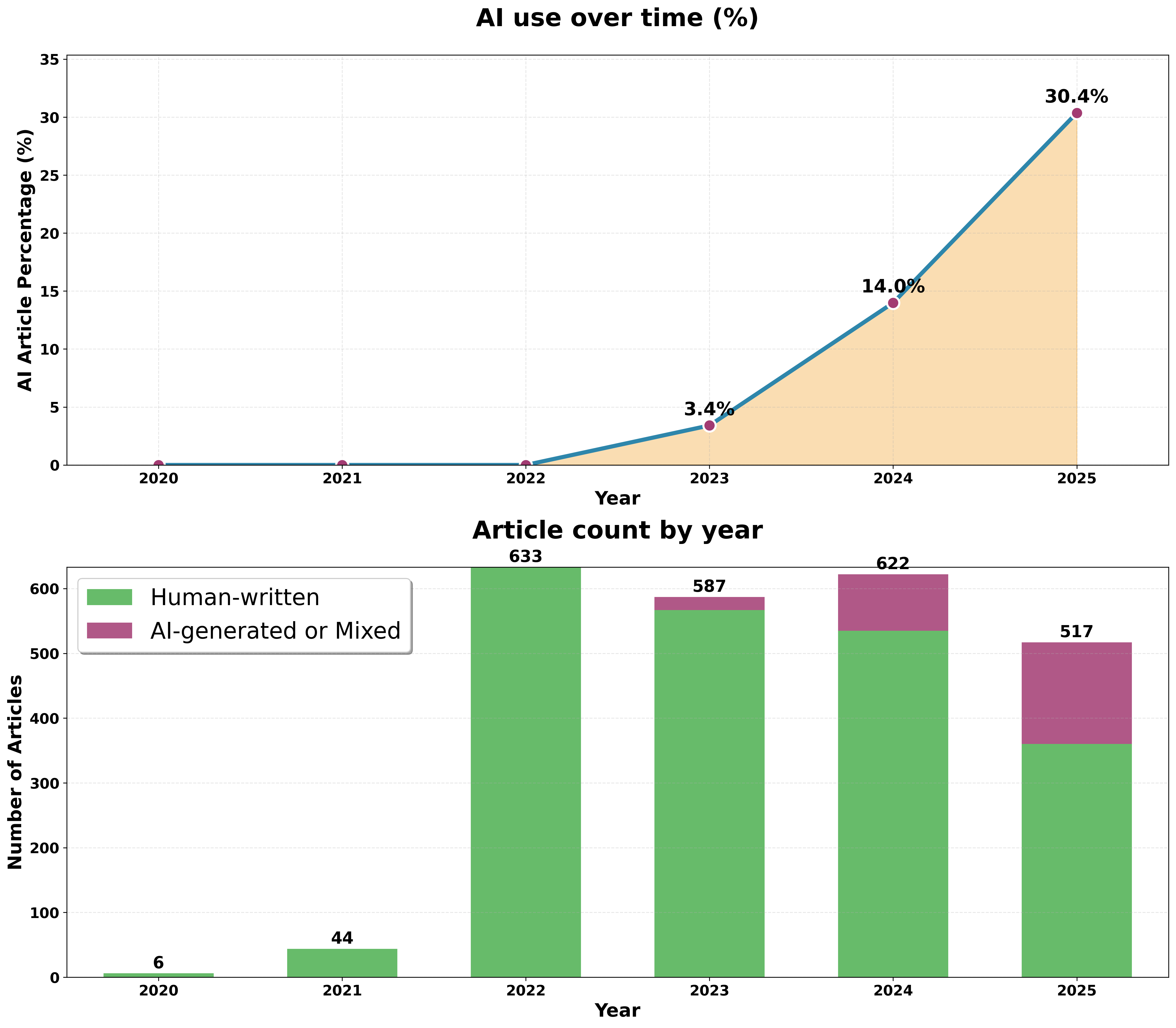}
    \caption{Reporter \faUser\ 6}
  \end{subfigure}

  \caption{AI content patterns in 2025 by reporters (part 1). See Table~\ref{tab:reporters_profiles} for profile alignment.}
  \label{fig:ai_reporters_part1}
\end{figure*}

\begin{figure*}[t]
  \centering
  \begin{subfigure}[b]{0.45\textwidth}
    \centering
    \includegraphics[width=\linewidth]{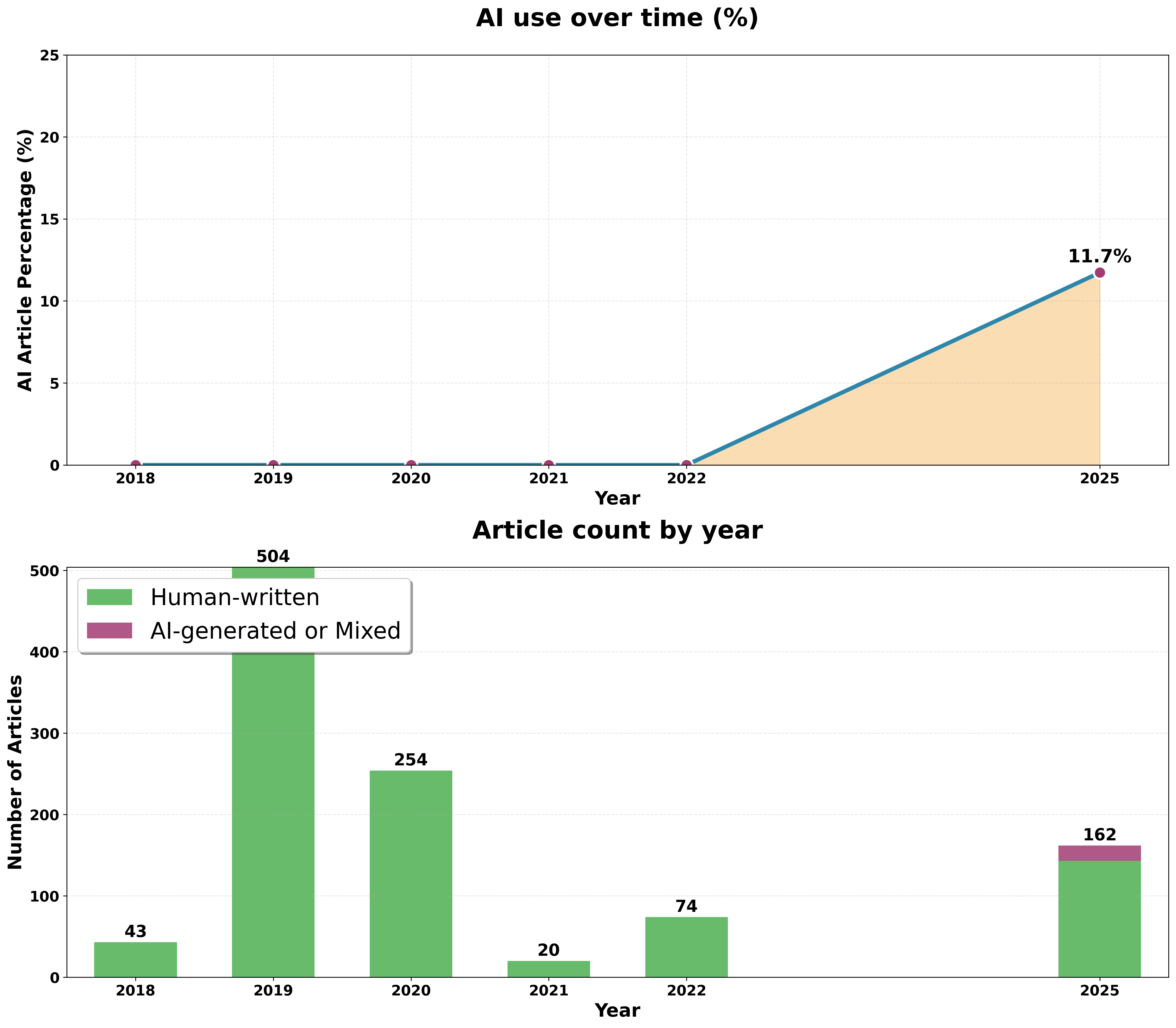}
    \caption{Reporter \faUser\ 7}
  \end{subfigure}
  \hfill
  \begin{subfigure}[b]{0.45\textwidth}
    \centering
    \includegraphics[width=\linewidth]{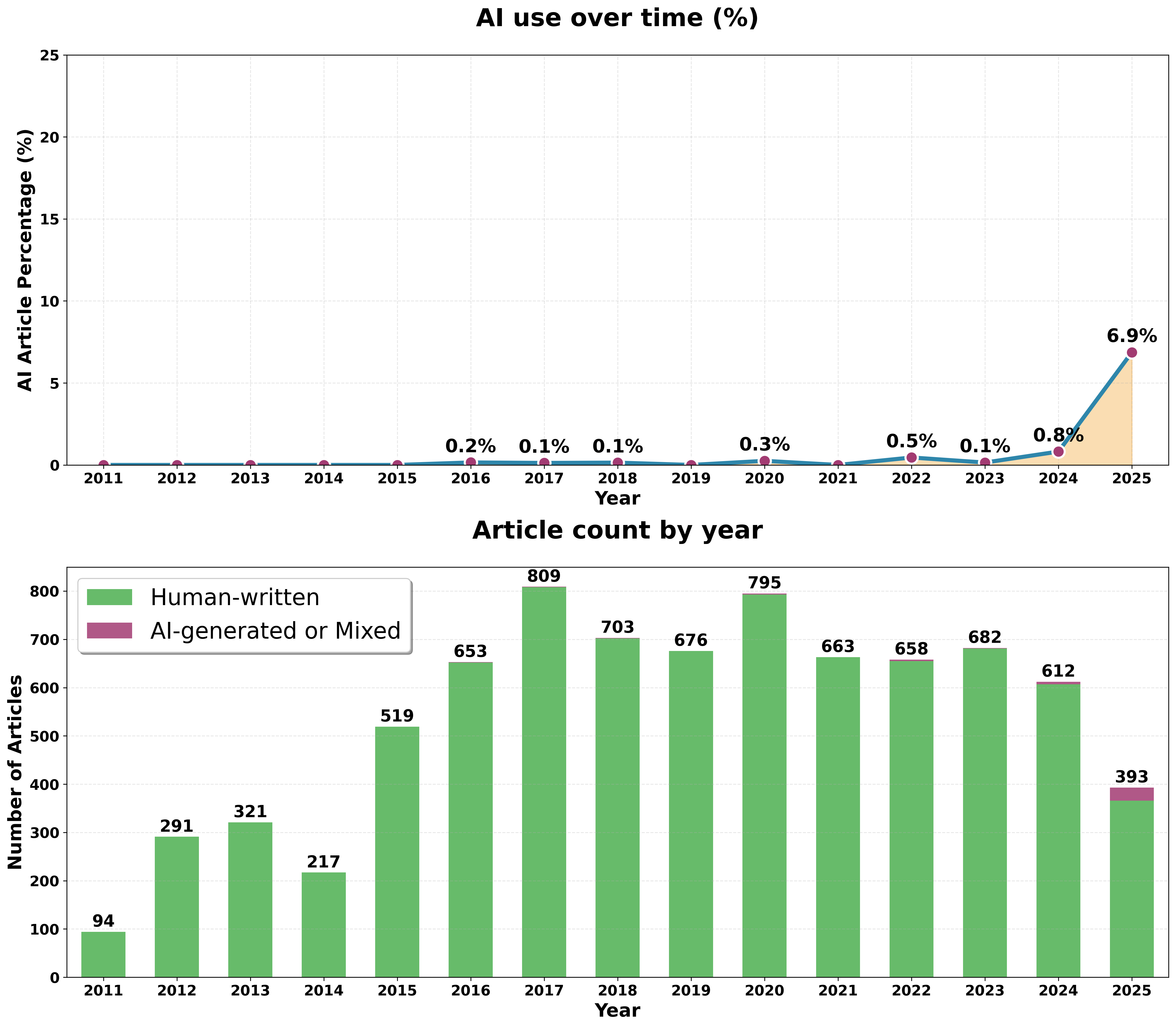}
    \caption{Reporter \faUser\ 8}
  \end{subfigure}

  \vspace{1em}
  \begin{subfigure}[b]{0.45\textwidth}
    \centering
    \includegraphics[width=\linewidth]{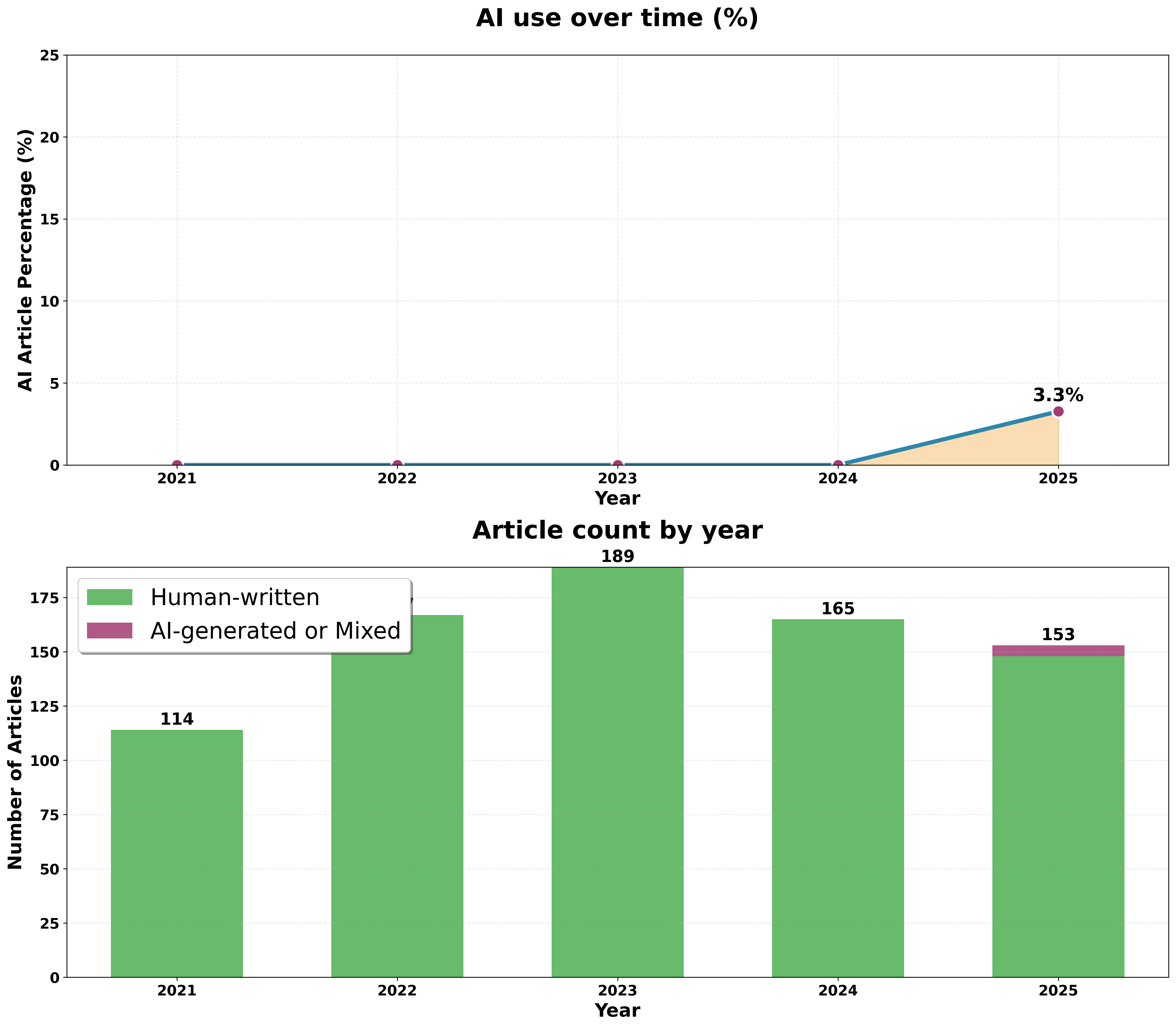}
    \caption{Reporter \faUser\ 9}
  \end{subfigure}
  \hfill
  \begin{subfigure}[b]{0.45\textwidth}
    \centering
    \includegraphics[width=\linewidth]{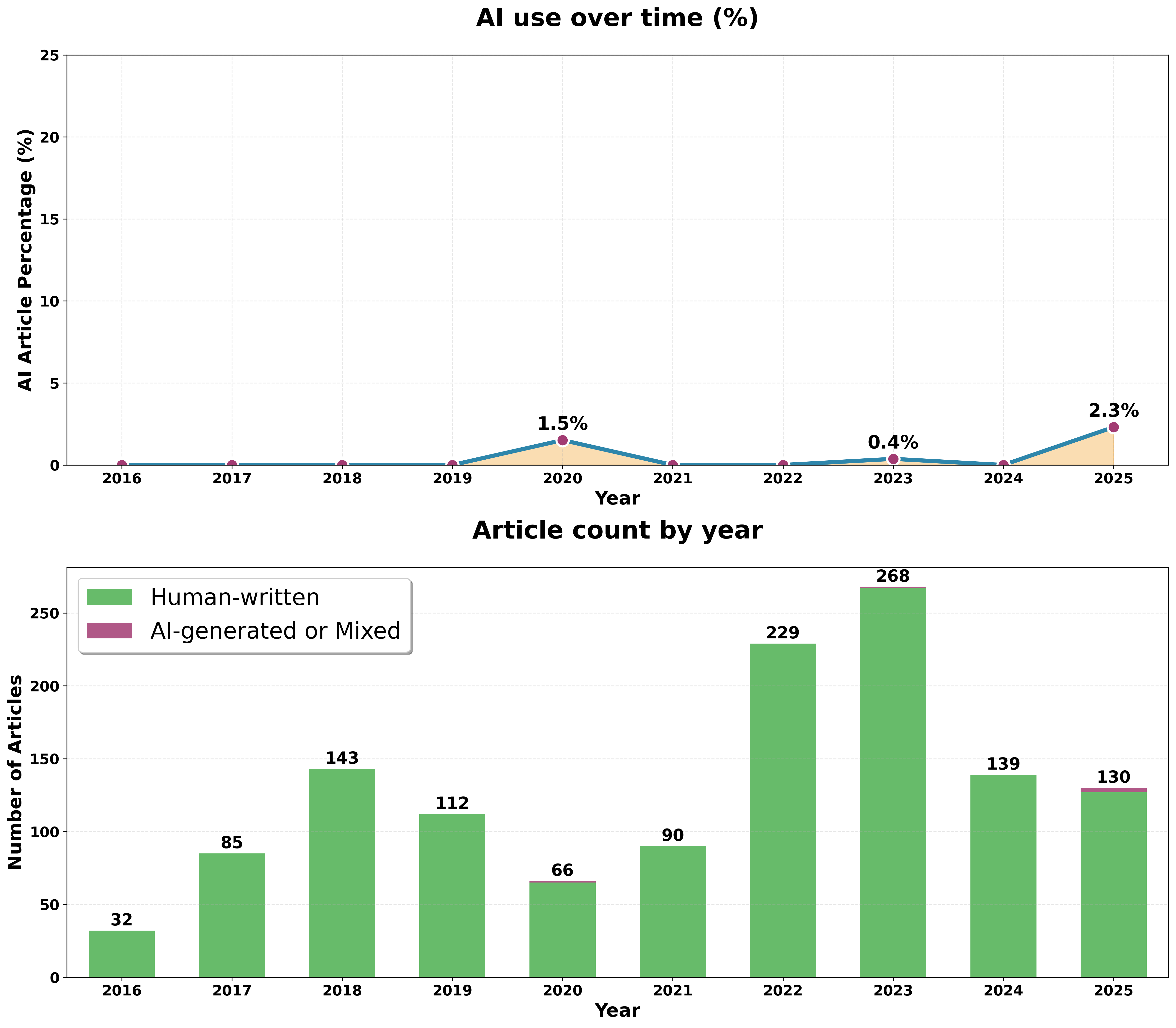}
    \caption{Reporter \faUser\ 10}
  \end{subfigure}

  \caption{AI content patterns in 2025 by reporters (part 2). See Table~\ref{tab:reporters_profiles} for profile alignment.}
  \label{fig:ai_reporters_part2}
\end{figure*}

\paragraph{AI use by reporter.} 
We report AI use by reporter in \autoref{fig:ai_reporters_part1} and \autoref{fig:ai_reporters_part2}. We note a limitation in our data collection: reporters often write for multiple outlets, and accessing historical articles from some of these sources is difficult. Consequently, we could not retrieve all possible articles for every reporter. We did, however, ensure we gathered most of their recent articles and as many as possible from before November 2022 (the ChatGPT release date). 

\end{document}